\documentclass{article} 
\usepackage{iclr2027_conference,times}

\usepackage{amsmath,amsfonts,bm}

\def\eqref#1{equation~\ref{#1}}

\def\1{\bm{1}}

\DeclareMathAlphabet{\mathsfit}{\encodingdefault}{\sfdefault}{m}{sl}
\SetMathAlphabet{\mathsfit}{bold}{\encodingdefault}{\sfdefault}{bx}{n}

\usepackage{hyperref}
\usepackage{url}
\usepackage[normalem]{ulem}

\usepackage{booktabs}
\usepackage{xcolor,colortbl}
\usepackage{pgfmath}
\definecolor{HeatmapRed}{RGB}{194,76,76}
\definecolor{HeatmapBlue}{RGB}{66,119,174}
\newcommand{\HeatmapMaxTint}{30}

\newcommand{\HeatmapShade}[2]{%
  \pgfmathparse{min(\HeatmapMaxTint,max(0,#2))}%
  \edef\heatmapcolor{#1!\pgfmathresult!white}%
  \expandafter\cellcolor\expandafter{\heatmapcolor}%
}

\newcommand{\PassCell}[2][]{%
  \ifdim #2pt<50pt
    \HeatmapShade{HeatmapRed}{\HeatmapMaxTint*(50-(#2))/50}%
  \else
    \HeatmapShade{HeatmapBlue}{\HeatmapMaxTint*((#2)-50)/50}%
  \fi
  {#1#2}%
}

\newcommand{\CostCell}[2][]{%
  \ifdim #2pt<1pt
    \HeatmapShade{HeatmapBlue}{\HeatmapMaxTint*(1-(#2))}%
  \else
    \HeatmapShade{HeatmapRed}{\HeatmapMaxTint*((#2)-1)/2}%
  \fi
  {#1#2}%
}

\usepackage[capitalise]{cleveref}
\makeatletter
\AddToHook{cmd/appendix/before}{\def\cref@section@alias{appendix}}
\makeatother

\crefname{appendix}{Appendix}{Appendices}
\Crefname{appendix}{Appendix}{Appendices}
\crefname{subappendix}{Appendix}{Appendices}
\Crefname{subappendix}{Appendix}{Appendices}
\crefname{subsubappendix}{Appendix}{Appendices}
\Crefname{subsubappendix}{Appendix}{Appendices}

\let\oldappendix\appendix
\renewcommand{\appendix}{%
  \oldappendix%
  \crefalias{section}{appendix}%
  \crefalias{subsection}{subappendix}%
  \crefalias{subsubsection}{subsubappendix}%
}

\newcommand{\myparagraph}[1]{\noindent\textbf{#1}}

\usepackage{graphicx}
\usepackage{paralist}
\usepackage{listings}
\usepackage{xspace}
\usepackage{enumitem}

\usepackage{xcolor}

\usepackage{soul} 

\title{Decentralized Multi-Agent Coordination and Communication in Shared Codebases}

\author{%
    \makebox[0.95\textwidth][c]{%
        \begin{tabular}{@{}c@{}}
            \textbf{Rishabh Mediratta}\thanks{Equal contribution.} \kern0.15em\textsuperscript{1}\quad
            \textbf{Lingyun Xiao}\footnotemark[1] \kern0.15em\textsuperscript{1} \quad
            \textbf{Peter Stone}\textsuperscript{1} \\
            \textbf{Joydeep Biswas}\textsuperscript{1, 2} \quad
            \textbf{Elias Stengel-Eskin}\textsuperscript{1} \\[0.5em]
            \normalfont\textsuperscript{1} The University of Texas at Austin    
            \quad \normalfont\textsuperscript{2}NVIDIA
        \end{tabular}%
    }%
}

\newcommand{\claudeopus}{Claude Opus 4.8\xspace}

\newcommand{\gpt}{GPT 5.6 Sol\xspace}

\newcommand{\deepseek}{DeepSeek V4 Flash\xspace}

\newcommand{\glm}{GLM 5.3 Flash\xspace}

\newcommand{\qwen}{Qwen3.8 Flash Next\xspace}

\newcommand{\swepro}{SWE-Bench Pro\xspace}

\newcommand{\dataset}{ParallACT\xspace}

\newif\ifpreprint
\preprinttrue
\ifpreprint
\iclrfinalcopy
\fi

\definecolor{color5}{HTML}{006795}
\hypersetup{
  colorlinks   = true, 
  urlcolor     = blue, 
  linkcolor    = color5, 
  citecolor   = color5 
}

\begin{document}

\maketitle

\pagestyle{plain}

  \begin{center}    
  \vspace{-3em}
    \textcolor{blue}{\url{https://lingyunxiao18.github.io/ParallACT/}}
    \end{center}

\begin{abstract}
    
    Coding agents working in shared repositories may receive separate, independently-motivated requests from human developers, creating a setting that is \emph{multi-agent by necessity}. When this scenario happens, their individual implementations may work separately but conflict when combined. 
    To study this setting, we introduce \dataset{}, a benchmark of orchestrator-free multi-agent coding tasks.
    \dataset{} contains both complementary and contradictory tasks. 
    In complementary tasks, each agent's task is separately useful and implementations must share an interface to be productively combined. 
    In contradictory pairs, both agents' tasks require editing the same code in incompatible ways, and naively combining independent solutions fails to produce a useful implementation. 
    We compare a single-agent oracle given the combined set of tasks against multi-agent settings that vary execution order, visibility, awareness of other agents, and access to a communication channel. 
    We find that the oracle solves most tasks while parallel agents perform substantially worse. 
     Their performance drop is notable for complementary tasks, and particularly sharp for contradictory tasks (e.g., from $62\%$ to $16\%$ for Claude Code with \claudeopus{}) where merged code typically satisfies only one specification, and the same gap appears in all six further configurations pairing two agent harnesses with three open-weight models.
    Simply telling an agent that a peer exists does not help significantly,
    and giving agents access to a communication channel recovers only a small part of the performance, at an additional token cost.
    \dataset{} thus identifies and isolates a core challenge that will become more common as  multi-agent coding by independent principals becomes standard, 
    and provides a controlled test of which coordination strategies help, and at what cost. 
\end{abstract}

\section{Introduction}
\label{sec:introduction}
\begin{figure}[t]
    \centering    \includegraphics[width=\linewidth,trim=17pt 0pt 0pt 0pt, clip]{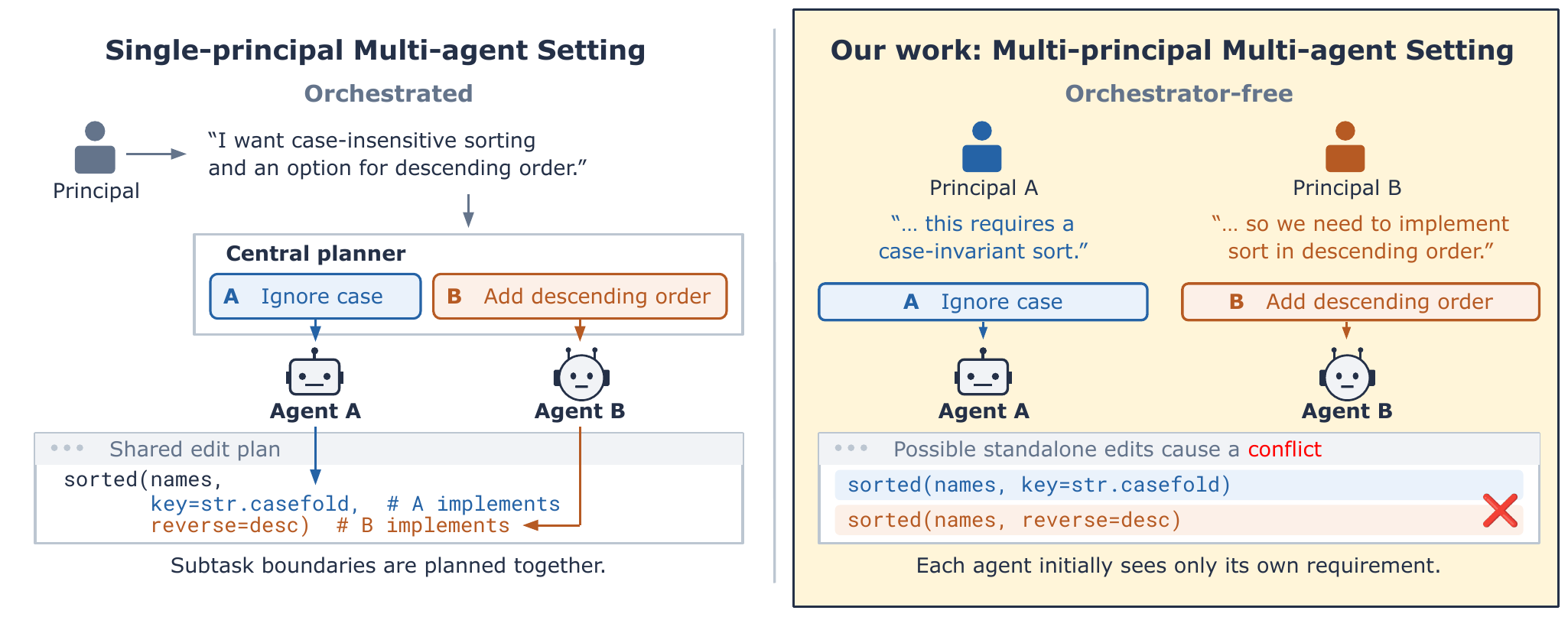}
    \caption{Centralized and decentralized assignment of coding tasks, illustrated with two sorting requirements: A requests case-insensitive sorting, while B requests support for descending order. Left: a central planner sees both requirements and coordinates edits to a single sorting call that combines case normalization with an optional reversal. Right: separate principals assign the requirements independently, and each agent initially sees only its own request. The agents may replace the same sorting call with individually valid implementations that conflict when combined. 
    } 
    \label{fig:fig1}
\end{figure}

The digital world is driven by large codebases
which hundreds or thousands of contributors work on simultaneously on independently-motivated and potentially conflicting tasks, often without knowledge of each other. 
Recently, coding workflows have begun to include a new kind of contributor: 
agents powered by large language models (LLMs), which can transform complex natural language specifications from a principal (in this case, a developer) into code \citep{yang2024swe}.
In principle, LLM agents seem to be a good fit for large-scale development: indeed, they are increasingly being used in multi-agent settings, with large swarms of agents working in parallel towards a shared goal \citep{qian2025scaling, lin2026scaling}.
This setting is illustrated in \cref{fig:fig1} (left), where a single principal gives a task to a central planner that decomposes it into compatible subtasks and distributes them across agents. 
However, real-world multi-principal coding environments often differ from this centralized picture.
In large-scale development, principals may not share goals, have conflicting specifications, and may not be aware of the other principals or their goals, as illustrated in \cref{fig:fig1} (right). 
Naively implementing specifications in this setting can lead to code that fails to meet either specification.

These independently-motivated settings differ in that they are \emph{multi-agent by necessity}, rather than convenience: in the centralized picture, it may happen to be that decomposing the task and assigning it to different agents improves performance or latency, but the tasks the agents receive ultimately stem from a single, shared intent. 
In contrast, the multi-principal setting shown on the right is irreducibly multi-agent; requests arise independently, and cannot be assigned to a single agent without aggregating them.
In large-scale development, this is impractical, as the number of tasks the centralized orchestrator must handle scales with the number of principals.
While it may be able to orchestrate a handful of tasks, many large-scale codebases would require aggregating requests from thousands or tens of thousands of principals in a central orchestrator, introducing a bottleneck, increasing latency, and creating a single point of failure \citep{Stone2000Multiagent}. 
Thus, agents must be able to implement their principals' requests in a decentralized fashion.

In this decentralized picture, one way to avoid conflicts is through communication and coordination, 
both explicit (e.g., sending messages to other developers, opening issues or pull requests) 
and implicit (e.g., inferring the likely goals of others from the environment or history, and shaping code to be compatible with those goals). 
For example, a human developer in a shared repository might infer another developer's goals from their commit history and avoid interfering, or might send that developer a message to coordinate on a shared interface. 
Past work has studied communication between coding agents \citep{khatua2026cooperbench, zhan2026agensh} but has focused on a team setting, in which agents are aware that they are working on shared tasks with teammates.
This kind of mutually-aware agent deployment has recently been validated at scale on problems such as proof search \citep{xin2025scaling, openai2026navierstokes}. 
We instead study \textbf{what happens when principals who may not initially be aware of each other delegate tasks to agents in environments which lack a central orchestrator and may feature conflicting or contingent specifications.} 
Understanding coordination in this setting is critical: assuming that agents will increasingly be taking over writing code from human developers, they will have to work in large, shared codebases where many other \emph{independently motivated} agents are concurrently implementing changes, without any explicit pre-coordination between agents.
Moreover, in addition to understanding the performance of agents under different multi-agent and coordination settings, we would like to measure the cost of different coordination approaches. 

To study these questions, we introduce \textbf{\dataset{}} (\underline{Parall}el \underline{A}gent \underline{C}oordination \underline{T}estbed), a diagnostic dataset of 155 challenging, self-contained multi-agent coding instances. The goal of \dataset{} is to provide a targeted testbed for understanding collaboration in orchestrator-free, multi-principal settings. While motivated by larger teams, to isolate coordination failures in a controlled fashion, we consider the minimal unit of a pair of agents. \dataset{} consists of a manually-constructed gold split (i.e., seed instances, each containing a repository at a given commit, two independently-motivated requirements, reference implementations, and an evaluation test suite) with manually-constructed competing specifications, and a silver split (i.e., model-generated instances), which is automatically constructed by \gpt{} using gold standard examples as seeds. 
In both sets, requirement A comes from \swepro{} \citep{deng2025swe}, while requirement B is written (manually or by a model) to be a separately-motivated counterpart.
In total, \dataset{} contains 85 contradictory and 70 complementary instances (\cref{appendix:dataset-composition,appendix:swepro-versions}); simplified examples of each family are shown in \cref{fig:dataset}.  
Contradictory instances provide competing specifications, and are filtered for examples that feature a true behavioral conflict: we ensure that reference solutions that meet each specification independently cannot be merged into a valid joint solution.
Complementary instances, on the other hand, require agents to agree upon a shared interface, and are tested via integration tests; here, we filter examples to ensure that no independent solution to either agent's problem passes the tests alone.
These filters ensure that success requires deep coordination between agents, and each instance is manually verified  (\cref{appendix:dataset-composition}).

Using our data, we test seven configurations of agent harness and model: Claude Code with the closed-weight \claudeopus{}, and both Claude Code and Codex with each of three open-weight models (\qwen{}, \deepseek{}, and \glm{}).
We first establish that the instances provided are achievable by a single developer: when both specifications are provided to a single agent, success rates are fairly high, with \claudeopus achieving a 62\% success rate on contradictory and 86\% on complementary instances, which increases to 98\% and 100\% when the agent is permitted to debug using provided tests. The open-weight configurations reach 60\% and 72\% on average, and 96\% and 97\% with tests.
We then test different multi-agent topologies and communication strategies. 
To set a topline of what can be done with a manually-specified topology, we test our problems under sequential and re-entrant settings, where agents take turns editing the codebase (e.g., Agent A, then Agent B, potentially iterating in the re-entrant case).
Under this fixed setting, we find that agents perform worse than the single-agent oracle but still succeed on around half of the cases in both conditions. 
However, this setting is unrealistic; while it may be possible to specify fixed topologies for $n=2$ agents, it quickly becomes infeasible as $n$ increases (e.g., imagine scaling to 100 agents, where each agent must block on 99 other agents before making its edits). 

We then test agents across multiple parallel settings, where agents work on their specifications simultaneously. 
Here, we study several ad-hoc communication strategies, optionally
providing agents with access to either a single shared file to write updates to, or a Redis cache (\cref{appendix:redis-documentation}) in which they can send messages.  
While each agent generally succeeds in implementing solutions that pass their own respective tests, they largely fail to coordinate on a joint solution, with two mutually-unaware \claudeopus agents achieving an average joint pass rate of only 16\% on contradictory and 57\% on complementary tasks (19\% and 53\% averaged over the open-weight configurations). 
While these drops can sometimes be mitigated by adding communication and making agents aware of each other, performance remains substantially lower than in the single-agent condition. 
Moreover, we find that communication increases costs. 

Building on these results, we analyze the failure modes of different topologies, finding that single-agent failures can be traced to capability shortcomings while multi-agent settings are bottlenecked by coordination failures.
Moreover, we study the communication channel in the parallel setting, finding that agents often succeed in terms of \emph{what to communicate} -- communicating about the right topics (e.g., the shared artifact) -- but often fail in terms of \emph{how to communicate and act on communication}, e.g., failing to read messages on time or at all, or agreeing to a design but not following through.  

\section{Problem Statement}
\label{sec:problem_statement}

\begin{figure}[t]
    \centering    \includegraphics[width=\linewidth,trim=15pt 0pt 15pt 0pt, clip]{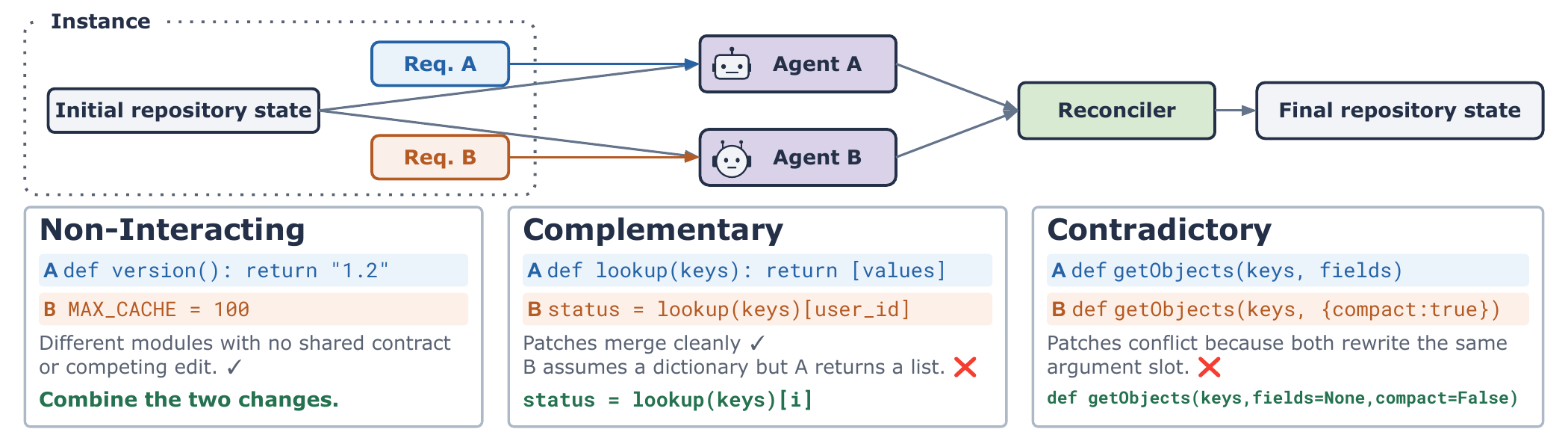}
    \caption{Task interactions in a shared codebase. Top: each agent receives its own requirement and the initial repository.     
    A reconciler (e.g. patch concatenation, Git merge, or model-based resolver) combines their edits into one codebase and may resolve some conflicts before the result is evaluated for joint correctness. Bottom: (a) Non-interacting: updating a version number and setting a cache limit affect separate modules. (b) Complementary: a lookup function returns a list, but its caller expects a dictionary, so the edits merge cleanly but fail to interoperate. (c) Contradictory: field selection and compact output compete for the same argument position, and separate parameters are needed to accommodate both. Green text shows possible joint solutions.}
    \label{fig:dataset}
\end{figure}

We study a scenario common in large-scale software engineering: \textit{multiple programmers pursuing separate tasks on a shared codebase}, each delegating work to their own coding agent. 
Each agent is responsible for its own programmer's request and may be \textit{unaware} of other agents or their requirements, even when their changes affect the same codebase or must interoperate through a shared interface. 
To isolate the coordination challenges in this broader scenario, we focus on the smallest nontrivial case: two agents with independently motivated and jointly satisfiable requirements. 
This setting captures conflicts and dependencies that can also arise within larger teams. 
We examine how peer awareness, communication, and execution schedules affect joint correctness and token cost.

An \textit{instance} consists of an \textit{initial repository state}, two separate \textit{requirements} in natural language, \textit{reference implementations}, and an \textit{evaluation test suite}. Each requirement is assigned to a separate coding agent and, by default, is disclosed only to that agent. Through code changes and, in some settings, communication, the agents must produce a single \textit{final repository state} that satisfies \textbf{both} requirements. Each requirement comes with a reference implementation, a known correct code change that satisfies the given requirement, and an evaluation test suite that scores the final repository state. Both the reference implementations and the evaluation test suite are withheld from the agents, except in settings that explicitly grant access to the tests.

Broadly speaking, requirements can interact in three different ways within the initial repository, shown in \cref{fig:dataset} (bottom): (a) \textbf{Non-interacting} 
requirements admit separate implementations that can be correctly combined with only surface-level reconciliation; (b) \textbf{Complementary} requirements specify individually-motivated tasks that must agree on a shared interface or contract to satisfy an integration test, and either implementation alone fails;
(c) \textbf{Contradictory} requirements specify 
tasks that require implementing or modifying the same code region in ways that would fail when mechanically combined, but that can in principle be reconciled to satisfy both tasks. 
We focus on the latter two families, as they require non-trivial coordination between agents. 
Each requirement is associated with a set of \emph{acceptance conditions} designed as executable tests.
We evaluate the final repository state against the acceptance conditions of both requirements, on the assumption that the final repository should satisfy both users' goals. 
\emph{Joint success} requires all conditions to be satisfied, while the fraction of satisfied conditions provides a more lenient measure of partial completion. \cref{appendix:example-instances} walks through one contradictory and one complementary instance with their requirements, reference implementations, and evaluation test suites.

A \emph{coordination strategy} specifies four aspects of execution: (a) The \textbf{execution schedule} determines when each agent acts and whether their actions overlap in time; 
(b) The \textbf{observation model} determines what each agent can see of the repository and the other’s work; 
(c) The \textbf{communication model} determines whether and how agents can exchange information beyond changes to the code; 
(d) The \textbf{combination rule} determines how separately produced changes become one final state, including any permitted reconciliation or repair. 
These choices determine both the information available to the agents during implementation and the opportunities to resolve incompatible decisions. 

We evaluate coordination strategies by \textbf{joint success rate}, \textbf{completion rate}, and relative effective \textbf{token cost}. 
A single agent given both requirements provides an \textit{oracle} for implementation with access to the full task specification -- we consider this to be an oracle because it deviates from our motivating setting, in which multiple principals implement changes in parallel, and this would require aggregating all principals' requests centrally before starting any task, which is impractical at scale. 
In this oracle setting, 
a single agent has access to all task-relevant information at once. The single-agent oracle measures to what degree the models we are testing are capable of executing on the requested specifications when all information is available, setting a topline for their abilities. 
Relative to this ceiling, we ask how joint success changes when agents work concurrently without awareness of one another. 
We then examine which components of a coordination strategy recover success and at what cost, whether effective communication is constrained by the available channel or by how agents use it, and how much a combination rule can repair after separate implementation. 

\section{Benchmark Construction and Experimental Settings}
\label{sec: benchmark_construction_and_experimental_settings}

This section describes how we construct \dataset{} and the coordination scenarios tested. 

\subsection{Benchmark Construction}

\paragraph{Data sourcing.} 
We build \dataset{} starting from  \swepro{} \citep{deng2025swe}. 
For any given instance in our dataset, we take requirement A from the original \swepro{} task and introduce a new requirement B, which is a separately-motivated task that a different developer could be tasked with, which merely happens to touch the same code. In contradictory instances, tasks A and B collide on shared code, while in complementary tasks they must agree on a shared interface. \swepro{} was revised after we had built our first instances.
We build instances from both the original release (v1) and the revision (v2), on disjoint sets of source tasks, and treat them as one dataset throughout;
\cref{appendix:swepro-versions} describes the two releases, how we verified that the v1 instances remain correct, and how the v2 instances were built.

\paragraph{Testing and filtering.}
Each instance starts from an initial repository state and includes a problem statement for each of task A and B, 
a per-side reference solution for each (a correct implementation of that one requirement), and a set of acceptance tests.
We store a reference solution as a \emph{patch}: a set of edits that, when applied to the original repository, implements that task's requirement. 
For contradictory instances, we additionally add a combined test set, which runs both tasks' acceptance tests together, 
and a \emph{reconciliation solution}: 
a single implementation that simultaneously satisfies both requirements. 
The reconciliation solution is distinct from the two per-side reference solutions since each of those solve their respective tasks; because A's and B's solutions collide on the shared code, stacking them does not give a working joint solution. For complementary instances, we define an integration test that passes only when both sides agree on the interface they share. 
Outside the oracle, communication oracle, and gold-test conditions (defined in \cref{subsec:experimental-settings}), problem statements refer only to their own requirements and do not mention the other task, its files, or any of the tests.

\begin{figure*}[t]
    \centering
    \includegraphics[width=\textwidth, trim=1pt 0pt 1pt 0pt, clip]{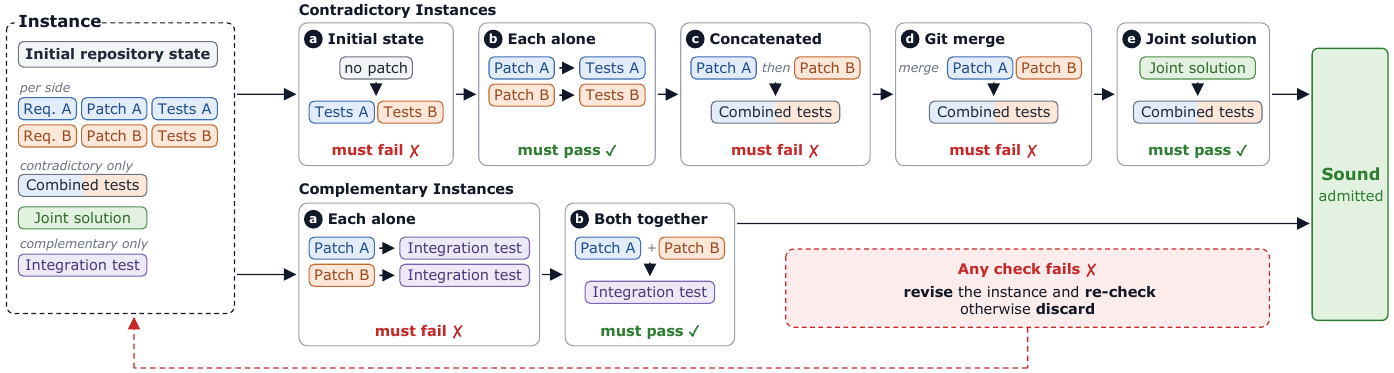}
    \caption{Soundness checks for benchmark instances. Each instance starts with an initial repository state and a separate requirement, reference patch, and acceptance tests for each task. Contradictory instances must pass all five checks in the upper row. Complementary instances must pass both checks in the lower row. An instance that fails a check is revised and rechecked or discarded.}
    \label{fig:benchmark-filtering}
\end{figure*}

We filter instances for soundness according to the following procedure, shown in \cref{fig:benchmark-filtering}.
Starting from the repository at its original commit as defined in \swepro{}, we apply:
task A's patch by itself, task B's patch by itself, both patches applied one after the other, and both patches combined through a Git merge. 
After each of these applications, we run the tests that apply to that case -- a single side's own acceptance tests, the combined test set, or the integration test -- and filter based on the following rules, which ensure that the two requirements interact as intended and that no naive combination of the per-side solutions already produces a valid joint solution.
A contradictory instance is sound when:
\begin{inparaenum}[(a)]
\item The unmodified initial repository fails both sides' tests.
\item Each agent's reference solution applied alone passes its respective tests.
\item Applying concatenated patches fails the combined test set.
\item Applying both via a git merge also fails the combined test set.
\item The reconciliation passes the combined set.
\end{inparaenum}
A complementary instance is sound when:
\begin{inparaenum}[(a)]
\item Each agent's reference solution applied alone fails the integration test.
\item Applying both together passes the integration test.
\end{inparaenum}

\myparagraph{Seed set.}
We first hand-build a gold  standard set of \emph{seed} instances (the manually
constructed instances recounted in \cref{appendix:dataset-composition}) 
each developed from one
\swepro{} task, taken as task A.
Task B, its reference solution, its tests, and the family-specific artifacts (the
combined test set and reconciliation, or the integration test) are written by the authors, ensuring they pass the filtering conditions. 
Each seed instance was further checked manually after passing the filters. \dataset{} contains 27 seed instances and 128 generated instances (58 built on v1 and 70 on v2; \cref{table:dataset-stats}).

\myparagraph{Data generation.}
We scale the benchmark beyond the seeds by generating further instances with \gpt{} in Codex (\cref{appendix:codex-configuration}), again keeping requirement A as a \swepro{} task and
authoring requirement B as an independent counterpart. 
Diversity is enforced by selecting candidate tasks via a deterministic, stratified sweep over
repository metadata (first programming language, then repository). 
We use a smaller model (\cref{appendix:qwen-configuration}) to route candidates to the complementary or contradictory group. 
We then prompt \gpt{} to produce task B requirements and artifacts, using tasks from the seed set as in-context examples.
After authoring, each instance is checked against the soundness conditions for its family (\cref{fig:benchmark-filtering}) 
, and any failure triggers an automatic correction round in which \gpt{} revises the instance and it is re-checked; we allow up to two such rounds, after which an instance that still fails any condition is discarded.
We give the full generation procedure in \cref{appendix:dataset-creation-details} and give the full breakdown by repository and family in \cref{appendix:dataset-composition}. 

\subsection{Experimental Settings}
\label{subsec:experimental-settings}

We evaluate each instance under \emph{coordination scenarios} that vary
three things: the \emph{schedule} (when each agent acts and whether they overlap),
the \emph{observation model} (what each agent sees of the repository and the
other's work), and the \emph{communication model} available beyond code.
A \emph{turn} is one agent invocation on a workspace.
We run 19 contradictory and 18 complementary scenarios; \cref{sec:methods} describes how changes are combined and scored, and \cref{fig:strategies} (Appendix) shows the four groups compared: single-agent oracle, serial hand-offs, parallel execution, as well as gold-test access. The exact prompt text for every scenario is shown in \cref{appendix:scenario-prompts}. 

\myparagraph{Single-agent references.} \emph{Oracle} gives one agent both requirements
and asks for the whole solution, an upper bound requiring no coordination.
\emph{Orchestrator} also gives one agent both requirements but asks it to delegate each to a separate subagent and do no implementation itself.

\myparagraph{Serial (hand-offs).} 
Two agents edit the same workspace in turn, 
each starting from the previous result. 
\emph{Sequential} runs two turns in one order
(A then B, or B then A); 
\emph{reentrant} runs four alternating turns (ABAB or
BABA). 
In both, we additionally vary the observation model: whether the acting agent sees only the current file
state (\emph{repo}) 
or also sees the earlier agent's file edit history (\emph{hist}).
Note that this strategy scales poorly as the number of agents increases. We expect a single handoff to help agents account for earlier edits. We test whether more turns and access to edit history improve joint success or lead agents to overwrite earlier work by contrast.

\myparagraph{Parallel.} Agents work simultaneously from the same base,
and their changes are kept separate until evaluation.
They are either \emph{unaware} that a peer exists,
\emph{aware} of a peer but without a communication channel,
or given a two-way message channel framed as free-form \emph{messaging} or as explicit \emph{synchronization} points. The framings differ only in how agents are prompted to use the channel: under \emph{messaging} they may send and read at any time and in any form, whereas under \emph{synchronization} they treat the channel as explicit checkpoints, announcing each intended change and reconciling it with the peer before continuing.
We run both framings over two communication modalities.
In the first, agents share a single append-only file: to send, an agent appends its message; to receive, it re-reads the file for new entries.
In the second, messages go through Redis, where each agent has its own inbox: sending pushes a message into the peer's inbox, and receiving reads and clears the agent's own. The send/receive interface is identical in both cases.
Finally, a \emph{communication oracle} extends parallel messaging by giving each agent its partner's requirement verbatim, with an instruction to communicate and solve both problems together.
This isolates whether using the channel -- rather than missing information about the peer's task -- is the bottleneck. We expect parallel-unaware agents to achieve lower joint success than the single-agent oracle and ask how large this gap is. We test whether peer awareness, messaging, or synchronization reduce the gap, and at what token cost. The communication oracle tests whether revealing the partner's requirement helps beyond providing a channel, and we expect it to outperform standard communication settings.

\myparagraph{Gold-test access.}
Two further oracle-style scenarios establish performance upper bounds by letting agents debug against the gold tests paired with each instance.
\emph{Oracle + gold tests} gives the single-agent oracle the gold tests, and \emph{parallel + gold tests} gives the
parallel-unaware agents their respective tests.
The parallel variant covers only contradictory instances, since complementary instances have only a shared integration test, which cannot evaluate one agent's implementation without the other's. In this setting, we test whether individually correct implementations are sufficient for joint success, and whether coordination failures persist even when agents can debug their solutions against the provided tests.

\subsection{Evaluation Methods and Metrics}
\label{sec:methods}

\paragraph{Merge policies.} Parallel scenarios require merging: each agent produces an
independent edit from the same starting point, and the two must be combined into a
single codebase before grading. 
We combine them in three ways of increasing
sophistication, following \citet{khatua2026cooperbench}. 
\emph{Concatenation} applies both edits together;
\emph{Git merge} merges them as two branches off the base and treats a merge conflict
as a failure; and \emph{reconciliation} goes further, resolving any conflict with an
off-the-shelf merge-resolver model from \citet{khatua2026cooperbench} (\cref{appendix:merge-resolver}) 
before grading. 
If any yields code that passes, the scenario is considered passed. Thus, the metric reflects the best a downstream merge step could recover without further agent work. \cref{appendix:merge-policy} shows a per-policy breakdown. 

\paragraph{Metrics.}
We report two primary axes: joint success (pass rate) and cost. 
For contradictory instances we also report partial completion: the fraction of instances where agent~A's tests pass, where agent~B's pass, where both
pass, and where either passes (\cref{appendix:per-agent-performance}).
The per-side results can identify cases where each side succeeds alone but the agents fail together. 
Because true agent cost differs across models and platforms, as well as across token type (input vs. output, cache hits vs. misses, etc.), to obtain a single token cost number, we report cost based on the following formula: 
$\text{input} + \text{cache-write} + 0.1\cdot\text{cache-read} + 5\cdot\text{output}$. 
The weights approximate \claudeopus{}'s per-token rates, with output counted at $5\times$ and cache reads at $0.1\times$ fresh input, and cache-write counted the same as fresh input as a proxy (\cref{appendix:openrouter-pricing}).
We report cost relative to the oracle rather than as absolute dollars, since prices are billed differently across providers and drift over time, and since the models use different tokenizers. Harnesses also report tokens differently.
For the open-weight configurations we count only fresh input: Codex reports input including cache reads, so we subtract them, and Claude Code connected to our model server reports no cache reads at all, so all of its input counts as fresh.
We drop the cache-write term for the open-weight configurations and in every comparison across configurations, since it is not reported for every model (\cref{appendix:token-accounting}).
Costs are only compared within a configuration, against that configuration's own oracle.
The underlying raw per-scenario token counts are given in \cref{appendix:raw-token-counts}.

\paragraph{Harnesses and models.} We run seven configurations of agent harness and model.
The first is Claude Code with \claudeopus{}, with reasoning effort set to \texttt{high} and its full 1M-token context window.
The other six pair each of two harnesses, Claude Code and Codex, with each of three open-weight models, \qwen{}, \deepseek{}, and \glm{}. We serve the open-weight models with vLLM, and both harnesses connect to the same server; reasoning effort is set to the highest level the server accepts (\texttt{xhigh}) in both harnesses. We give software version and model details in \cref{appendix:details}.

\paragraph{Evaluation sets.} Claude Code with \claudeopus{} is evaluated on the 85 instances built on \swepro{} v1, and the six open-weight configurations are run on all 155 instances.
Every configuration is run under the full set of scenarios, and every comparison between scenarios is computed within one configuration on identical instances.
For comparisons across configurations, \cref{appendix:same-subset} reports all seven configurations on the 85 instances they share.

\section{Results \& Analysis}

\label{sec:results-and-analysis}

We report results for Claude Code with \claudeopus{} (\cref{table:opus-results-main}) and for the six open-weight configurations on all 155 instances (\cref{table:configs-results-main}).
We compare scenarios on the same instances using two-sided exact McNemar tests for joint success and paired Wilcoxon signed-rank tests for effective token costs.
For the open-weight configurations, we give the mean over configurations and pool the McNemar test by summing the discordant instances over configurations. We also report in how many of the six configurations a difference is significant on its own ($k$/6).
Full significance test results appear in \cref{appendix:significance}.

\subsection{Results}
\label{subsec:results}

\myparagraph{Performance drops dramatically in parallel settings.}
We first compare the single-agent oracle with parallel-unaware, where each agent receives only its own requirement and works independently.
\textbf{Joint pass rates drop significantly} for \claudeopus{}, from 62\% to 16\% on contradictory instances ($p<0.001$) and from 86\% to 57\% ($p=0.02$) on complementary instances (\cref{table:opus-results-main}; \cref{appendix:significance}).
The same drop appears in all six open-weight configurations: joint success falls from 60\% to 19\% on contradictory instances on average (significant in 6/6 configurations; pooled $p<0.001$) and from 72\% to 53\% on complementary instances (4/6, with all six in the same direction; pooled $p<0.001$; \cref{table:configs-results-main}).
This drop is also reflected when agents have access to gold tests: here, the oracle agent achieves a near-perfect pass rate of 98\% on contradictory examples.
In contrast, the parallel-unaware agent with gold tests achieves only 38\% success ($p<0.001$).
The open-weight configs show the same contrast: 96\% for the oracle with gold tests and 32\% for parallel agents with gold tests (6/6).
Thus, \textbf{access to gold tests alone does not recover the oracle's joint success} in the parallel setting, indicating that agents fail to coordinate on shared problems even when they can debug against gold tests. 

\begin{table}[t]
\centering
\small
\setlength{\tabcolsep}{6pt}
\caption{
Pass rate (\%) and relative cost of \claudeopus{} on 85 instances, for a subset of the conditions.
\emph{Cost} is mean effective tokens relative to the family's \texttt{oracle} run (\cref{sec:methods}). Among the deployable strategies the top two pass rates per family are in \textbf{bold}. 
Darker red shading indicates worse, darker blue shading indicates better.
Full per-split results and variants are in \cref{table:opus-results-full}.
}
\label{table:opus-results-main}
\resizebox{\textwidth}{!}{
\begin{tabular}{@{}l cc cc  l cc cc@{}}
\toprule
\multicolumn{5}{c}{\textit{Upper Bounds}} & \multicolumn{5}{c}{\textit{Deployable}} \\
\cmidrule(lr){1-5}\cmidrule(lr){6-10}
 & \multicolumn{2}{c}{Contradictory} & \multicolumn{2}{c}{Complementary} &
 & \multicolumn{2}{c}{Contradictory} & \multicolumn{2}{c}{Complementary} \\
\cmidrule(lr){2-3}\cmidrule(lr){4-5}\cmidrule(lr){7-8}\cmidrule(lr){9-10}
Scenario & Pass & Cost & Pass & Cost &
Scenario & Pass & Cost & Pass & Cost \\
\midrule
Oracle           & \PassCell{62} & \CostCell{1.00} & \PassCell{86} & \CostCell{1.00} & Seq.\ A$\to$B     & \PassCell[\bfseries]{54} & \CostCell{1.18} & \PassCell[\bfseries]{77} & \CostCell{1.12} \\
Orchestrator     & \PassCell{58} & \CostCell{1.58} & \PassCell{77} & \CostCell{1.18} & Reentr.\ ABAB    & \PassCell[\bfseries]{50} & \CostCell{1.88} & \PassCell[\bfseries]{80} & \CostCell{1.49} \\
Comm-Oracle      & \PassCell{56} & \CostCell{2.74} & \PassCell{83} & \CostCell{1.31} & Parallel Unaware & \PassCell{16} & \CostCell{1.17} & \PassCell{57} & \CostCell{1.14} \\
Oracle+Gold      & \PassCell{98} & \CostCell{0.74} & \PassCell{100} & \CostCell{0.54} & Parallel Aware   & \PassCell{26} & \CostCell{1.14} & \PassCell{60} & \CostCell{1.21} \\
Parallel+Gold    & \PassCell{38} & \CostCell{0.77} & n/a & n/a & Parallel Message & \PassCell{20} & \CostCell{1.42} & \PassCell{63} & \CostCell{1.43} \\
                 &    &      &     &     & Parallel Sync    & \PassCell{22} & \CostCell{1.63} & \PassCell{71} & \CostCell{1.43} \\
\bottomrule
\end{tabular}
}
\end{table}

\begin{table}[t]
\centering
\setlength{\tabcolsep}{3pt}
\caption{
Pass rate (\%) and relative cost of the six open-weight configurations on all 155 instances, for the conditions of \cref{table:opus-results-main}; \emph{Mean} averages the six configurations.
\emph{Cost} is mean effective tokens relative to the same configuration's \texttt{oracle} in the same family (\cref{sec:methods}).
Among the deployable strategies (below the rule in each family), the top two pass rates per column are in \textbf{bold} (ties shared).
Darker red shading indicates worse, darker blue shading indicates better.
All scenarios are in \cref{table:configs-results-full-contra,table:configs-results-full-comple}.
}
\label{table:configs-results-main}
\resizebox{\textwidth}{!}{
\begin{tabular}{@{}l cc cc cc cc cc cc cc@{}}
\toprule
 & \multicolumn{6}{c}{Claude Code} & \multicolumn{6}{c}{Codex} & & \\
\cmidrule(lr){2-7}\cmidrule(lr){8-13}
 & \multicolumn{2}{c}{Qwen} & \multicolumn{2}{c}{DeepSeek} & \multicolumn{2}{c}{GLM}
 & \multicolumn{2}{c}{Qwen} & \multicolumn{2}{c}{DeepSeek} & \multicolumn{2}{c}{GLM} & \multicolumn{2}{c}{Mean} \\
\cmidrule(lr){2-3}\cmidrule(lr){4-5}\cmidrule(lr){6-7}\cmidrule(lr){8-9}\cmidrule(lr){10-11}\cmidrule(lr){12-13}\cmidrule(lr){14-15}
Scenario & Pass & Cost & Pass & Cost & Pass & Cost & Pass & Cost & Pass & Cost & Pass & Cost & Pass & Cost \\
\midrule
\multicolumn{15}{l}{\textit{Contradictory} ($n=85$)} \\
Oracle & \PassCell{48} & \CostCell{1.00} & \PassCell{46} & \CostCell{1.00} & \PassCell{71} & \CostCell{1.00} & \PassCell{66} & \CostCell{1.00} & \PassCell{68} & \CostCell{1.00} & \PassCell{60} & \CostCell{1.00} & \PassCell{60} & \CostCell{1.00} \\
Orchestrator & \PassCell{34} & \CostCell{1.36} & \PassCell{36} & \CostCell{0.77} & \PassCell{53} & \CostCell{0.73} & \PassCell{49} & \CostCell{0.42} & \PassCell{60} & \CostCell{0.44} & \PassCell{44} & \CostCell{0.26} & \PassCell{46} & \CostCell{0.66} \\
Comm-Oracle & \PassCell{35} & \CostCell{1.53} & \PassCell{46} & \CostCell{1.70} & \PassCell{54} & \CostCell{1.32} & \PassCell{60} & \CostCell{1.72} & \PassCell{61} & \CostCell{1.94} & \PassCell{33} & \CostCell{1.32} & \PassCell{48} & \CostCell{1.59} \\
Oracle + Gold Tests & \PassCell{94} & \CostCell{0.64} & \PassCell{95} & \CostCell{0.58} & \PassCell{98} & \CostCell{0.65} & \PassCell{98} & \CostCell{0.71} & \PassCell{99} & \CostCell{0.57} & \PassCell{94} & \CostCell{0.53} & \PassCell{96} & \CostCell{0.61} \\
Parallel + Gold Tests & \PassCell{33} & \CostCell{0.77} & \PassCell{28} & \CostCell{0.67} & \PassCell{29} & \CostCell{0.61} & \PassCell{29} & \CostCell{0.80} & \PassCell{33} & \CostCell{0.60} & \PassCell{39} & \CostCell{0.70} & \PassCell{32} & \CostCell{0.69} \\
\cmidrule(lr){1-15}
Sequential A$\to$B & \PassCell[\bfseries]{42} & \CostCell{1.14} & \PassCell[\bfseries]{52} & \CostCell{1.11} & \PassCell[\bfseries]{66} & \CostCell{0.92} & \PassCell[\bfseries]{76} & \CostCell{1.08} & \PassCell[\bfseries]{66} & \CostCell{1.02} & \PassCell[\bfseries]{55} & \CostCell{1.07} & \PassCell[\bfseries]{60} & \CostCell{1.06} \\
Reentrant ABAB & \PassCell[\bfseries]{38} & \CostCell{2.03} & \PassCell[\bfseries]{46} & \CostCell{2.10} & \PassCell[\bfseries]{59} & \CostCell{1.40} & \PassCell[\bfseries]{66} & \CostCell{2.37} & \PassCell[\bfseries]{53} & \CostCell{2.09} & \PassCell[\bfseries]{53} & \CostCell{1.72} & \PassCell[\bfseries]{52} & \CostCell{1.95} \\
Parallel Unaware & \PassCell{15} & \CostCell{1.07} & \PassCell{12} & \CostCell{1.07} & \PassCell{16} & \CostCell{0.94} & \PassCell{20} & \CostCell{1.04} & \PassCell{26} & \CostCell{0.92} & \PassCell{25} & \CostCell{1.05} & \PassCell{19} & \CostCell{1.02} \\
Parallel Aware & \PassCell{14} & \CostCell{1.07} & \PassCell{15} & \CostCell{0.96} & \PassCell{18} & \CostCell{0.91} & \PassCell{20} & \CostCell{1.05} & \PassCell{25} & \CostCell{0.90} & \PassCell{20} & \CostCell{1.04} & \PassCell{19} & \CostCell{0.99} \\
Parallel Message & \PassCell{14} & \CostCell{1.40} & \PassCell{20} & \CostCell{1.40} & \PassCell{29} & \CostCell{1.11} & \PassCell{22} & \CostCell{1.48} & \PassCell{29} & \CostCell{1.32} & \PassCell{29} & \CostCell{1.11} & \PassCell{24} & \CostCell{1.30} \\
Parallel Sync & \PassCell{15} & \CostCell{1.57} & \PassCell{22} & \CostCell{1.67} & \PassCell{34} & \CostCell{1.28} & \PassCell{27} & \CostCell{1.64} & \PassCell{39} & \CostCell{1.74} & \PassCell{27} & \CostCell{1.22} & \PassCell{27} & \CostCell{1.52} \\
\midrule
\multicolumn{15}{l}{\textit{Complementary} ($n=70$)} \\
Oracle & \PassCell{61} & \CostCell{1.00} & \PassCell{70} & \CostCell{1.00} & \PassCell{74} & \CostCell{1.00} & \PassCell{84} & \CostCell{1.00} & \PassCell{74} & \CostCell{1.00} & \PassCell{66} & \CostCell{1.00} & \PassCell{72} & \CostCell{1.00} \\
Orchestrator & \PassCell{51} & \CostCell{0.94} & \PassCell{51} & \CostCell{0.68} & \PassCell{64} & \CostCell{0.62} & \PassCell{66} & \CostCell{0.40} & \PassCell{60} & \CostCell{0.37} & \PassCell{60} & \CostCell{0.26} & \PassCell{59} & \CostCell{0.55} \\
Comm-Oracle & \PassCell{71} & \CostCell{1.12} & \PassCell{69} & \CostCell{1.25} & \PassCell{71} & \CostCell{1.07} & \PassCell{80} & \CostCell{1.41} & \PassCell{77} & \CostCell{1.31} & \PassCell{60} & \CostCell{1.29} & \PassCell{71} & \CostCell{1.24} \\
Oracle + Gold Tests & \PassCell{94} & \CostCell{0.74} & \PassCell{96} & \CostCell{0.71} & \PassCell{94} & \CostCell{0.76} & \PassCell{100} & \CostCell{0.86} & \PassCell{97} & \CostCell{0.67} & \PassCell{99} & \CostCell{0.66} & \PassCell{97} & \CostCell{0.73} \\
Parallel + Gold Tests & n/a & n/a & n/a & n/a & n/a & n/a & n/a & n/a & n/a & n/a & n/a & n/a & n/a & n/a \\
\cmidrule(lr){1-15}
Sequential A$\to$B & \PassCell[\bfseries]{63} & \CostCell{0.98} & \PassCell[\bfseries]{73} & \CostCell{0.89} & \PassCell[\bfseries]{71} & \CostCell{0.82} & \PassCell[\bfseries]{81} & \CostCell{1.18} & \PassCell[\bfseries]{71} & \CostCell{0.92} & \PassCell[\bfseries]{66} & \CostCell{1.00} & \PassCell[\bfseries]{71} & \CostCell{0.96} \\
Reentrant ABAB & \PassCell[\bfseries]{66} & \CostCell{1.45} & \PassCell{67} & \CostCell{1.74} & \PassCell{66} & \CostCell{1.40} & \PassCell[\bfseries]{79} & \CostCell{2.32} & \PassCell{70} & \CostCell{1.82} & \PassCell[\bfseries]{71} & \CostCell{1.63} & \PassCell[\bfseries]{70} & \CostCell{1.73} \\
Parallel Unaware & \PassCell{49} & \CostCell{1.03} & \PassCell{57} & \CostCell{1.10} & \PassCell{56} & \CostCell{0.94} & \PassCell{56} & \CostCell{1.17} & \PassCell{50} & \CostCell{1.00} & \PassCell{50} & \CostCell{1.07} & \PassCell{53} & \CostCell{1.05} \\
Parallel Aware & \PassCell{46} & \CostCell{0.97} & \PassCell{59} & \CostCell{0.96} & \PassCell{66} & \CostCell{0.87} & \PassCell{69} & \CostCell{1.16} & \PassCell{63} & \CostCell{1.01} & \PassCell{47} & \CostCell{1.24} & \PassCell{58} & \CostCell{1.04} \\
Parallel Message & \PassCell{59} & \CostCell{1.21} & \PassCell{70} & \CostCell{1.26} & \PassCell[\bfseries]{69} & \CostCell{0.98} & \PassCell{70} & \CostCell{1.38} & \PassCell[\bfseries]{73} & \CostCell{1.19} & \PassCell{54} & \CostCell{1.08} & \PassCell{66} & \CostCell{1.18} \\
Parallel Sync & \PassCell{57} & \CostCell{1.37} & \PassCell[\bfseries]{71} & \CostCell{1.29} & \PassCell[\bfseries]{71} & \CostCell{0.94} & \PassCell{69} & \CostCell{1.44} & \PassCell{69} & \CostCell{1.29} & \PassCell{47} & \CostCell{1.48} & \PassCell{64} & \CostCell{1.30} \\
\bottomrule
\end{tabular}
}
\end{table}

\myparagraph{Sequential baselines recover performance but do not scale.}
To assess sequential access to the evolving codebase, we compare parallel-unaware with a single A$\to$B hand-off, in this case without access to edit history.
On \claudeopus{}, joint success rises significantly from 16\% to 54\% on contradictory instances ($p<0.001$).
It also increases from 57\% to 77\% on complementary instances, though this is non-significant ($p=0.065$).
The mean effective token cost remains similar: 1.17 to 1.18 times the oracle for contradictory instances, and 1.14 to 1.12 times for complementary instances (\cref{table:opus-results-main}).
Therefore, \textbf{a single hand-off recovers much of the observed performance gap at comparable token cost}.
We further extend this in the reentrant scenario (ABAB) that allows both agents a second turn.
This has no significant effect on the pass rate, achieving 50\% on contradictory ($4\%$ drop) and 80\% ($3\%$ gain),
while raising costs significantly to 1.88 and 1.49 times the oracle.
The open-weight configurations show the same pattern more strongly.
A single hand-off raises joint success from 19\% to 60\% on contradictory and from 53\% to 71\% on complementary instances (6/6 in both families) at a similar cost (1.06 and 0.96 times the oracle), and on contradictory instances it matches the oracle (60\%).
Reentrance again costs significantly more than the single hand-off (1.95 and 1.73 times the oracle; 6/6 in both families) without raising joint success (52\% and 70\%).
In \cref{appendix:full-result-tables} we show similar results for sequential and reentrant baselines when the orders are reversed, as well as when we add history.
For the open-weight configurations, reversing the order lowers contradictory joint success (B$\to$A 46\% vs.\ A$\to$B 60\%; 4/6), and adding history lowers it slightly (53\%; 2/6).
Based on these results, we conclude that \textbf{reentrance costs significantly more and provides no discernible performance gains.}
Moreover, \textit{enforcing sequential or reentrant turn-taking is impractical with more agents}; if hundreds or thousands of developers independently deploy agents on a shared codebase -- as might be expected in large organizations and teams using monorepos -- sequential turn-taking becomes completely infeasible.

\myparagraph{Peer awareness and access to communication marginally improve performance but fail to close the gap.}
We compare parallel agents with peer awareness, messaging, and synchronization. Peer awareness slightly raises joint success from 16\% to 26\% on contradictory instances and from 57\% to 60\% on complementary instances, but neither gain is significant.
Messaging and synchronization also provide limited improvements, reaching 20\%/63\% (contradictory/complementary) and 22\%/71\%, respectively (\cref{table:opus-results-main}).
For \claudeopus{}, Redis Sync reaches 32\%/77\%, the highest success rates among the messaging and synchronization variants, and is significantly above parallel-unaware on contradictory instances ($p = 0.04$); its advantage over file-based Sync (22\%/71\%) is not significant ($p = 0.13$ and $p = 0.63$; \cref{appendix:significance}). It also comes at a greater effective token cost of 2.23/1.39 times the single-agent oracle (\cref{appendix:full-result-tables}), and still trails the single-agent oracle by 30 and 9 percentage points.
Note that all four messaging and synchronization variants use more effective tokens than parallel-unaware.
The communication oracle reaches 56\%/83\%, exceeding Redis Sync by 24 and 6 percentage points (\cref{appendix:full-result-tables}).
The open-weight configurations confirm that awareness alone does not help (19\% vs.\ 19\% on contradictory and 58\% vs.\ 53\% on complementary instances; significant in 0/6 and 2/6 configurations), and that communication gives small but consistent gains at a higher cost: messaging and synchronization reach 24\%/66\% and 27\%/64\%, above parallel-aware in most configurations (pooled $p<0.01$ on both families), and on contradictory instances both cost significantly more than parallel-aware in every configuration.
As for \claudeopus{}, Redis is not significantly better than the shared file (Redis Sync 28\%/64\% vs.\ Sync 27\%/64\%).
The communication oracle again exceeds the best communication setting on contradictory instances (48\% vs.\ 28\% for Redis Sync) while remaining below the single-agent oracle (60\%).
The fact that a gap remains to the single-agent oracle performance even when the agent is exactly instructed on what to communicate about and how suggests that providing a communication channel is insufficient,
particularly for contradictory tasks.
We conclude that \textbf{communication improves joint success at a higher token cost, but leaves a substantial gap to the communication oracle on contradictory tasks.}
We further analyze the contents of the communication channel in \cref{subsec:communication_analysis}.

\myparagraph{Post-hoc reconciliation recovers only a small part of the gap.}
To separate joint success from the choice of combination rule, \cref{appendix:merge-policy} evaluates concatenated diffs, Git merging, and a model-based reconciler taken from \citet{khatua2026cooperbench} on the integration of code changes from parallel agents.
For parallel-unaware, contradictory success is 10\% (concatenation), 14\% (merging), and 16\% (model-based), while complementary success is 51\%, 54\%, and 54\%.
The reconciler's gain is larger in the Redis sync setting, where it
raises contradictory success from 22\% to 32\%, and complementary success from 69\% to 77\%.
For \claudeopus{}, none of these differences is significant (\cref{appendix:merge-policy}).
For the open-weight configurations, the model-based reconciler does help: on contradictory instances, parallel-unaware success rises from 9\% with Git merging to 19\% with reconciliation, and the reconciler's gain over Git merging is significant in at least four of the six configurations for every contradictory parallel scenario (\cref{table:configs-merge-policy}).
Even so, across all four non-oracle parallel conditions shown in \cref{table:opus-results-main} (and indeed, across all six such conditions in the full results in \cref{appendix:full-result-tables}),
the reconciler remains below the sequential A$\to$B baseline's 54\% contradictory success for \claudeopus{}; for the open-weight configurations, the best parallel scenario with reconciliation reaches 28\%, against 60\% for a single hand-off.
In other words, \textbf{changing the combination rule recovers some failures, but does not resolve the coordination failures observed in the parallel setting.}

\myparagraph{The parallel gap holds across harnesses and models.}
\cref{table:configs-results-main} reports the six open-weight configurations on all 155 instances, and \cref{appendix:same-subset} compares all seven configurations on the 85 instances they share.
On the shared instances, the open-weight configurations reach single-agent oracle rates comparable to \claudeopus{} (72\% vs.\ 62\% on contradictory and 78\% vs.\ 86\% on complementary instances, averaged over the six), and every configuration shows a large oracle-to-parallel gap on contradictory instances (parallel-unaware 16--28\%).
In every configuration, a single hand-off recovers most or all of the gap, awareness does not significantly help on contradictory instances, and communication raises cost while recovering only part of the gap.
The level of performance varies across configurations (e.g., contradictory oracle 46--71\% on all 155 instances), as does the size of individual effects, such as the gain from the model-based reconciler (\cref{appendix:merge-policy}).
Thus, \textbf{all configurations have a parallel performance gap and a higher communication cost}.

\subsection{Analysis}
\label{sec:analysis}

\myparagraph{Error analysis and failure modes.}
\label{subsec:failure_modes}
We manually annotate every failing \claudeopus{} run in the oracle and parallel scenarios using open coding: rather than starting from a fixed taxonomy, we annotate each failure with a short descriptive code grounded in its produced patches, the gold solutions, and the message logs, and then group the codes into a small set of failure classes.
The isolated per-agent test results let us label each contradictory failure objectively as either a \emph{competence} failure (an agent's patch fails its own tests) or an \emph{integration} failure (both patches pass alone, but none of their combinations passes). We find that \textbf{single-agent failures are mostly competence failures:} 92\% of oracle failures involve the oracle producing an implementation that diverges from the test requirements (e.g.,  logic placed in the wrong function or layer, wrong API signatures, return values, or configurations). On the other hand, \textbf{parallel failures are mostly integration failures:} across parallel conditions, in 58\% of contradictory failures each agent's half passes its own tests but no combination of the two patches passes the joint tests, 
and 98\% of those produce a git merge conflict. 
The dominant complementary mechanism is different: 53\% of complementary failures merge successfully but fail because the two halves diverge on a name,  signature, or data shape. 

Furthermore, \textbf{sequential agents often overwrite their predecessor's work:}
we track added lines by file and exact text with \textit{survival}, \textit{overwrite}, and \textit{redo} metrics defined in \cref{appendix:agent-topologies}.
On contradictory tasks, 79\% of the first agent's lines survive and 23\% of handoffs are overwrites.
On complementary tasks, these rates are 95\% and 5\%.
In sequential runs with overwrites, the first agent's requirement passes in 37\% of runs compared with 73\% without overwrites.
The second agent's pass rate stays similar, while joint success is 26\% versus 55\%.
Access to history results in \textit{last-mover advantage}, which means the same agent on the same instance passes more often when it moves last with history.
Surprisingly, \textbf{history actually increases overwriting and redos:}
with history, overwrites rise from 15\% to 30\% of handoffs and redos from 15\% to 39\% of reentrant runs (both $p<0.001$).
Transcripts suggest that agents often compare attributed edits with their own requirement and replace them as incorrect attempts.
Without history, agents sometimes extend the same code as existing behavior.
History also makes earlier edits easy to restore because agents use \texttt{git checkout} or \texttt{git revert} in 31\% of redos with history, compared with 7\% without.
Finally, \textbf{communication does not prevent conflicting edits and sometimes causes them:} even when the agents agree on a correct joint design in the channel, the run still fails due to changing text style (variable names, quote style, import order), both agents delegating merging to each other, or avoiding the conflicting code entirely.
This finding extends the findings of \citet{khatua2026cooperbench} that agents deviate from commitments even after effective communication. In our setting the agents often attempt to keep their commitments and still fail.

\myparagraph{Communication analysis.}
\label{subsec:communication_analysis}
We further analyze the message logs and transcripts of all valid \claudeopus{} channel runs, track when each message is read relative to the recipient's first edit, and label the conversation content with an outcome-blind LLM judge (Claude Sonnet 5) validated against human labels; full details are in \cref{appendix:communication_analysis}. 
In summary, agents know \emph{what} to communicate but largely fail in terms of \emph{who} should act and \emph{how} to act on information in a timely fashion.
First, \textbf{agents send early but read late.} Communication is two-way in at least 94\% of runs and most agents send messages before editing, yet about $1/3$ of messages are never read, and only 21--23\% are read before the recipient's first edit, because agents check for messages mainly at the start and end of a turn. This gels with the findings of \citet{khatua2026cooperbench}. 
However, unlike \citet{khatua2026cooperbench}, who found that agents often polluted the channel ``with vague, ill-timed, and inaccurate messages'', we find that the agents we tested \textbf{discuss the right thing}. 
88\% of contradictory and 76\% of complementary conversations address the annotated shared artifact, and 59\% and 74\% respectively reach a concrete agreement on names, signatures, or edit ownership. 
Finally, \textbf{knowing the partner's task leads to subtask ownership.} In the communication oracle, 85\% of contradictory conversations assign an owner to the contested code, conflicts fall from 67\% to 18\% of runs, and runs that assign an owner pass at a higher rate of 61\% versus 29\%. 
Interestingly, in 15\% of conversations, one agent implements \emph{both} requirements while the other withdraws. Note that this is an unscalable strategy that defeats the purpose of multi-agent collaboration, since with $n$ agents it is completely infeasible for one agent to do all the work on a company scale.

\section{Related Work}
\label{sec:related_work}

\myparagraph{Coordination in Classical Multi-agent Systems.}
Dependencies, conflicts, and communication costs are longstanding concerns in multi-agent systems, where prior work has studied the selective exchange of local plans \citep{durfee1988using}, beneficial and conflicting plan interactions \citep{von1992coordinating}, and trade-offs between communication and miscoordination costs \citep{tambe1997towards}.
We study similar problems in a complex, real-world setting: \textit{decentralized software development}, where individual agents must implement natural language requirements in a shared codebase.

\myparagraph{Coding Agents and Multi-agent Development.} 
Coding agent harnesses (e.g., Claude Code \citep{anthropic2026claudecode}, Codex \citep{jif2026openai}, SWE-Agent \citep{yang2024swe}, OpenHands \citep{wang2025openhands}, etc.) equip LLMs with tools to navigate repositories, edit files, and execute code.
We study how multiple agents operating in such harnesses coordinate on shared code. 
Multi-LLM systems such as MetaGPT \citep{hong2024metagpt}, ChatDev \citep{qian2024chatdev}, MAGIS \citep{ tao2024magis}, and AgileCoder \citep{nguyen2025agile} organize development work through predefined roles or structured dialogue, coordinating around a single shared development request. 
CAID \citep{geng2026effective} uses a central manager to plan dependent subtasks and runs isolated workers asynchronously. 
Our setting fundamentally differs: 
in \dataset{}, multiple agents receive \textit{separate} assignments without any central planner.
This distinction is key: 
in centralized, decomposition-based methods 
coordination can be partly addressed through the decomposition itself \citep{hong2024metagpt, geng2026effective}. 
In contrast, we study a setting where requirements are created \textit{separately} by multiple principals, with independence enforced during dataset construction (\cref{appendix:dataset-creation-details}).   
Agents must discover and resolve dependencies or conflicts that are not apparent from either requirement. 

\myparagraph{Benchmarks for Agent Coordination.} 
The Collaboration Gap \citep{davidson2025collaboration} studies collaborative maze solving and shows that strong solo performance does not guarantee successful teamwork. 
CooperBench \citep{khatua2026cooperbench} is closest to our setting: agents operate and communicate to implement separate features on a shared repository, potentially coordinating via messages, and are scored based on a merged result. 
However, CooperBench focuses on a collaborative setting in which agents implementing jointly compatible features are explicitly framed as teammates, are aware of each other, and are told to communicate.
We are instead motivated natively by a large-scale coding scenario in which agents
implementing independently-motivated requests may be mutually unaware, and focus on independent tasks that conflict at a behavioral level. 
\citet{xia2026passesalone} studies a setting similar to our parallel-unaware setting, in which two agents work in parallel on Django tasks without seeing each other's changes, and find that interference between agents occurs in the majority of tasks.
\dataset{} encompasses this setting as well as  agents that communicate \emph{while} working in parallel, with requirements from independent principals that may be contradictory or complementary, across 11 repositories.

\myparagraph{Communication between Agents.}
DeLM uses a shared context to help coding agents reuse discoveries across attempts to resolve the same software issue \citep{mao2026decentralized}.
{Agent-Radio} supports asynchronous communication among agents investigating different parts of a code-comprehension question, with source modification prohibited in its evaluation \citep{ren2026agentradio}.
Our study examines whether communication helps agents coordinate \textit{conflicting} implementation decisions across \textit{independently specified requirements}.
This problem setting requires agents to agree on shared interfaces or reconcile conflicting behavioral requirements, and we analyze the tradeoff between integration correctness and execution cost.

\myparagraph{Multi-agent Benefits.}
\citet{kim2026capable} evaluate 260 configurations of different language models and coordination architectures across six benchmarks and find that collaboration benefits depend on task structure and can diminish as individual agents become more capable. 
A complementary analysis by \citet{cemri2026multi} identifies 14 failure modes in multi-agent LLM systems involving system design, inter-agent misalignment, and task verification.
These studies ask when collaboration improves performance on a shared task, while our study examines a different coordination setting when separate coding assignments are distributed across multiple agents on the same codebase and result in conflicts.

\section{Conclusion}
As agents take on work delegated by many programmers who share a codebase but not   a plan, they will increasingly face coordination challenges. 
We introduced \dataset{} to study this decentralized multi-agent coordination setting in its most targeted form: two agents with independent requirements that must agree on a shared interface (complementary) or reconcile incompatible edits (contradictory), scored on whether their combined code satisfies both.

We find that much of the difficulty lies in coordination: across all seven configurations of agent harness and model, a single agent with both requirements succeeds more often than parallel agents. Peer awareness and access to communication channels only yield marginal gains, and post-hoc reconciliation recovers only a small part of the gap.
Sequential topologies or explicit ownership assignment through communication recover much of the performance gap, but are not feasible in large-scale development settings.
Finally, in our analysis of the Claude Code harness with \claudeopus{}, agents often discuss the right topics, but fail in terms of how to communicate and how to act on communication.

\section*{Limitations and Future Work}

\paragraph{Limitations.}
Our evaluation focuses on two agents and a small dataset of coding tasks. 
This setting helps identify coordination failures but does not establish how the findings generalize to larger teams or longer development workflows. 
\dataset{} is also diagnostic rather than a prevalence estimate: it is a targeted benchmark with instances that have been selected to require coordination.
This allows us to measure how well agents handle such conflicts when they arise, but does not predict how often they arise in practice.
Like prior work \citep{deng2025swe, mao2026decentralized, khatua2026cooperbench}, we focus on coding because executable tests allow us to check joint correctness. 
This leaves open many multi-agent tasks, such as collaborative writing and open-ended research, which lack executable tests that reliably measure joint correctness.
\paragraph{Future work.}
Future work should extend \dataset{} beyond two-agent pairs and design robust coordination strategies in $n$-agent scenarios. 
\dataset{} also provides a training signal for agents to learn to negotiate interfaces and edit ownership rather than being prompted to.

\section*{Acknowledgments}
Elias Stengel-Eskin is supported by Schmidt Sciences and Coefficient Giving.
A portion of this work has taken place in the Learning Agents Research Group (LARG) at the Artificial Intelligence Laboratory, The University of Texas at Austin. LARG research is supported in part by the National Science Foundation (FAIN-2019844, NRT-2125858, OIC-2535195), the Office of Naval Research (N00014-24-1-2550), Army Research Office (W911NF-17-2-0181, W911NF-23-2-0004, W911NF-25-1-0065), Lockheed Martin, Lyda Hill Philanthropies, and Good Systems, a research grand challenge at the University of Texas at Austin. The views and conclusions contained in this document are those of the authors alone. Peter Stone serves as the Chief Scientist of Sony Corporate Technology Center and receives financial compensation for that role. The terms of this arrangement have been reviewed and approved by the University of Texas at Austin in accordance with its policy on objectivity in research.
The authors acknowledge the Texas Advanced
Computing Center (TACC) at The University of Texas at Austin for providing computational resources that have contributed to the research results reported within this paper.

\bibliography{iclr2027_conference}
\bibliographystyle{iclr2027_conference}
\clearpage 
\appendix

\section{Dataset Composition}
\label{appendix:dataset-composition}

\cref{table:dataset-stats} shows the composition of \dataset{} by repository and family, split into seed instances and generated instances, and by the \swepro{} release each instance is built on (\cref{appendix:swepro-versions}).

\begin{table}[htp]
\centering
\small
\setlength{\tabcolsep}{5pt}
\caption{Dataset composition. Instance counts per repository and family, split into manually constructed \emph{Seed} instances and \emph{Generated} instances produced by \gpt{} using seed examples for guidance, and by the \swepro{} release (v1 or v2) the instance is built on.
Instances are produced by adding complementary or contradictory tasks to \swepro{} instances; thus, they reflect coding problems from real-world repositories.
\dataset{} comprises 155 instances (85 contradictory $+$ 70 complementary; 85 built on v1 $+$ 70 on v2), drawn from 11 open-source repositories across four languages.}
\label{table:dataset-stats}
\begin{tabular}{l l cccc | cccc}
\toprule
 & & \multicolumn{4}{c}{Contradictory} & \multicolumn{4}{c}{Complementary} \\
\cmidrule(lr){3-6}\cmidrule(lr){7-10}
 & & \multicolumn{2}{c}{v1} & v2 & & \multicolumn{2}{c}{v1} & v2 & \\
\cmidrule(lr){3-4}\cmidrule(lr){5-5}\cmidrule(lr){7-8}\cmidrule(lr){9-9}
Repo & Language & Seed & Gen. & Gen. & Total & Seed & Gen. & Gen. & Total \\
\midrule
\href{https://github.com/flipt-io/flipt}{\textcolor{blue}{flipt}} & Go & 2 & 2 & 4 & 8 & 1 & 1 & 4 & 6 \\
\href{https://github.com/future-architect/vuls}{\textcolor{blue}{vuls}} & Go & 1 & 4 & 4 & 9 & 1 & 3 & 4 & 8 \\
\href{https://github.com/gravitational/teleport}{\textcolor{blue}{teleport}} & Go & 0 & 1 & 3 & 4 & 1 & 1 & 4 & 6 \\
\href{https://github.com/navidrome/navidrome}{\textcolor{blue}{navidrome}} & Go & 0 & 6 & 3 & 9 & 3 & 4 & 4 & 11 \\
\midrule
\href{https://github.com/ansible/ansible}{\textcolor{blue}{ansible}} & Python & 2 & 7 & 4 & 13 & 2 & 1 & 5 & 8 \\
\href{https://github.com/internetarchive/openlibrary}{\textcolor{blue}{openlibrary}} & Python & 2 & 1 & 3 & 6 & 2 & 2 & 3 & 7 \\
\href{https://github.com/qutebrowser/qutebrowser}{\textcolor{blue}{qutebrowser}} & Python & 2 & 4 & 3 & 9 & 1 & 1 & 3 & 5 \\
\midrule
\href{https://github.com/element-hq/element-web}{\textcolor{blue}{element-web}} & JS/TS & 1 & 2 & 4 & 7 & 2 & 3 & 4 & 9 \\
\href{https://github.com/NodeBB/NodeBB}{\textcolor{blue}{NodeBB}} & JS/TS & 2 & 6 & 4 & 12 & 2 & 3 & 3 & 8 \\
\href{https://github.com/protonmail/webclients}{\textcolor{blue}{webclients}} & JS/TS & 0 & 5 & 2 & 7 & 0 & 1 & 1 & 2 \\
\href{https://github.com/tutao/tutanota}{\textcolor{blue}{tutanota}} & JS/TS & 0 & 0 & 1 & 1 & 0 & 0 & 0 & 0 \\
\midrule
Total &  & 12 & 38 & 35 & 85 & 15 & 20 & 35 & 70 \\
\bottomrule
\end{tabular}
\end{table}

\section{Dataset Generation Details}
\label{appendix:dataset-creation-details}

This section of the appendix gives details on how candidate tasks are selected, and how the scout, author, and auditor stages operate.

\textbf{Candidate selection and diversity by construction.}
Diversity is enforced before any model inspects the code, so the candidate
set is spread by construction. Starting from
the pool of \swepro{} tasks, each task is labeled from  row
metadata alone: its language, its repository, and the subsystem
it edits (the dominant changed source directory in the task's gold patch). A deterministic, stratified selector then fills a fixed number of
candidates by balancing first on language, then on repository (with a floor so every
repository, including ones untouched by the seeds, is represented), then round-robin
over distinct subsystems, capped at two candidates per
(repository,~subsystem) pair. The selection uses a fixed random seed and a stable
sort so it is reproducible, and it prefers subsystems not already used by an existing
seed to maximize cross-batch novelty.

\textbf{Suggestion stage.}
On this pre-spread candidate set, a scout model (Qwen3.8 27B; \cref{appendix:qwen-configuration}) explores each task's repository and proposes which family the task best
supports and how strong the conflict or interface signal is, under the
independent-task-pair framing (requirement A the original task, requirement B a separately motivated counterpart). The scout's judgment is an explicit,
non-authoritative suggestion, and is used only to rank and route candidates
into the authoring queue.

\textbf{Authoring.}
For each selected candidate, \gpt{} (in Codex) authors requirement B together with its reference solution, its tests, and the family-specific artifacts (the combined
test set and reconciliation for contradictory instances; the integration test for complementary ones). The author is shown two in-context examples drawn from the seeds of the same family, chosen by a uniform similarity score (same
repository, then same language, then same code area) so the harness, test-id, and formatting conventions match the repository being authored. The authored instance is
then validated (every diff must apply and every schema field must be present) before verification.

\textbf{Verification and correction.}
Verification runs inside the original \swepro{} execution environment: each
source task ships a prebuilt per-task Docker image, and we build and test every
instance in that same image (inherited unchanged from the source task) so the
runtime matches the upstream benchmark rather than a reconstructed one. Each
authored instance is verified against the soundness conditions of
\cref{sec: benchmark_construction_and_experimental_settings} by applying every relevant combination of changes in this
container and running the instance's tests. An instance that fails any check is
handed the failing check's diagnostic and re-authored, up to two correction rounds;
instances that never become sound are discarded.

\textbf{Independence and blind audit.}
Soundness alone does not guarantee that the two requirements read as
independent tickets rather than two halves of one decomposed task. A separate
auditing pass with \gpt{} therefore checks each sound instance against the
independence criterion and the blind rule (no problem statement may name the other
requirement, its agent, its files, or its tests), and instances judged to be
decomposition or to leak are rejected. Every instance that passes the model audit is
additionally reviewed by hand.

\textbf{Cross-set diversity audit.}
Diversity is a property of the set, not of any single instance, so in addition to
diversity-by-construction at selection we periodically run a cross-set audit over the
finalized instances of each family. It labels each instance by conflict archetype or
capability type and by code layer, computes the resulting distributions, and flags
over-concentration (a repeated (repository,~subsystem) pair, a repository or an
archetype that dominates the set) together with the buckets to avoid next. This audit
is advisory and changes no finalized instance.

The prompt for each of these stages can be found in our code repository under \texttt{prompts/} folder.

\section{SWE-Bench Pro v1 and v2}
\label{appendix:swepro-versions}

\swepro{} was revised after we had built our first set of instances.
We call the original release v1 and the revision v2.
The revision rewrote parts of many tasks' problem statements, tests, and execution environments, and removed some tasks.
Our 85 v1 instances are built on v1 tasks, and we built 70 further instances on v2 tasks.
No \swepro{} task is used in both sets. Because five tasks in each release are the source of both a contradictory and a complementary instance, the 85 v1 instances come from 80 distinct tasks and the 70 v2 instances from 65. We treat the two sets as a single dataset of 155 instances in the main sections of this paper.

\textbf{Why the v1 instances remain correct.}
Each instance is self-contained: it fixes a repository and base commit, and carries its own requirements, reference solutions, tests, and execution environment (the source task's container image, run script, and test parser).
It does not use the upstream task's problem statement or tests.
A revision of \swepro{} can therefore affect an instance only through its base commit or its execution environment, and every instance was verified sound at creation (\cref{sec: benchmark_construction_and_experimental_settings}).
We nonetheless checked the 80 source tasks of the v1 instances against v2:
\begin{enumerate}[leftmargin=1.5em, itemsep=1pt]
\item \emph{Availability.} All 80 v1 images remain available. 71 of the 80 tasks are still in v2, and the other 9 were removed upstream.
\item \emph{Base commits.} Identical in v1 and v2 for all 71 tasks present in both.
\item \emph{Environment.} Comparing the installed packages of the v1 and v2 images, system, Python, and Go packages are unchanged for all 71 tasks, and Node packages for 69. The other two (both \texttt{openlibrary}) have an npm lockfile regenerated to older pinned versions.
\item \emph{Behavior.} Running each task's own gold patch and tests, 75 of the 80 tasks reproduce under v1. The other five fail in the same way under v1 and v2, for environment reasons unrelated to the revision. For the 71 tasks in both releases, the v1 gold patch receives the same verdict under the v1 and v2 tests for 68. For the other three (all \texttt{flipt}), v2 rewrote the gold patch and added tests that the v1 gold patch does not pass. Our instances use their own tests which their reference solutions pass.
\item \emph{Text and tests.} Upstream, 3 of the 71 tasks changed their gold patch, 28 changed their tests but not the gold patch, 37 changed only the text of the problem statement or requirements, and 3 changed nothing material. An LLM-assisted review of the most-changed tasks, read against our instances' requirements, tests, and reference solutions, found rewordings, scope changes in behavior our instances do not test, and clarifications consistent with our instances, but no change that contradicts what an instance asserts.
\end{enumerate}

\textbf{The v2 instances.}
We built the 70 v2 instances with the pipeline of \cref{appendix:dataset-creation-details}, applied to v2 tasks: 200 candidates were selected from v2, routed by a scout model (\cref{appendix:qwen-configuration}), and authored by \gpt{} in Codex with the 27 seed instances as in-context examples.
Each v2 instance is verified in its source task's v2 image with that task's own v2 run script and parser.
All 70 are generated instances (35 contradictory and 35 complementary) from 11 repositories, one of which (\texttt{tutanota}) does not appear among the v1 instances.

\textbf{Results by release.}
\cref{table:v1-v2} splits the open-weight results by release.
The v2 instances are harder on average (e.g., contradictory oracle 72\% on v1 and 43\% on v2), but the pattern across scenarios is the same in both.
Claude Code with \claudeopus{} was run on the v1 instances only.

\begin{table}[htp]
\centering
\small
\setlength{\tabcolsep}{5pt}
\caption{Pass rate (\%) of the six open-weight configurations by \swepro{} release, averaged over configurations.}
\label{table:v1-v2}
\begin{tabular}{l rrr rrr}
\toprule
 & \multicolumn{3}{c}{Contradictory} & \multicolumn{3}{c}{Complementary} \\
\cmidrule(lr){2-4}\cmidrule(lr){5-7}
Scenario & v1 & v2 & All & v1 & v2 & All \\
\midrule
Oracle & \PassCell{72} & \PassCell{43} & \PassCell{60} & \PassCell{78} & \PassCell{65} & \PassCell{72} \\
Oracle + Gold Tests & \PassCell{96} & \PassCell{96} & \PassCell{96} & \PassCell{97} & \PassCell{96} & \PassCell{97} \\
Sequential A$\to$B & \PassCell{69} & \PassCell{46} & \PassCell{60} & \PassCell{80} & \PassCell{62} & \PassCell{71} \\
Parallel Unaware & \PassCell{22} & \PassCell{15} & \PassCell{19} & \PassCell{58} & \PassCell{48} & \PassCell{53} \\
Parallel Sync & \PassCell{34} & \PassCell{18} & \PassCell{27} & \PassCell{72} & \PassCell{56} & \PassCell{64} \\
Comm-Oracle & \PassCell{59} & \PassCell{32} & \PassCell{48} & \PassCell{82} & \PassCell{61} & \PassCell{71} \\
\bottomrule
\end{tabular}
\end{table}

\section{Example Instances}
\label{appendix:example-instances}


\lstdefinestyle{parallactexample}{
  basicstyle=\ttfamily\scriptsize,
  columns=fullflexible,
  keepspaces=true,
  breaklines=true,
  breakatwhitespace=false,
  postbreak=\mbox{\textcolor{gray}{$\hookrightarrow$}\space},
  frame=single,
  rulecolor=\color{gray!50},
  framesep=3pt,
  xleftmargin=4pt,
  xrightmargin=4pt,
  showstringspaces=false,
  upquote=false,
  tabsize=4,
  aboveskip=4pt,
  belowskip=8pt,
}

This section walks through one contradictory and one complementary instance from \dataset{}, giving each instance's requirements, reference solutions, and tests (\cref{sec:problem_statement}).
Both are seed instances built on \swepro{} v1 (\cref{appendix:dataset-creation-details}), and both pass the soundness checks of \cref{fig:benchmark-filtering}.
Code is shown as excerpts of the reference patches; lines marked \texttt{[...]} are abridged.
Outside the oracle settings, each agent receives only its own requirement; the reference solutions and tests are withheld from the agents except in the gold-test scenarios.

\subsection{Contradictory example: identifier validation in Ansible}
\label{appendix:example-contradictory}

\begin{center}
\small
\begin{tabular}{@{}ll@{}}
\toprule
Repository & \texttt{ansible/ansible} (Python) \\
\swepro{} task & \texttt{ansible\_\_ansible-1ee70fc} (v1) \\
Shared code & \texttt{isidentifier()} in \texttt{lib/ansible/utils/vars.py} \\
Scored tests & A's and B's tests together (combined test set) \\
\bottomrule
\end{tabular}
\end{center}

\textbf{Requirement A} (the original \swepro{} issue, abridged).
\begin{quote}\small
\emph{Inconsistent Python identifier validation behavior between Python 2 and Python 3 in \texttt{ansible.utils.vars.isidentifier}.}
The \texttt{isidentifier} function in \texttt{ansible.utils.vars} presents inconsistent behavior between Python 2 and Python 3 for identifier validation. Specifically, Python 2 does not consider \texttt{True}, \texttt{False}, and \texttt{None} as reserved keywords, while Python 3 does. Additionally, Python 3 allows non-ASCII characters in identifiers, while Python 2 does not. [...]
Expected results: non-ASCII characters are not allowed as valid identifiers; \texttt{True}, \texttt{False}, \texttt{None} are treated as reserved keywords in both versions.
\end{quote}

\textbf{Requirement B} (written for \dataset{}, verbatim).
\begin{quote}\small
Ansible variable validation should optionally accept dotted variable paths. Modify \texttt{ansible.utils.vars.isidentifier} to take a new keyword argument \texttt{allow\_dotted} (default \texttt{False}). When \texttt{allow\_dotted=True} and the candidate contains a dot, validate it as a dotted path: split on `.' and treat it as valid only if it is non-empty and EVERY dot-separated segment is itself a valid identifier (e.g.\ \texttt{isidentifier('foo.bar', allow\_dotted=True)} is True, \texttt{isidentifier('foo.1bad', allow\_dotted=True)} is False, \texttt{isidentifier('.foo', allow\_dotted=True)} is False). With \texttt{allow\_dotted=False} (the default) a dotted string is not a valid identifier, so existing callers are unaffected. The public entry point remains exactly \texttt{ansible.utils.vars.isidentifier}.
\end{quote}

\textbf{Why the requirements conflict.}
Both requirements change the same existing function.
A replaces its body, which validates a name by parsing it with Python's \texttt{ast} module, with ASCII-only, keyword-aware rules, and keeps the one-argument signature; B adds a parameter to the signature and a dotted-path branch on top of the old body.
The two reference patches edit the same lines of the function, and neither concatenating nor git-merging them passes the combined test set, so a joint solution has to combine both changes by hand.

\textbf{Reference solution A} (excerpt; the patch also replaces \texttt{import ast} with \texttt{import keyword} and adds a docstring).
\begin{lstlisting}[style=parallactexample]
-def isidentifier(ident):
-    """
-    Determines, if string is valid Python identifier using the ast module.
-    Originally posted at: http://stackoverflow.com/a/29586366
-    """
-
+def _isidentifier_PY3(ident):
     if not isinstance(ident, string_types):
         return False

     try:
-        root = ast.parse(ident)
-    except SyntaxError:
+        ident.encode('ascii')
+    except UnicodeEncodeError:
         return False

-    [... four checks on the parsed AST ...]
+    if not ident.isidentifier():
+        return False
+
+    if keyword.iskeyword(ident):
+        return False

     return True
+
+
+def _isidentifier_PY2(ident):
+    [... the same rules for Python 2, with True, False and None treated as keywords ...]
+
+
+if PY3:
+    isidentifier = _isidentifier_PY3
+else:
+    isidentifier = _isidentifier_PY2
\end{lstlisting}

\textbf{Reference solution B} (the full change to the function).
\begin{lstlisting}[style=parallactexample]
-def isidentifier(ident):
+def isidentifier(ident, allow_dotted=False):
     """
     Determines, if string is valid Python identifier using the ast module.
+    When allow_dotted is True, each dot-separated segment is validated
+    individually so dotted variable paths (e.g. "foo.bar") are accepted.
     Originally posted at: http://stackoverflow.com/a/29586366
     """

+    if allow_dotted and isinstance(ident, string_types) and '.' in ident:
+        parts = ident.split('.')
+        return bool(parts) and all(isidentifier(part) for part in parts)
+
     if not isinstance(ident, string_types):
         return False

     try:
         root = ast.parse(ident)
     [... rest of the original ast-based body, unchanged ...]
\end{lstlisting}

\textbf{Reconciliation solution} (the resulting function, docstring omitted).
A single function keeps B's signature and dotted-path branch and applies A's rules to each name:
\begin{lstlisting}[style=parallactexample]
def isidentifier(ident, allow_dotted=False):
    if allow_dotted and isinstance(ident, string_types) and '.' in ident:
        parts = ident.split('.')
        return bool(parts) and all(isidentifier(part) for part in parts)

    if not isinstance(ident, string_types):
        return False

    try:
        ident.encode('ascii')
    except UnicodeEncodeError:
        return False

    if not ident.isidentifier():
        return False

    if keyword.iskeyword(ident):
        return False

    return True
\end{lstlisting}

\textbf{Tests.}
A's tests come from the \swepro{} task; B's are new.
\begin{center}
\small
\begin{tabular}{@{}lp{0.62\linewidth}@{}}
\toprule
Test file & Cases \\
\midrule
A: \texttt{test\_isidentifier.py} & accepts \texttt{foo}, \texttt{foo1\_23}; rejects \texttt{pass}, \texttt{1234}, \texttt{1234abc}, \texttt{foo bar}, \texttt{no-dashed-names-for-you}, the empty string and names with leading, trailing or only spaces; rejects \texttt{True}, \texttt{False}, \texttt{None}; rejects the non-ASCII name \texttt{k\v{r}\'{\i}\v{z}ek} (the case that fails on the initial repository) \\
B: \texttt{test\_isidentifier\_dotted.py} & with \texttt{allow\_dotted=True}: accepts \texttt{foo.bar}, \texttt{a.b.c}, \texttt{foo}; rejects \texttt{foo.1bad}, \texttt{.foo}. With \texttt{allow\_dotted=False}: rejects \texttt{foo.bar} \\
\bottomrule
\end{tabular}
\end{center}

\textbf{Outcome.}
Claude Code with \claudeopus{} solves this instance in the oracle and in all eight sequential and reentrant scenarios, but fails every parallel scenario under all three combination rules, including the communication oracle.
In the parallel-unaware run, A's agent rewrote the function body without the new parameter and B's agent added \texttt{allow\_dotted} to the old \texttt{ast}-based body.
Each patch passes its own side's tests in isolation, so this is an integration failure (\cref{subsec:failure_modes}): git merge reports a conflict, and neither concatenation nor the reconciler passes the combined test set.
In the A$\to$B hand-off, B's agent starts from A's edited code, and the result passes both test files.
The six open-weight configurations show the same pattern: the oracle and the A$\to$B hand-off succeed in 6 of 6, parallel-unaware in 1 of 6.

\subsection{Complementary example: FortiSwitch CPEs in Vuls}
\label{appendix:example-complementary}

\begin{center}
\small
\begin{tabular}{@{}ll@{}}
\toprule
Repository & \texttt{future-architect/vuls} (Go) \\
\swepro{} task & \texttt{future-architect\_\_vuls-0144135} (v1) \\
Shared interface & the FortiSwitch hardware CPE that A emits and B reads, in \texttt{Convert()}\\
 & (\texttt{contrib/snmp2cpe/pkg/cpe/cpe.go}) \\
Scored tests & the integration test \\
\bottomrule
\end{tabular}
\end{center}

Vuls converts SNMP device data into CPE identifiers (standard names for hardware, operating systems and firmware) that it matches against vulnerability databases.
The \swepro{} issue asks the converter to emit correct hardware, OS, and firmware CPEs for Fortinet FortiSwitch devices.
In this instance, requirement A covers recognizing the FortiSwitch hardware, and requirement B covers choosing the software CPEs from the recognized hardware family.

\textbf{Requirement A} (verbatim).
\begin{quote}\small
Extend Fortinet SNMP conversion to recognize FortiSwitch chassis names. When \texttt{EntPhysicalName} starts with \texttt{FS\_}, normalize the suffix to lowercase and emit a Fortinet hardware CPE whose product is \texttt{fortiswitch-}$\langle$model$\rangle$, while preserving existing FortiGate handling.
\end{quote}

\textbf{Requirement B} (verbatim).
\begin{quote}\small
Correct Fortinet software CPE generation so a validated firmware version is mapped according to the recognized hardware family. FortiGate and FortiWiFi identities should produce \texttt{fortios}, while a FortiSwitch identity should produce both \texttt{fortiswitch} OS and \texttt{fortiswitch\_firmware} CPEs. Unsupported or unidentified families must not receive a generic \texttt{fortios} CPE.
\end{quote}

\textbf{Why the requirements depend on each other.}
A produces the hardware identity and B consumes it, and neither is useful for a FortiSwitch device without the other.
With only A, the converter emits the FortiSwitch hardware CPE but still labels the firmware with the legacy \texttt{fortios} CPE.
With only B, no FortiSwitch hardware CPE exists for B to recognize, so no software CPE is emitted.
The two reference patches edit different lines and merge cleanly; together they emit the full set of CPEs.

\textbf{Reference solution A} (Fortinet branch of \texttt{Convert()}).
\begin{lstlisting}[style=parallactexample]
     case "Fortinet":
         if t, ok := result.EntPhysicalTables[1]; ok {
-            if strings.HasPrefix(t.EntPhysicalName, "FGT_") {
+            switch {
+            case strings.HasPrefix(t.EntPhysicalName, "FGT_"):
                 cpes = append(cpes, fmt.Sprintf("cpe:2.3:h:fortinet:fortigate-%s:-:*:*:*:*:*:*:*", strings.ToLower(strings.TrimPrefix(t.EntPhysicalName, "FGT_"))))
+            case strings.HasPrefix(t.EntPhysicalName, "FS_"):
+                cpes = append(cpes, fmt.Sprintf("cpe:2.3:h:fortinet:fortiswitch-%s:-:*:*:*:*:*:*:*", strings.ToLower(strings.TrimPrefix(t.EntPhysicalName, "FS_"))))
             }
\end{lstlisting}

\textbf{Reference solution B} (the firmware-version case in \texttt{Convert()}, and a new helper that reads the hardware CPEs already emitted).
\begin{lstlisting}[style=parallactexample]
                 case strings.HasPrefix(s, "v") && strings.Contains(s, "build"):
                     if v, _, found := strings.Cut(strings.TrimPrefix(s, "v"), ",build"); found {
                         if _, err := version.NewVersion(v); err == nil {
-                            cpes = append(cpes, fmt.Sprintf("cpe:2.3:o:fortinet:fortios:%s:*:*:*:*:*:*:*", v))
+                            cpes = appendFortinetSoftwareCPEs(cpes, v)
                         }

+// contrib/snmp2cpe/pkg/cpe/fortinet_software.go (new file)
+func appendFortinetSoftwareCPEs(cpes []string, version string) []string {
+    for _, c := range cpes {
+        switch {
+        case strings.HasPrefix(c, "cpe:2.3:h:fortinet:fortigate-"), strings.HasPrefix(c, "cpe:2.3:h:fortinet:fortiwifi-"):
+            cpes = append(cpes, fmt.Sprintf("cpe:2.3:o:fortinet:fortios:%s:*:*:*:*:*:*:*", version))
+        case strings.HasPrefix(c, "cpe:2.3:h:fortinet:fortiswitch-"):
+            cpes = append(cpes,
+                fmt.Sprintf("cpe:2.3:o:fortinet:fortiswitch:%s:*:*:*:*:*:*:*", version),
+                fmt.Sprintf("cpe:2.3:o:fortinet:fortiswitch_firmware:%s:*:*:*:*:*:*:*", version),
+            )
+        }
+    }
+    return cpes
+}
\end{lstlisting}

\textbf{Tests.}
Each side has its own test, and the integration test checks the combined behavior and is the only test scored.
A's test converts a device named \texttt{FS\_108E} and expects exactly the hardware CPE \texttt{fortiswitch-108e}.
B's test converts an unrecognized Fortinet device and expects no CPEs, in particular no \texttt{fortios} CPE.
The integration test converts a FortiSwitch with both a name and a firmware version and checks the exact list:
\begin{lstlisting}[style=parallactexample]
func TestFortiSwitchConversion(t *testing.T) {
    result := snmp.Result{EntPhysicalTables: map[int]snmp.EntPhysicalTable{1: {
        EntPhysicalMfgName:     "Fortinet",
        EntPhysicalName:        "FS_108E",
        EntPhysicalSoftwareRev: "FortiSwitch-108E v6.4.6,build0000,000000 (GA)",
    }}}
    want := []string{
        "cpe:2.3:h:fortinet:fortiswitch-108e:-:*:*:*:*:*:*:*",
        "cpe:2.3:o:fortinet:fortiswitch:6.4.6:*:*:*:*:*:*:*",
        "cpe:2.3:o:fortinet:fortiswitch_firmware:6.4.6:*:*:*:*:*:*:*",
    }
    if diff := cmp.Diff(want, cpe.Convert(result)); diff != "" {
        t.Fatalf("unexpected CPEs (-want +got):\n%s", diff)
    }
}
\end{lstlisting}

\textbf{Outcome.}
Claude Code with \claudeopus{} solves this instance in the oracle but fails it in parallel-unaware under all three combination rules.
Unaware of A, B's agent built the FortiSwitch hardware CPE itself, from the product name in the firmware string, and chose the software CPEs from that name.
Its edits overlap A's in the Fortinet branch, so git merge reports a conflict, and neither concatenation nor the reconciler passes the integration test; keeping both sides' logic would list the hardware CPE twice.
When the agents are aware of each other or can communicate (parallel-aware, messaging, synchronization, and the communication oracle), the combined result passes.
Across the six open-weight configurations, the communication oracle succeeds in 6 of 6, the oracle in 3 of 6, and parallel-unaware in 1 of 6.

\section{Agent Topologies}
\label{appendix:agent-topologies}
\begin{figure}[htp]
    \centering    \includegraphics[width=\linewidth,pagebox=cropbox]{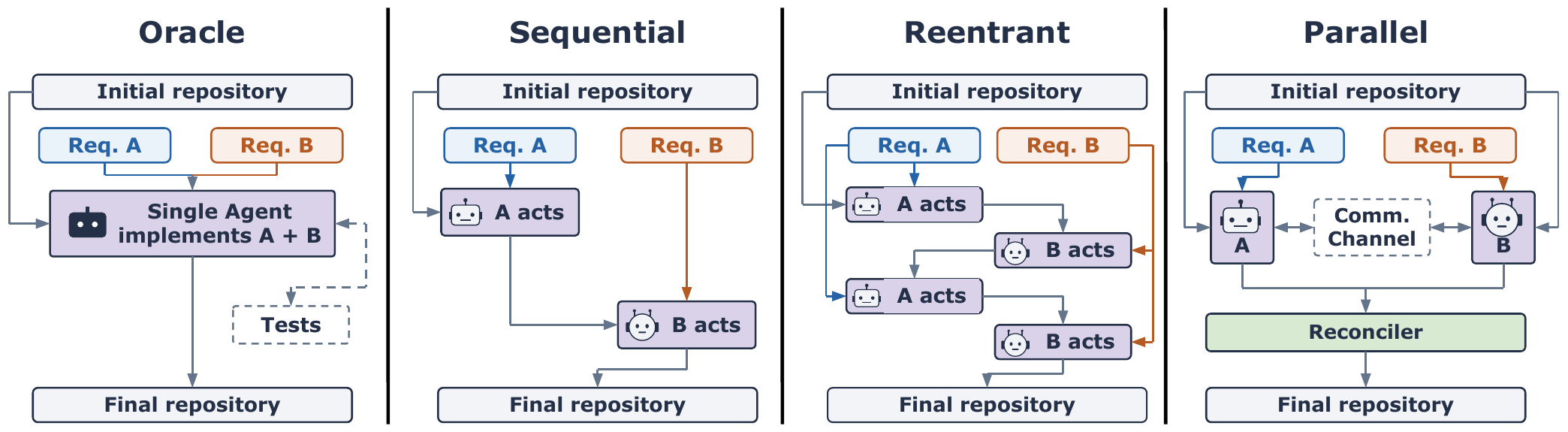}
    \caption{Coordination strategies: single-agent oracle with optional test access, sequential hand-off, reentrant, and parallel execution with optional communication and reconciliation. }
    \label{fig:strategies}
\end{figure}

\cref{fig:strategies} illustrates the topologies and coordination strategies we test.
Oracle agents optionally debug against gold tests. 
The sequential setting is a special case of the reentrant setting, with only one iteration.
In the reentrant setting, we optionally provide agents with access to the edit history. 

In the experiments, each turn leaves a patch against the base commit.
We track added lines by file and exact text.
\emph{Survival} is the share of the first agent's added lines present in the final state.
An \emph{overwrite} is a handoff where the receiving agent removes at least half of the lines its predecessor added on the preceding turn.
A \emph{redo} is a reentrant run where at least three lines added by one agent are removed by the other and restored on the next turn.
These metrics measure textual persistence rather than semantic preservation.

We provide edit history to test whether knowing who made earlier changes helps agents preserve each other's work. Our initial expectation was that history would discourage repeated code reversals. Instead, we observe \textit{more overwrites and redos with history}. Transcripts suggest that agents often interpret a predecessor's edits as an incorrect attempt at their own task, rather than as work serving a separate requirement.
History also makes it easier to restore their own earlier implementation. In the parallel setting, agents have access to a communication channel, whose implementation varies. 

\newpage

\section{Results for all scenarios}
\label{appendix:full-result-tables}

\cref{table:opus-results-full} shows the performance of the agents with \claudeopus{} for all the scenarios tested.

\cref{table:configs-results-full-contra,table:configs-results-full-comple} show all scenarios for the six open-weight configurations on all 155 instances.

\begin{table}[htp]
\centering
\setlength{\tabcolsep}{5pt}
\caption{Pass rate (\%) and relative cost of \claudeopus{} across all 19 scenarios, on its 85 instances (\swepro{} v1). \emph{Pass} is the fraction of instances
whose combined patch passes the scored test group (contradictory: union of both sides'
tests; complementary: the integration test). \emph{Cost} is mean effective tokens (\cref{sec:methods}) relative to that family's \texttt{oracle} run.
Per pass column, the highest and second-highest values are in \textbf{bold} (ties shared).
}
\label{table:opus-results-full}
\begin{tabular}{l rrrr rrrr}
\toprule
 & \multicolumn{4}{c}{Contradictory} & \multicolumn{4}{c}{Complementary} \\
\cmidrule(lr){2-5}\cmidrule(lr){6-9}
 & \multicolumn{3}{c}{Pass (\%)} &  & \multicolumn{3}{c}{Pass (\%)} &  \\
\cmidrule(lr){2-4}\cmidrule(lr){6-8}
Scenario & Seeds & Gen. & All & Cost & Seeds & Gen. & All & Cost \\
\midrule
\multicolumn{9}{l}{\textit{Reference}} \\
Oracle & \PassCell{67} & \PassCell{61} & \PassCell{62} & \CostCell{1.00} & \PassCell{87} & \PassCell{85} & \PassCell{86} & \CostCell{1.00} \\
Orchestrator & \PassCell{58} & \PassCell{58} & \PassCell{58} & \CostCell{1.58} & \PassCell{73} & \PassCell{80} & \PassCell{77} & \CostCell{1.18} \\
Comm-Oracle & \PassCell{50} & \PassCell{58} & \PassCell{56} & \CostCell{2.74} & \PassCell{93} & \PassCell{75} & \PassCell{83} & \CostCell{1.31} \\
Oracle + Gold Tests & \PassCell{100} & \PassCell{97} & \PassCell{98} & \CostCell{0.74} & \PassCell{100} & \PassCell{100} & \PassCell{100} & \CostCell{0.54} \\
Parallel + Gold Tests & \PassCell{17} & \PassCell{45} & \PassCell{38} & \CostCell{0.77} & n/a & n/a & n/a & n/a \\
\midrule
\multicolumn{9}{l}{\textit{Sequential}} \\
\quad A$\to$B (Repo) & \PassCell{50} & \PassCell[\bfseries]{55} & \PassCell[\bfseries]{54} & \CostCell{1.18} & \PassCell{80} & \PassCell{75} & \PassCell{77} & \CostCell{1.12} \\
\quad A$\to$B (Hist) & \PassCell{50} & \PassCell{50} & \PassCell{50} & \CostCell{1.25} & \PassCell{80} & \PassCell{75} & \PassCell{77} & \CostCell{1.15} \\
\quad B$\to$A (Repo) & \PassCell[\bfseries]{58} & \PassCell{47} & \PassCell{50} & \CostCell{1.27} & \PassCell{67} & \PassCell{75} & \PassCell{71} & \CostCell{1.11} \\
\quad B$\to$A (Hist) & \PassCell[\bfseries]{58} & \PassCell{42} & \PassCell{46} & \CostCell{1.26} & \PassCell{53} & \PassCell[\bfseries]{80} & \PassCell{69} & \CostCell{1.11} \\
\midrule
\multicolumn{9}{l}{\textit{Reentrant}} \\
\quad ABAB (Repo) & \PassCell{42} & \PassCell[\bfseries]{53} & \PassCell{50} & \CostCell{1.88} & \PassCell[\bfseries]{87} & \PassCell{75} & \PassCell{80} & \CostCell{1.49} \\
\quad ABAB (Hist) & \PassCell[\bfseries]{58} & \PassCell{42} & \PassCell{46} & \CostCell{2.05} & \PassCell[\bfseries]{93} & \PassCell[\bfseries]{80} & \PassCell[\bfseries]{86} & \CostCell{1.51} \\
\quad BABA (Repo) & \PassCell[\bfseries]{67} & \PassCell{50} & \PassCell[\bfseries]{54} & \CostCell{1.92} & \PassCell[\bfseries]{87} & \PassCell[\bfseries]{85} & \PassCell[\bfseries]{86} & \CostCell{1.51} \\
\quad BABA (Hist) & \PassCell[\bfseries]{58} & \PassCell{50} & \PassCell[\bfseries]{52} & \CostCell{2.01} & \PassCell{80} & \PassCell[\bfseries]{85} & \PassCell[\bfseries]{83} & \CostCell{1.57} \\
\midrule
\multicolumn{9}{l}{\textit{Parallel}} \\
\quad Unaware & \PassCell{0} & \PassCell{21} & \PassCell{16} & \CostCell{1.17} & \PassCell{53} & \PassCell{60} & \PassCell{57} & \CostCell{1.14} \\
\quad Aware & \PassCell{0} & \PassCell{34} & \PassCell{26} & \CostCell{1.14} & \PassCell{60} & \PassCell{60} & \PassCell{60} & \CostCell{1.21} \\
\quad Message & \PassCell{8} & \PassCell{24} & \PassCell{20} & \CostCell{1.42} & \PassCell{67} & \PassCell{60} & \PassCell{63} & \CostCell{1.43} \\
\quad Sync & \PassCell{0} & \PassCell{29} & \PassCell{22} & \CostCell{1.63} & \PassCell{67} & \PassCell{75} & \PassCell{71} & \CostCell{1.43} \\
\quad Message (Redis) & \PassCell{8} & \PassCell{37} & \PassCell{30} & \CostCell{1.61} & \PassCell{67} & \PassCell{70} & \PassCell{69} & \CostCell{1.32} \\
\quad Sync (Redis) & \PassCell{8} & \PassCell{39} & \PassCell{32} & \CostCell{2.23} & \PassCell{80} & \PassCell{75} & \PassCell{77} & \CostCell{1.39} \\
\bottomrule
\end{tabular}
\end{table}
\begin{table}[htp]
\centering
\setlength{\tabcolsep}{3pt}
\caption{Pass rate (\%) and relative cost of the six open-weight configurations across all 19 scenarios on the 85 contradictory instances. \emph{Cost} is mean effective tokens relative to the same configuration's \texttt{oracle}. The \emph{Reference} block grants privileged information and marks upper bounds; among the other scenarios, the highest and second-highest pass rate per column is in \textbf{bold} (ties shared).}
\label{table:configs-results-full-contra}
\resizebox{\textwidth}{!}{
\begin{tabular}{l rr rr rr rr rr rr}
\toprule
 & \multicolumn{6}{c}{Claude Code} & \multicolumn{6}{c}{Codex} \\
\cmidrule(lr){2-7}\cmidrule(lr){8-13}
 & \multicolumn{2}{c}{Qwen} & \multicolumn{2}{c}{DeepSeek} & \multicolumn{2}{c}{GLM} & \multicolumn{2}{c}{Qwen} & \multicolumn{2}{c}{DeepSeek} & \multicolumn{2}{c}{GLM} \\
\cmidrule(lr){2-3}\cmidrule(lr){4-5}\cmidrule(lr){6-7}\cmidrule(lr){8-9}\cmidrule(lr){10-11}\cmidrule(lr){12-13}
Scenario & Pass & Cost & Pass & Cost & Pass & Cost & Pass & Cost & Pass & Cost & Pass & Cost \\
\midrule
\multicolumn{13}{l}{\textit{Reference}} \\
Oracle & \PassCell{48} & \CostCell{1.00} & \PassCell{46} & \CostCell{1.00} & \PassCell{71} & \CostCell{1.00} & \PassCell{66} & \CostCell{1.00} & \PassCell{68} & \CostCell{1.00} & \PassCell{60} & \CostCell{1.00} \\
Orchestrator & \PassCell{34} & \CostCell{1.36} & \PassCell{36} & \CostCell{0.77} & \PassCell{53} & \CostCell{0.73} & \PassCell{49} & \CostCell{0.42} & \PassCell{60} & \CostCell{0.44} & \PassCell{44} & \CostCell{0.26} \\
Comm-Oracle & \PassCell{35} & \CostCell{1.53} & \PassCell{46} & \CostCell{1.70} & \PassCell{54} & \CostCell{1.32} & \PassCell{60} & \CostCell{1.72} & \PassCell{61} & \CostCell{1.94} & \PassCell{33} & \CostCell{1.32} \\
Oracle + Gold Tests & \PassCell{94} & \CostCell{0.64} & \PassCell{95} & \CostCell{0.58} & \PassCell{98} & \CostCell{0.65} & \PassCell{98} & \CostCell{0.71} & \PassCell{99} & \CostCell{0.57} & \PassCell{94} & \CostCell{0.53} \\
Parallel + Gold Tests & \PassCell{33} & \CostCell{0.77} & \PassCell{28} & \CostCell{0.67} & \PassCell{29} & \CostCell{0.61} & \PassCell{29} & \CostCell{0.80} & \PassCell{33} & \CostCell{0.60} & \PassCell{39} & \CostCell{0.70} \\
\midrule
\multicolumn{13}{l}{\textit{Sequential}} \\
\quad A$\to$B (Repo) & \PassCell[\bfseries]{42} & \CostCell{1.14} & \PassCell[\bfseries]{52} & \CostCell{1.11} & \PassCell[\bfseries]{66} & \CostCell{0.92} & \PassCell[\bfseries]{76} & \CostCell{1.08} & \PassCell[\bfseries]{66} & \CostCell{1.02} & \PassCell{55} & \CostCell{1.07} \\
\quad A$\to$B (Hist) & \PassCell{34} & \CostCell{1.08} & \PassCell{36} & \CostCell{1.09} & \PassCell[\bfseries]{64} & \CostCell{0.90} & \PassCell{60} & \CostCell{1.20} & \PassCell[\bfseries]{62} & \CostCell{1.03} & \PassCell[\bfseries]{59} & \CostCell{1.11} \\
\quad B$\to$A (Repo) & \PassCell{31} & \CostCell{1.21} & \PassCell{41} & \CostCell{1.22} & \PassCell{54} & \CostCell{0.99} & \PassCell{44} & \CostCell{1.24} & \PassCell{52} & \CostCell{1.11} & \PassCell{54} & \CostCell{1.02} \\
\quad B$\to$A (Hist) & \PassCell{27} & \CostCell{1.03} & \PassCell{40} & \CostCell{1.14} & \PassCell{52} & \CostCell{0.96} & \PassCell{47} & \CostCell{1.27} & \PassCell{42} & \CostCell{0.99} & \PassCell{52} & \CostCell{1.18} \\
\midrule
\multicolumn{13}{l}{\textit{Reentrant}} \\
\quad ABAB (Repo) & \PassCell{38} & \CostCell{2.03} & \PassCell[\bfseries]{46} & \CostCell{2.10} & \PassCell{59} & \CostCell{1.40} & \PassCell{66} & \CostCell{2.37} & \PassCell{53} & \CostCell{2.09} & \PassCell{53} & \CostCell{1.72} \\
\quad ABAB (Hist) & \PassCell[\bfseries]{41} & \CostCell{1.70} & \PassCell[\bfseries]{46} & \CostCell{1.99} & \PassCell{60} & \CostCell{1.42} & \PassCell[\bfseries]{67} & \CostCell{2.24} & \PassCell{59} & \CostCell{1.98} & \PassCell[\bfseries]{60} & \CostCell{1.71} \\
\quad BABA (Repo) & \PassCell{33} & \CostCell{2.17} & \PassCell{42} & \CostCell{2.25} & \PassCell{52} & \CostCell{1.73} & \PassCell{31} & \CostCell{2.52} & \PassCell{53} & \CostCell{2.22} & \PassCell{54} & \CostCell{1.72} \\
\quad BABA (Hist) & \PassCell{32} & \CostCell{1.74} & \PassCell{39} & \CostCell{2.03} & \PassCell{46} & \CostCell{1.64} & \PassCell{54} & \CostCell{2.38} & \PassCell{55} & \CostCell{2.00} & \PassCell{51} & \CostCell{1.96} \\
\midrule
\multicolumn{13}{l}{\textit{Parallel}} \\
\quad Unaware & \PassCell{15} & \CostCell{1.07} & \PassCell{12} & \CostCell{1.07} & \PassCell{16} & \CostCell{0.94} & \PassCell{20} & \CostCell{1.04} & \PassCell{26} & \CostCell{0.92} & \PassCell{25} & \CostCell{1.05} \\
\quad Aware & \PassCell{14} & \CostCell{1.07} & \PassCell{15} & \CostCell{0.96} & \PassCell{18} & \CostCell{0.91} & \PassCell{20} & \CostCell{1.05} & \PassCell{25} & \CostCell{0.90} & \PassCell{20} & \CostCell{1.04} \\
\quad Message & \PassCell{14} & \CostCell{1.40} & \PassCell{20} & \CostCell{1.40} & \PassCell{29} & \CostCell{1.11} & \PassCell{22} & \CostCell{1.48} & \PassCell{29} & \CostCell{1.32} & \PassCell{29} & \CostCell{1.11} \\
\quad Sync & \PassCell{15} & \CostCell{1.57} & \PassCell{22} & \CostCell{1.67} & \PassCell{34} & \CostCell{1.28} & \PassCell{27} & \CostCell{1.64} & \PassCell{39} & \CostCell{1.74} & \PassCell{27} & \CostCell{1.22} \\
\quad Message (Redis) & \PassCell{12} & \CostCell{1.35} & \PassCell{19} & \CostCell{1.46} & \PassCell{27} & \CostCell{1.16} & \PassCell{31} & \CostCell{1.49} & \PassCell{25} & \CostCell{1.29} & \PassCell{25} & \CostCell{1.29} \\
\quad Sync (Redis) & \PassCell{21} & \CostCell{1.65} & \PassCell{21} & \CostCell{1.71} & \PassCell{28} & \CostCell{1.23} & \PassCell{26} & \CostCell{1.61} & \PassCell{40} & \CostCell{1.55} & \PassCell{29} & \CostCell{1.26} \\
\bottomrule
\end{tabular}
}
\end{table}

\begin{table}[htp]
\centering
\setlength{\tabcolsep}{3pt}
\caption{As \cref{table:configs-results-full-contra}, on the 70 complementary instances (18 scenarios).}
\label{table:configs-results-full-comple}
\resizebox{\textwidth}{!}{
\begin{tabular}{l rr rr rr rr rr rr}
\toprule
 & \multicolumn{6}{c}{Claude Code} & \multicolumn{6}{c}{Codex} \\
\cmidrule(lr){2-7}\cmidrule(lr){8-13}
 & \multicolumn{2}{c}{Qwen} & \multicolumn{2}{c}{DeepSeek} & \multicolumn{2}{c}{GLM} & \multicolumn{2}{c}{Qwen} & \multicolumn{2}{c}{DeepSeek} & \multicolumn{2}{c}{GLM} \\
\cmidrule(lr){2-3}\cmidrule(lr){4-5}\cmidrule(lr){6-7}\cmidrule(lr){8-9}\cmidrule(lr){10-11}\cmidrule(lr){12-13}
Scenario & Pass & Cost & Pass & Cost & Pass & Cost & Pass & Cost & Pass & Cost & Pass & Cost \\
\midrule
\multicolumn{13}{l}{\textit{Reference}} \\
Oracle & \PassCell{61} & \CostCell{1.00} & \PassCell{70} & \CostCell{1.00} & \PassCell{74} & \CostCell{1.00} & \PassCell{84} & \CostCell{1.00} & \PassCell{74} & \CostCell{1.00} & \PassCell{66} & \CostCell{1.00} \\
Orchestrator & \PassCell{51} & \CostCell{0.94} & \PassCell{51} & \CostCell{0.68} & \PassCell{64} & \CostCell{0.62} & \PassCell{66} & \CostCell{0.40} & \PassCell{60} & \CostCell{0.37} & \PassCell{60} & \CostCell{0.26} \\
Comm-Oracle & \PassCell{71} & \CostCell{1.12} & \PassCell{69} & \CostCell{1.25} & \PassCell{71} & \CostCell{1.07} & \PassCell{80} & \CostCell{1.41} & \PassCell{77} & \CostCell{1.31} & \PassCell{60} & \CostCell{1.29} \\
Oracle + Gold Tests & \PassCell{94} & \CostCell{0.74} & \PassCell{96} & \CostCell{0.71} & \PassCell{94} & \CostCell{0.76} & \PassCell{100} & \CostCell{0.86} & \PassCell{97} & \CostCell{0.67} & \PassCell{99} & \CostCell{0.66} \\
\midrule
\multicolumn{13}{l}{\textit{Sequential}} \\
\quad A$\to$B (Repo) & \PassCell{63} & \CostCell{0.98} & \PassCell[\bfseries]{73} & \CostCell{0.89} & \PassCell[\bfseries]{71} & \CostCell{0.82} & \PassCell[\bfseries]{81} & \CostCell{1.18} & \PassCell{71} & \CostCell{0.92} & \PassCell{66} & \CostCell{1.00} \\
\quad A$\to$B (Hist) & \PassCell{61} & \CostCell{0.97} & \PassCell{66} & \CostCell{0.99} & \PassCell[\bfseries]{71} & \CostCell{0.78} & \PassCell{74} & \CostCell{1.15} & \PassCell[\bfseries]{74} & \CostCell{0.93} & \PassCell{67} & \CostCell{0.97} \\
\quad B$\to$A (Repo) & \PassCell{57} & \CostCell{1.18} & \PassCell{63} & \CostCell{1.06} & \PassCell[\bfseries]{71} & \CostCell{1.13} & \PassCell{61} & \CostCell{1.28} & \PassCell{64} & \CostCell{1.09} & \PassCell{63} & \CostCell{1.17} \\
\quad B$\to$A (Hist) & \PassCell{56} & \CostCell{1.08} & \PassCell{56} & \CostCell{1.13} & \PassCell[\bfseries]{71} & \CostCell{1.02} & \PassCell{70} & \CostCell{1.31} & \PassCell{66} & \CostCell{1.12} & \PassCell{64} & \CostCell{1.16} \\
\midrule
\multicolumn{13}{l}{\textit{Reentrant}} \\
\quad ABAB (Repo) & \PassCell[\bfseries]{66} & \CostCell{1.45} & \PassCell{67} & \CostCell{1.74} & \PassCell{66} & \CostCell{1.40} & \PassCell[\bfseries]{79} & \CostCell{2.32} & \PassCell{70} & \CostCell{1.82} & \PassCell[\bfseries]{71} & \CostCell{1.63} \\
\quad ABAB (Hist) & \PassCell[\bfseries]{69} & \CostCell{1.41} & \PassCell{64} & \CostCell{1.73} & \PassCell[\bfseries]{73} & \CostCell{1.46} & \PassCell{76} & \CostCell{2.32} & \PassCell{70} & \CostCell{1.86} & \PassCell{66} & \CostCell{1.62} \\
\quad BABA (Repo) & \PassCell{61} & \CostCell{1.60} & \PassCell{60} & \CostCell{1.92} & \PassCell{69} & \CostCell{1.55} & \PassCell{47} & \CostCell{2.57} & \PassCell{61} & \CostCell{1.97} & \PassCell{67} & \CostCell{1.76} \\
\quad BABA (Hist) & \PassCell[\bfseries]{66} & \CostCell{1.52} & \PassCell{69} & \CostCell{1.82} & \PassCell{66} & \CostCell{1.40} & \PassCell{69} & \CostCell{2.48} & \PassCell{69} & \CostCell{2.00} & \PassCell[\bfseries]{74} & \CostCell{1.87} \\
\midrule
\multicolumn{13}{l}{\textit{Parallel}} \\
\quad Unaware & \PassCell{49} & \CostCell{1.03} & \PassCell{57} & \CostCell{1.10} & \PassCell{56} & \CostCell{0.94} & \PassCell{56} & \CostCell{1.17} & \PassCell{50} & \CostCell{1.00} & \PassCell{50} & \CostCell{1.07} \\
\quad Aware & \PassCell{46} & \CostCell{0.97} & \PassCell{59} & \CostCell{0.96} & \PassCell{66} & \CostCell{0.87} & \PassCell{69} & \CostCell{1.16} & \PassCell{63} & \CostCell{1.01} & \PassCell{47} & \CostCell{1.24} \\
\quad Message & \PassCell{59} & \CostCell{1.21} & \PassCell{70} & \CostCell{1.26} & \PassCell{69} & \CostCell{0.98} & \PassCell{70} & \CostCell{1.38} & \PassCell[\bfseries]{73} & \CostCell{1.19} & \PassCell{54} & \CostCell{1.08} \\
\quad Sync & \PassCell{57} & \CostCell{1.37} & \PassCell[\bfseries]{71} & \CostCell{1.29} & \PassCell[\bfseries]{71} & \CostCell{0.94} & \PassCell{69} & \CostCell{1.44} & \PassCell{69} & \CostCell{1.29} & \PassCell{47} & \CostCell{1.48} \\
\quad Message (Redis) & \PassCell{51} & \CostCell{1.24} & \PassCell{63} & \CostCell{1.23} & \PassCell{70} & \CostCell{1.01} & \PassCell{63} & \CostCell{1.31} & \PassCell{66} & \CostCell{1.16} & \PassCell{53} & \CostCell{1.14} \\
\quad Sync (Redis) & \PassCell{57} & \CostCell{1.27} & \PassCell{61} & \CostCell{1.26} & \PassCell[\bfseries]{73} & \CostCell{1.09} & \PassCell{74} & \CostCell{1.50} & \PassCell{66} & \CostCell{1.26} & \PassCell{50} & \CostCell{1.20} \\
\bottomrule
\end{tabular}
}
\end{table}

\clearpage

\section{Comparison of all configurations on the shared instances}
\label{appendix:same-subset}

\cref{table:same-subset-contra,table:same-subset-comple} compare all seven configurations on the 85 instances they share (the instances built on \swepro{} v1; \cref{appendix:swepro-versions}), so that every cell is computed on identical instances.
Cost excludes cache-write tokens for every configuration, including Claude Code with \claudeopus{}, for comparability (\cref{appendix:token-accounting}); this is why its costs differ slightly from \cref{table:opus-results-full}.

\begin{table}[htp]
\centering
\setlength{\tabcolsep}{3pt}
\caption{All seven configurations on the 50 contradictory instances they share (built on \swepro{} v1). \emph{Cost} is relative to each configuration's own \texttt{oracle}, excluding cache-write tokens. Outside the \emph{Reference} block, the highest and second-highest pass rate per column is in \textbf{bold} (ties shared).}
\label{table:same-subset-contra}
\resizebox{\textwidth}{!}{
\begin{tabular}{l rr rr rr rr rr rr rr}
\toprule
 & \multicolumn{8}{c}{Claude Code} & \multicolumn{6}{c}{Codex} \\
\cmidrule(lr){2-9}\cmidrule(lr){10-15}
 & \multicolumn{2}{c}{Opus~4.8} & \multicolumn{2}{c}{Qwen} & \multicolumn{2}{c}{DeepSeek} & \multicolumn{2}{c}{GLM} & \multicolumn{2}{c}{Qwen} & \multicolumn{2}{c}{DeepSeek} & \multicolumn{2}{c}{GLM} \\
\cmidrule(lr){2-3}\cmidrule(lr){4-5}\cmidrule(lr){6-7}\cmidrule(lr){8-9}\cmidrule(lr){10-11}\cmidrule(lr){12-13}\cmidrule(lr){14-15}
Scenario & Pass & Cost & Pass & Cost & Pass & Cost & Pass & Cost & Pass & Cost & Pass & Cost & Pass & Cost \\
\midrule
\multicolumn{15}{l}{\textit{Reference}} \\
Oracle & \PassCell{62} & \CostCell{1.00} & \PassCell{56} & \CostCell{1.00} & \PassCell{56} & \CostCell{1.00} & \PassCell{84} & \CostCell{1.00} & \PassCell{78} & \CostCell{1.00} & \PassCell{78} & \CostCell{1.00} & \PassCell{78} & \CostCell{1.00} \\
Orchestrator & \PassCell{58} & \CostCell{1.43} & \PassCell{44} & \CostCell{1.36} & \PassCell{48} & \CostCell{0.68} & \PassCell{70} & \CostCell{0.78} & \PassCell{60} & \CostCell{0.45} & \PassCell{76} & \CostCell{0.44} & \PassCell{58} & \CostCell{0.32} \\
Comm-Oracle & \PassCell{56} & \CostCell{2.90} & \PassCell{44} & \CostCell{1.56} & \PassCell{54} & \CostCell{1.65} & \PassCell{72} & \CostCell{1.40} & \PassCell{66} & \CostCell{1.84} & \PassCell{80} & \CostCell{1.84} & \PassCell{40} & \CostCell{1.51} \\
Oracle + Gold Tests & \PassCell{98} & \CostCell{0.74} & \PassCell{90} & \CostCell{0.70} & \PassCell{92} & \CostCell{0.54} & \PassCell{100} & \CostCell{0.64} & \PassCell{100} & \CostCell{0.79} & \PassCell{98} & \CostCell{0.57} & \PassCell{98} & \CostCell{0.59} \\
Parallel + Gold Tests & \PassCell{38} & \CostCell{0.72} & \PassCell{34} & \CostCell{0.82} & \PassCell{30} & \CostCell{0.66} & \PassCell{34} & \CostCell{0.58} & \PassCell{26} & \CostCell{0.85} & \PassCell{36} & \CostCell{0.64} & \PassCell{42} & \CostCell{0.78} \\
\midrule
\multicolumn{15}{l}{\textit{Sequential}} \\
\quad A$\to$B (Repo) & \PassCell[\bfseries]{54} & \CostCell{1.14} & \PassCell[\bfseries]{52} & \CostCell{1.12} & \PassCell[\bfseries]{60} & \CostCell{0.96} & \PassCell[\bfseries]{86} & \CostCell{0.88} & \PassCell[\bfseries]{84} & \CostCell{1.12} & \PassCell[\bfseries]{72} & \CostCell{0.97} & \PassCell{62} & \CostCell{1.19} \\
\quad A$\to$B (Hist) & \PassCell{50} & \CostCell{1.21} & \PassCell{42} & \CostCell{1.09} & \PassCell{42} & \CostCell{1.04} & \PassCell[\bfseries]{80} & \CostCell{0.88} & \PassCell{72} & \CostCell{1.23} & \PassCell[\bfseries]{74} & \CostCell{1.00} & \PassCell[\bfseries]{68} & \CostCell{1.26} \\
\quad B$\to$A (Repo) & \PassCell{50} & \CostCell{1.26} & \PassCell{40} & \CostCell{1.13} & \PassCell{48} & \CostCell{1.13} & \PassCell{70} & \CostCell{0.94} & \PassCell{52} & \CostCell{1.27} & \PassCell{60} & \CostCell{1.12} & \PassCell{60} & \CostCell{1.21} \\
\quad B$\to$A (Hist) & \PassCell{46} & \CostCell{1.25} & \PassCell{36} & \CostCell{1.04} & \PassCell{46} & \CostCell{1.07} & \PassCell{64} & \CostCell{0.98} & \PassCell{56} & \CostCell{1.26} & \PassCell{52} & \CostCell{1.05} & \PassCell{60} & \CostCell{1.41} \\
\midrule
\multicolumn{15}{l}{\textit{Reentrant}} \\
\quad ABAB (Repo) & \PassCell{50} & \CostCell{1.80} & \PassCell{46} & \CostCell{2.04} & \PassCell{52} & \CostCell{1.91} & \PassCell{74} & \CostCell{1.47} & \PassCell{74} & \CostCell{2.42} & \PassCell{62} & \CostCell{2.13} & \PassCell[\bfseries]{64} & \CostCell{2.04} \\
\quad ABAB (Hist) & \PassCell{46} & \CostCell{1.98} & \PassCell[\bfseries]{48} & \CostCell{1.75} & \PassCell[\bfseries]{54} & \CostCell{1.78} & \PassCell[\bfseries]{80} & \CostCell{1.39} & \PassCell[\bfseries]{76} & \CostCell{2.39} & \PassCell[\bfseries]{72} & \CostCell{2.04} & \PassCell[\bfseries]{68} & \CostCell{1.98} \\
\quad BABA (Repo) & \PassCell[\bfseries]{54} & \CostCell{1.86} & \PassCell{40} & \CostCell{2.17} & \PassCell{50} & \CostCell{1.97} & \PassCell{70} & \CostCell{1.75} & \PassCell{38} & \CostCell{2.62} & \PassCell{66} & \CostCell{2.10} & \PassCell[\bfseries]{64} & \CostCell{2.05} \\
\quad BABA (Hist) & \PassCell[\bfseries]{52} & \CostCell{1.95} & \PassCell{42} & \CostCell{1.78} & \PassCell{44} & \CostCell{1.83} & \PassCell{60} & \CostCell{1.51} & \PassCell{66} & \CostCell{2.43} & \PassCell{66} & \CostCell{2.02} & \PassCell{62} & \CostCell{2.09} \\
\midrule
\multicolumn{15}{l}{\textit{Parallel}} \\
\quad Unaware & \PassCell{16} & \CostCell{1.13} & \PassCell{16} & \CostCell{1.01} & \PassCell{16} & \CostCell{1.01} & \PassCell{20} & \CostCell{0.87} & \PassCell{24} & \CostCell{1.14} & \PassCell{28} & \CostCell{0.90} & \PassCell{28} & \CostCell{1.15} \\
\quad Aware & \PassCell{26} & \CostCell{1.10} & \PassCell{16} & \CostCell{1.04} & \PassCell{18} & \CostCell{0.89} & \PassCell{26} & \CostCell{0.88} & \PassCell{24} & \CostCell{1.08} & \PassCell{26} & \CostCell{0.91} & \PassCell{22} & \CostCell{1.27} \\
\quad Message & \PassCell{20} & \CostCell{1.43} & \PassCell{18} & \CostCell{1.46} & \PassCell{22} & \CostCell{1.32} & \PassCell{42} & \CostCell{1.17} & \PassCell{24} & \CostCell{1.52} & \PassCell{34} & \CostCell{1.33} & \PassCell{34} & \CostCell{1.29} \\
\quad Sync & \PassCell{22} & \CostCell{1.64} & \PassCell{18} & \CostCell{1.52} & \PassCell{26} & \CostCell{1.51} & \PassCell{48} & \CostCell{1.28} & \PassCell{30} & \CostCell{1.70} & \PassCell{50} & \CostCell{1.86} & \PassCell{32} & \CostCell{1.26} \\
\quad Message (Redis) & \PassCell{30} & \CostCell{1.64} & \PassCell{14} & \CostCell{1.29} & \PassCell{22} & \CostCell{1.43} & \PassCell{36} & \CostCell{1.10} & \PassCell{36} & \CostCell{1.57} & \PassCell{30} & \CostCell{1.28} & \PassCell{30} & \CostCell{1.60} \\
\quad Sync (Redis) & \PassCell{32} & \CostCell{2.30} & \PassCell{24} & \CostCell{1.72} & \PassCell{26} & \CostCell{1.62} & \PassCell{36} & \CostCell{1.23} & \PassCell{30} & \CostCell{1.75} & \PassCell{46} & \CostCell{1.62} & \PassCell{36} & \CostCell{1.45} \\
\bottomrule
\end{tabular}
}
\end{table}

\begin{table}[htp]
\centering
\setlength{\tabcolsep}{3pt}
\caption{As \cref{table:same-subset-contra}, on the 35 complementary instances.}
\label{table:same-subset-comple}
\resizebox{\textwidth}{!}{
\begin{tabular}{l rr rr rr rr rr rr rr}
\toprule
 & \multicolumn{8}{c}{Claude Code} & \multicolumn{6}{c}{Codex} \\
\cmidrule(lr){2-9}\cmidrule(lr){10-15}
 & \multicolumn{2}{c}{Opus~4.8} & \multicolumn{2}{c}{Qwen} & \multicolumn{2}{c}{DeepSeek} & \multicolumn{2}{c}{GLM} & \multicolumn{2}{c}{Qwen} & \multicolumn{2}{c}{DeepSeek} & \multicolumn{2}{c}{GLM} \\
\cmidrule(lr){2-3}\cmidrule(lr){4-5}\cmidrule(lr){6-7}\cmidrule(lr){8-9}\cmidrule(lr){10-11}\cmidrule(lr){12-13}\cmidrule(lr){14-15}
Scenario & Pass & Cost & Pass & Cost & Pass & Cost & Pass & Cost & Pass & Cost & Pass & Cost & Pass & Cost \\
\midrule
\multicolumn{15}{l}{\textit{Reference}} \\
Oracle & \PassCell{86} & \CostCell{1.00} & \PassCell{66} & \CostCell{1.00} & \PassCell{74} & \CostCell{1.00} & \PassCell{83} & \CostCell{1.00} & \PassCell{94} & \CostCell{1.00} & \PassCell{80} & \CostCell{1.00} & \PassCell{71} & \CostCell{1.00} \\
Orchestrator & \PassCell{77} & \CostCell{1.08} & \PassCell{63} & \CostCell{0.90} & \PassCell{57} & \CostCell{0.62} & \PassCell{74} & \CostCell{0.59} & \PassCell{74} & \CostCell{0.37} & \PassCell{63} & \CostCell{0.34} & \PassCell{69} & \CostCell{0.22} \\
Comm-Oracle & \PassCell{83} & \CostCell{1.33} & \PassCell{80} & \CostCell{1.09} & \PassCell{89} & \CostCell{1.27} & \PassCell{80} & \CostCell{1.05} & \PassCell{91} & \CostCell{1.32} & \PassCell{83} & \CostCell{1.18} & \PassCell{69} & \CostCell{0.88} \\
Oracle + Gold Tests & \PassCell{100} & \CostCell{0.53} & \PassCell{91} & \CostCell{0.76} & \PassCell{94} & \CostCell{0.72} & \PassCell{97} & \CostCell{0.72} & \PassCell{100} & \CostCell{0.75} & \PassCell{100} & \CostCell{0.63} & \PassCell{100} & \CostCell{0.48} \\
\midrule
\multicolumn{15}{l}{\textit{Sequential}} \\
\quad A$\to$B (Repo) & \PassCell{77} & \CostCell{1.09} & \PassCell{66} & \CostCell{0.91} & \PassCell[\bfseries]{83} & \CostCell{0.84} & \PassCell{83} & \CostCell{0.74} & \PassCell[\bfseries]{89} & \CostCell{1.09} & \PassCell[\bfseries]{83} & \CostCell{0.85} & \PassCell{74} & \CostCell{0.88} \\
\quad A$\to$B (Hist) & \PassCell{77} & \CostCell{1.13} & \PassCell[\bfseries]{69} & \CostCell{0.81} & \PassCell{74} & \CostCell{0.91} & \PassCell{86} & \CostCell{0.73} & \PassCell[\bfseries]{89} & \CostCell{1.04} & \PassCell[\bfseries]{83} & \CostCell{0.95} & \PassCell{74} & \CostCell{0.80} \\
\quad B$\to$A (Repo) & \PassCell{71} & \CostCell{1.11} & \PassCell{66} & \CostCell{1.11} & \PassCell{77} & \CostCell{1.10} & \PassCell{83} & \CostCell{1.02} & \PassCell{71} & \CostCell{1.20} & \PassCell{69} & \CostCell{1.02} & \PassCell[\bfseries]{80} & \CostCell{0.91} \\
\quad B$\to$A (Hist) & \PassCell{69} & \CostCell{1.12} & \PassCell{66} & \CostCell{1.01} & \PassCell{63} & \CostCell{1.16} & \PassCell[\bfseries]{89} & \CostCell{1.04} & \PassCell{80} & \CostCell{1.26} & \PassCell{77} & \CostCell{0.97} & \PassCell{77} & \CostCell{1.01} \\
\midrule
\multicolumn{15}{l}{\textit{Reentrant}} \\
\quad ABAB (Repo) & \PassCell{80} & \CostCell{1.44} & \PassCell[\bfseries]{69} & \CostCell{1.38} & \PassCell{66} & \CostCell{1.78} & \PassCell{80} & \CostCell{1.34} & \PassCell[\bfseries]{89} & \CostCell{2.23} & \PassCell[\bfseries]{89} & \CostCell{1.82} & \PassCell[\bfseries]{86} & \CostCell{1.34} \\
\quad ABAB (Hist) & \PassCell[\bfseries]{86} & \CostCell{1.49} & \PassCell[\bfseries]{77} & \CostCell{1.31} & \PassCell{69} & \CostCell{1.81} & \PassCell{86} & \CostCell{1.36} & \PassCell[\bfseries]{86} & \CostCell{2.21} & \PassCell{77} & \CostCell{1.83} & \PassCell{71} & \CostCell{1.39} \\
\quad BABA (Repo) & \PassCell[\bfseries]{86} & \CostCell{1.46} & \PassCell{66} & \CostCell{1.50} & \PassCell{69} & \CostCell{1.90} & \PassCell{86} & \CostCell{1.58} & \PassCell{60} & \CostCell{2.40} & \PassCell{66} & \CostCell{1.89} & \PassCell[\bfseries]{86} & \CostCell{1.42} \\
\quad BABA (Hist) & \PassCell[\bfseries]{83} & \CostCell{1.55} & \PassCell[\bfseries]{77} & \CostCell{1.46} & \PassCell[\bfseries]{80} & \CostCell{1.92} & \PassCell{80} & \CostCell{1.43} & \PassCell{80} & \CostCell{2.36} & \PassCell{80} & \CostCell{1.80} & \PassCell[\bfseries]{80} & \CostCell{1.58} \\
\midrule
\multicolumn{15}{l}{\textit{Parallel}} \\
\quad Unaware & \PassCell{57} & \CostCell{1.12} & \PassCell{54} & \CostCell{0.92} & \PassCell{63} & \CostCell{1.15} & \PassCell{63} & \CostCell{0.82} & \PassCell{57} & \CostCell{1.08} & \PassCell{54} & \CostCell{0.90} & \PassCell{57} & \CostCell{0.95} \\
\quad Aware & \PassCell{60} & \CostCell{1.20} & \PassCell{57} & \CostCell{0.90} & \PassCell{69} & \CostCell{1.02} & \PassCell{77} & \CostCell{0.87} & \PassCell{77} & \CostCell{1.10} & \PassCell{74} & \CostCell{0.97} & \PassCell{49} & \CostCell{1.12} \\
\quad Message & \PassCell{63} & \CostCell{1.46} & \PassCell[\bfseries]{69} & \CostCell{1.08} & \PassCell[\bfseries]{80} & \CostCell{1.27} & \PassCell{80} & \CostCell{0.92} & \PassCell[\bfseries]{89} & \CostCell{1.20} & \PassCell[\bfseries]{83} & \CostCell{1.05} & \PassCell{71} & \CostCell{0.84} \\
\quad Sync & \PassCell{71} & \CostCell{1.45} & \PassCell{63} & \CostCell{1.38} & \PassCell[\bfseries]{80} & \CostCell{1.28} & \PassCell{80} & \CostCell{0.99} & \PassCell{80} & \CostCell{1.35} & \PassCell{77} & \CostCell{1.27} & \PassCell{54} & \CostCell{1.20} \\
\quad Message (Redis) & \PassCell{69} & \CostCell{1.33} & \PassCell{60} & \CostCell{1.21} & \PassCell{74} & \CostCell{1.15} & \PassCell{83} & \CostCell{0.97} & \PassCell{74} & \CostCell{1.17} & \PassCell{80} & \CostCell{1.08} & \PassCell{60} & \CostCell{1.02} \\
\quad Sync (Redis) & \PassCell{77} & \CostCell{1.41} & \PassCell{60} & \CostCell{1.25} & \PassCell{66} & \CostCell{1.26} & \PassCell[\bfseries]{94} & \CostCell{1.05} & \PassCell{80} & \CostCell{1.35} & \PassCell[\bfseries]{83} & \CostCell{1.16} & \PassCell{60} & \CostCell{0.94} \\
\bottomrule
\end{tabular}
}
\end{table}

\clearpage

\section{Performance By Agent}
\label{appendix:per-agent-performance}

\cref{table:per-agent-pass-rate-opus} shows the per-agent pass rate across all the scenarios tested for \claudeopus{}, and \cref{table:configs-per-agent-claude,table:configs-per-agent-codex} show the same for the six open-weight configurations on the 85 contradictory instances.

\begin{table}[htp]
\centering
\setlength{\tabcolsep}{6pt}
\caption{Per-agent pass rates (\%) for \claudeopus{} on its 50 contradictory instances (\swepro{} v1; contradictory tasks). \emph{A} and \emph{B} are the fractions where agent~A's and agent~B's patch passes its own tests in isolation, and \emph{Either} is at least one of them. \emph{Both} is the combined patch passing the scored test group (the joint success rate of \cref{table:opus-results-full}), a different, merged artifact. Because A/B/Either measure isolated patches while \emph{Both} measures the merged patch, \emph{Both} may exceed \emph{B} and \emph{Either} need not equal $A+B-\text{Both}$. Per column, the highest and second-highest values are in \textbf{bold} (ties shared).}
\label{table:per-agent-pass-rate-opus}
\begin{tabular}{l rrrr} 
\toprule
 & \multicolumn{4}{c}{Contradictory} \\ 
\cmidrule(lr){2-5} 
Scenario & A & B & Both & Either \\ 
\midrule
Oracle & \PassCell{78} & \PassCell{76} & \PassCell[\bfseries]{62} & \PassCell{88} \\ 
Orchestrator & \PassCell{68} & \PassCell{76} & \PassCell{58} & \PassCell{86} \\ 
Oracle + Gold Tests & \PassCell[\bfseries]{98} & \PassCell[\bfseries]{92} & \PassCell[\bfseries]{98} & \PassCell[\bfseries]{98} \\ 
Parallel + Gold Tests & \PassCell[\bfseries]{98} & \PassCell[\bfseries]{98} & \PassCell{38} & \PassCell[\bfseries]{98} \\ 
\midrule
Sequential A$\to$B (Repo) & \PassCell{74} & \PassCell{74} & \PassCell{54} & \PassCell{92} \\ 
Sequential A$\to$B (Hist) & \PassCell{64} & \PassCell{80} & \PassCell{50} & \PassCell{92} \\ 
Sequential B$\to$A (Repo) & \PassCell{76} & \PassCell{66} & \PassCell{50} & \PassCell{92} \\ 
Sequential B$\to$A (Hist) & \PassCell[\bfseries]{80} & \PassCell{62} & \PassCell{46} & \PassCell{92} \\ 
\midrule
Reentrant ABAB (Repo) & \PassCell{72} & \PassCell{74} & \PassCell{50} & \PassCell{94} \\ 
Reentrant ABAB (Hist) & \PassCell{58} & \PassCell{76} & \PassCell{46} & \PassCell{88} \\ 
Reentrant BABA (Repo) & \PassCell{76} & \PassCell{74} & \PassCell{54} & \PassCell{92} \\ 
Reentrant BABA (Hist) & \PassCell{78} & \PassCell{66} & \PassCell{52} & \PassCell{90} \\ 
\midrule
Parallel Unaware & \PassCell{68} & \PassCell[\bfseries]{92} & \PassCell{16} & \PassCell[\bfseries]{98} \\ 
Parallel Aware & \PassCell{78} & \PassCell[\bfseries]{92} & \PassCell{26} & \PassCell[\bfseries]{96} \\ 
Parallel Message & \PassCell{72} & \PassCell[\bfseries]{92} & \PassCell{20} & \PassCell{94} \\ 
Parallel Sync & \PassCell{70} & \PassCell[\bfseries]{92} & \PassCell{22} & \PassCell[\bfseries]{96} \\ 
Parallel Message (Redis) & \PassCell{74} & \PassCell{88} & \PassCell{30} & \PassCell[\bfseries]{96} \\ 
Parallel Sync (Redis) & \PassCell{70} & \PassCell{84} & \PassCell{32} & \PassCell{92} \\ 
Parallel Comm-Oracle & \PassCell{56} & \PassCell{70} & \PassCell{56} & \PassCell{86} \\ 
\bottomrule
\end{tabular}
\end{table}
\begin{table}[htp]
\centering
\setlength{\tabcolsep}{3pt}
\caption{Per-agent pass rates (\%) for Claude Code with each open-weight model on the contradictory instances. Columns as in \cref{table:per-agent-pass-rate-opus}: \emph{A} and \emph{B} are the fractions where agent~A's and agent~B's patch passes its own tests in isolation, \emph{Either} is at least one of them, and \emph{Both} is joint success. Per column, the highest and second-highest values are in \textbf{bold} (ties shared).}
\label{table:configs-per-agent-claude}
\resizebox{\textwidth}{!}{
\begin{tabular}{l rrrr rrrr rrrr}
\toprule
 & \multicolumn{4}{c}{Qwen} & \multicolumn{4}{c}{DeepSeek} & \multicolumn{4}{c}{GLM} \\
\cmidrule(lr){2-5}\cmidrule(lr){6-9}\cmidrule(lr){10-13}
Scenario & A & B & Both & Either & A & B & Both & Either & A & B & Both & Either \\
\midrule
Oracle & \PassCell{64} & \PassCell{76} & \PassCell[\bfseries]{48} & \PassCell{92} & \PassCell{68} & \PassCell{71} & \PassCell{46} & \PassCell{91} & \PassCell[\bfseries]{86} & \PassCell{75} & \PassCell[\bfseries]{71} & \PassCell{92} \\
Orchestrator & \PassCell{46} & \PassCell{73} & \PassCell{34} & \PassCell{87} & \PassCell{56} & \PassCell{73} & \PassCell{36} & \PassCell{91} & \PassCell{66} & \PassCell{74} & \PassCell{53} & \PassCell{89} \\
Comm-Oracle & \PassCell{54} & \PassCell{74} & \PassCell{35} & \PassCell{85} & \PassCell{55} & \PassCell{65} & \PassCell{46} & \PassCell{81} & \PassCell{72} & \PassCell{68} & \PassCell{54} & \PassCell{91} \\
Oracle + Gold Tests & \PassCell[\bfseries]{94} & \PassCell[\bfseries]{82} & \PassCell[\bfseries]{94} & \PassCell[\bfseries]{95} & \PassCell[\bfseries]{92} & \PassCell{85} & \PassCell[\bfseries]{95} & \PassCell[\bfseries]{95} & \PassCell[\bfseries]{96} & \PassCell{88} & \PassCell[\bfseries]{98} & \PassCell[\bfseries]{98} \\
Parallel + Gold Tests & \PassCell[\bfseries]{92} & \PassCell[\bfseries]{94} & \PassCell{33} & \PassCell[\bfseries]{94} & \PassCell[\bfseries]{93} & \PassCell[\bfseries]{99} & \PassCell{28} & \PassCell[\bfseries]{99} & \PassCell[\bfseries]{96} & \PassCell[\bfseries]{99} & \PassCell{29} & \PassCell[\bfseries]{99} \\
\midrule
Sequential A$\to$B (Repo) & \PassCell{64} & \PassCell{69} & \PassCell{42} & \PassCell{88} & \PassCell{69} & \PassCell{72} & \PassCell[\bfseries]{52} & \PassCell{89} & \PassCell{81} & \PassCell{79} & \PassCell{66} & \PassCell{93} \\
Sequential A$\to$B (Hist) & \PassCell{56} & \PassCell{72} & \PassCell{34} & \PassCell{89} & \PassCell{60} & \PassCell{73} & \PassCell{36} & \PassCell{87} & \PassCell{79} & \PassCell{81} & \PassCell{64} & \PassCell{93} \\
Sequential B$\to$A (Repo) & \PassCell{60} & \PassCell{51} & \PassCell{31} & \PassCell{82} & \PassCell{65} & \PassCell{54} & \PassCell{41} & \PassCell{79} & \PassCell{80} & \PassCell{59} & \PassCell{54} & \PassCell{86} \\
Sequential B$\to$A (Hist) & \PassCell{56} & \PassCell{49} & \PassCell{27} & \PassCell{81} & \PassCell{72} & \PassCell{48} & \PassCell{40} & \PassCell{78} & \PassCell{81} & \PassCell{58} & \PassCell{52} & \PassCell{87} \\
\midrule
Reentrant ABAB (Repo) & \PassCell{52} & \PassCell{73} & \PassCell{38} & \PassCell{87} & \PassCell{67} & \PassCell{74} & \PassCell{46} & \PassCell{93} & \PassCell{76} & \PassCell{75} & \PassCell{59} & \PassCell{92} \\
Reentrant ABAB (Hist) & \PassCell{60} & \PassCell{71} & \PassCell{41} & \PassCell{88} & \PassCell{64} & \PassCell{74} & \PassCell{46} & \PassCell{87} & \PassCell{74} & \PassCell{78} & \PassCell{60} & \PassCell{89} \\
Reentrant BABA (Repo) & \PassCell{61} & \PassCell{52} & \PassCell{33} & \PassCell{81} & \PassCell{69} & \PassCell{58} & \PassCell{42} & \PassCell{85} & \PassCell{78} & \PassCell{61} & \PassCell{52} & \PassCell{85} \\
Reentrant BABA (Hist) & \PassCell{62} & \PassCell{53} & \PassCell{32} & \PassCell{79} & \PassCell{75} & \PassCell{54} & \PassCell{39} & \PassCell{87} & \PassCell{81} & \PassCell{56} & \PassCell{46} & \PassCell{91} \\
\midrule
Parallel Unaware & \PassCell{61} & \PassCell[\bfseries]{82} & \PassCell{15} & \PassCell{93} & \PassCell{67} & \PassCell{82} & \PassCell{12} & \PassCell{94} & \PassCell{84} & \PassCell[\bfseries]{89} & \PassCell{16} & \PassCell[\bfseries]{98} \\
Parallel Aware & \PassCell{64} & \PassCell{79} & \PassCell{14} & \PassCell{91} & \PassCell{71} & \PassCell[\bfseries]{86} & \PassCell{15} & \PassCell{94} & \PassCell{80} & \PassCell{86} & \PassCell{18} & \PassCell{96} \\
Parallel Message & \PassCell{66} & \PassCell{78} & \PassCell{14} & \PassCell{91} & \PassCell{73} & \PassCell{82} & \PassCell{20} & \PassCell{93} & \PassCell{85} & \PassCell{85} & \PassCell{29} & \PassCell{94} \\
Parallel Sync & \PassCell{60} & \PassCell{81} & \PassCell{15} & \PassCell{89} & \PassCell{66} & \PassCell{75} & \PassCell{22} & \PassCell{87} & \PassCell{80} & \PassCell{84} & \PassCell{34} & \PassCell{94} \\
Parallel Message (Redis) & \PassCell{60} & \PassCell{78} & \PassCell{12} & \PassCell{89} & \PassCell{74} & \PassCell{84} & \PassCell{19} & \PassCell{94} & \PassCell{75} & \PassCell{85} & \PassCell{27} & \PassCell{92} \\
Parallel Sync (Redis) & \PassCell{65} & \PassCell{80} & \PassCell{21} & \PassCell{91} & \PassCell{68} & \PassCell{76} & \PassCell{21} & \PassCell{93} & \PassCell{73} & \PassCell{80} & \PassCell{28} & \PassCell{94} \\
\bottomrule
\end{tabular}
}
\end{table}

\begin{table}[htp]
\centering
\setlength{\tabcolsep}{3pt}
\caption{As \cref{table:configs-per-agent-claude}, for Codex.}
\label{table:configs-per-agent-codex}
\resizebox{\textwidth}{!}{
\begin{tabular}{l rrrr rrrr rrrr}
\toprule
 & \multicolumn{4}{c}{Qwen} & \multicolumn{4}{c}{DeepSeek} & \multicolumn{4}{c}{GLM} \\
\cmidrule(lr){2-5}\cmidrule(lr){6-9}\cmidrule(lr){10-13}
Scenario & A & B & Both & Either & A & B & Both & Either & A & B & Both & Either \\
\midrule
Oracle & \PassCell{75} & \PassCell{76} & \PassCell{66} & \PassCell{89} & \PassCell{80} & \PassCell{76} & \PassCell[\bfseries]{68} & \PassCell{92} & \PassCell{67} & \PassCell{74} & \PassCell[\bfseries]{60} & \PassCell{85} \\
Orchestrator & \PassCell{60} & \PassCell{69} & \PassCell{49} & \PassCell{82} & \PassCell{72} & \PassCell{75} & \PassCell{60} & \PassCell{91} & \PassCell{52} & \PassCell{72} & \PassCell{44} & \PassCell{82} \\
Comm-Oracle & \PassCell{78} & \PassCell{73} & \PassCell{60} & \PassCell{92} & \PassCell{72} & \PassCell{73} & \PassCell{61} & \PassCell{88} & \PassCell{62} & \PassCell{75} & \PassCell{33} & \PassCell{89} \\
Oracle + Gold Tests & \PassCell[\bfseries]{99} & \PassCell[\bfseries]{91} & \PassCell[\bfseries]{98} & \PassCell[\bfseries]{99} & \PassCell[\bfseries]{100} & \PassCell[\bfseries]{89} & \PassCell[\bfseries]{99} & \PassCell[\bfseries]{100} & \PassCell[\bfseries]{96} & \PassCell[\bfseries]{91} & \PassCell[\bfseries]{94} & \PassCell[\bfseries]{100} \\
Parallel + Gold Tests & \PassCell[\bfseries]{100} & \PassCell[\bfseries]{99} & \PassCell{29} & \PassCell[\bfseries]{100} & \PassCell[\bfseries]{100} & \PassCell[\bfseries]{100} & \PassCell{33} & \PassCell[\bfseries]{100} & \PassCell[\bfseries]{98} & \PassCell[\bfseries]{100} & \PassCell{39} & \PassCell[\bfseries]{100} \\
\midrule
Sequential A$\to$B (Repo) & \PassCell{80} & \PassCell{84} & \PassCell[\bfseries]{76} & \PassCell{91} & \PassCell{69} & \PassCell{79} & \PassCell{66} & \PassCell{89} & \PassCell{59} & \PassCell{76} & \PassCell{55} & \PassCell{82} \\
Sequential A$\to$B (Hist) & \PassCell{73} & \PassCell{75} & \PassCell{60} & \PassCell{91} & \PassCell{66} & \PassCell{81} & \PassCell{62} & \PassCell{88} & \PassCell{65} & \PassCell{80} & \PassCell{59} & \PassCell{88} \\
Sequential B$\to$A (Repo) & \PassCell{82} & \PassCell{48} & \PassCell{44} & \PassCell{88} & \PassCell{85} & \PassCell{54} & \PassCell{52} & \PassCell{92} & \PassCell{67} & \PassCell{66} & \PassCell{54} & \PassCell{84} \\
Sequential B$\to$A (Hist) & \PassCell{84} & \PassCell{55} & \PassCell{47} & \PassCell{92} & \PassCell{88} & \PassCell{48} & \PassCell{42} & \PassCell{95} & \PassCell{73} & \PassCell{62} & \PassCell{52} & \PassCell{87} \\
\midrule
Reentrant ABAB (Repo) & \PassCell{72} & \PassCell{85} & \PassCell{66} & \PassCell{93} & \PassCell{59} & \PassCell{81} & \PassCell{53} & \PassCell{89} & \PassCell{60} & \PassCell{75} & \PassCell{53} & \PassCell{84} \\
Reentrant ABAB (Hist) & \PassCell{75} & \PassCell{84} & \PassCell{67} & \PassCell{94} & \PassCell{68} & \PassCell{79} & \PassCell{59} & \PassCell{89} & \PassCell{64} & \PassCell{80} & \PassCell[\bfseries]{60} & \PassCell{87} \\
Reentrant BABA (Repo) & \PassCell{84} & \PassCell{36} & \PassCell{31} & \PassCell{91} & \PassCell{81} & \PassCell{60} & \PassCell{53} & \PassCell{91} & \PassCell{67} & \PassCell{67} & \PassCell{54} & \PassCell{84} \\
Reentrant BABA (Hist) & \PassCell{86} & \PassCell{56} & \PassCell{54} & \PassCell{91} & \PassCell{85} & \PassCell{56} & \PassCell{55} & \PassCell{91} & \PassCell{74} & \PassCell{61} & \PassCell{51} & \PassCell{87} \\
\midrule
Parallel Unaware & \PassCell{85} & \PassCell{87} & \PassCell{20} & \PassCell{96} & \PassCell{89} & \PassCell{88} & \PassCell{26} & \PassCell{96} & \PassCell{74} & \PassCell{85} & \PassCell{25} & \PassCell{93} \\
Parallel Aware & \PassCell{87} & \PassCell{85} & \PassCell{20} & \PassCell[\bfseries]{99} & \PassCell{87} & \PassCell{87} & \PassCell{25} & \PassCell{98} & \PassCell{64} & \PassCell{81} & \PassCell{20} & \PassCell{89} \\
Parallel Message & \PassCell{80} & \PassCell{87} & \PassCell{22} & \PassCell{95} & \PassCell[\bfseries]{91} & \PassCell{85} & \PassCell{29} & \PassCell{96} & \PassCell{68} & \PassCell{87} & \PassCell{29} & \PassCell{91} \\
Parallel Sync & \PassCell{76} & \PassCell{82} & \PassCell{27} & \PassCell{92} & \PassCell{84} & \PassCell{81} & \PassCell{39} & \PassCell{96} & \PassCell{61} & \PassCell{86} & \PassCell{27} & \PassCell{93} \\
Parallel Message (Redis) & \PassCell{87} & \PassCell{80} & \PassCell{31} & \PassCell{95} & \PassCell{87} & \PassCell{82} & \PassCell{25} & \PassCell[\bfseries]{99} & \PassCell{69} & \PassCell{87} & \PassCell{25} & \PassCell[\bfseries]{94} \\
Parallel Sync (Redis) & \PassCell{86} & \PassCell{86} & \PassCell{26} & \PassCell{96} & \PassCell{86} & \PassCell{81} & \PassCell{40} & \PassCell{98} & \PassCell{66} & \PassCell{79} & \PassCell{29} & \PassCell{92} \\
\bottomrule
\end{tabular}
}
\end{table}

\clearpage 

\section{Merge policy}
\label{appendix:merge-policy}

\cref{table:merge-policy-opus} breaks down parallel pass rates by merge policy for \claudeopus{}, and \cref{table:merge-policy-significance} reports paired significance tests between merge policies for \claudeopus{} on the parallel scenarios.

\cref{table:configs-merge-policy} gives the same breakdown for the open-weight configurations, averaged over the six configurations, with the number of configurations in which each step of the ladder changes outcomes significantly (exact binomial test on the instances it adds and loses, $p<0.05$).

\begin{table}[htp]
\centering
\small
\setlength{\tabcolsep}{5pt}
\caption{Pass rate (\%) of \claudeopus{} on the parallel scenarios, broken out by merge
policy: \emph{concat} (concatenate the two agents' diffs), \emph{git-merge},
\emph{reconciler} (a post-hoc merger model), and \emph{any} (best of the three, i.e). Per column, the
highest and second-highest values are in \textbf{bold} (ties shared).}
\label{table:merge-policy-opus}
\begin{tabular}{l rrrr rrrr}
\toprule
 & \multicolumn{4}{c}{Contradictory} & \multicolumn{4}{c}{Complementary} \\
\cmidrule(lr){2-5}\cmidrule(lr){6-9}
Scenario & concat & git-merge & reconciler & any & concat & git-merge & reconciler & any \\
\midrule
Parallel Unaware & 10 & 14 & 16 & 16 & 51 & 54 & 54 & 57 \\
Parallel Aware & 16 & 18 & 26 & 26 & 46 & 49 & 57 & 60 \\
Parallel Message &16 & 14 & 20 & 20 & 51 & 57 & 63 & 63 \\
Parallel Sync &16 & 16 & 22 & 22 & 60 & \textbf{69} & 71 & 71 \\
Parallel Message (Redis) & 16 & 20 & 30 & 30 & 60 & 57 & 63 & 69 \\
Parallel Sync (Redis) & \textbf{22} & \textbf{22} & \textbf{32} & \textbf{32} & \textbf{63} & \textbf{69} & \textbf{77} & \textbf{77} \\
Parallel Comm-Oracle & \textbf{38} & \textbf{50} & \textbf{56} & \textbf{56} & \textbf{71} & \textbf{80} & \textbf{80} & \textbf{83} \\
\bottomrule
\end{tabular}
\end{table}

\begin{table}[htp]
\centering
\small
\setlength{\tabcolsep}{4pt}
\caption{Paired comparison of merge policies for \claudeopus{} on the parallel scenarios. \emph{Passed} gives the number of instances each policy passes; \emph{added} gives the instances gained or lost by git-merge over concatenation and by the reconciler over git-merge, with $p$ from a binomial test on those instances. $^{*}$: $p<0.05$.}
\label{table:merge-policy-significance}
\begin{tabular}{l rrr rr rr}
\toprule
 & \multicolumn{3}{c}{Instances passed} & \multicolumn{2}{c}{git-merge $-$ concat} & \multicolumn{2}{c}{reconciler $-$ git-merge} \\
\cmidrule(lr){2-4}\cmidrule(lr){5-6}\cmidrule(lr){7-8}
Scenario & concat & git-merge & reconciler & added & $p$ & added & $p$ \\
\midrule
\multicolumn{8}{l}{\textit{Contradictory} ($n=50$)} \\
Parallel Unaware & 5 & 7 & 8 & +2 & 0.500 & +1 & 1.000 \\
Parallel Aware & 8 & 9 & 13 & +1 & 1.000 & +4 & 0.125 \\
Parallel Message & 8 & 7 & 10 & +0/$-$1 & 1.000 & +3 & 0.250 \\
Parallel Sync & 8 & 8 & 11 & +0 & 1.000 & +3 & 0.250 \\
Parallel Message (Redis) & 8 & 10 & 15 & +2 & 0.500 & +5 & 0.062 \\
Parallel Sync (Redis) & 11 & 11 & 16 & +0 & 1.000 & +5 & 0.062 \\
Communication Oracle & 19 & 25 & 28 & +6 & 0.031$^{*}$ & +3 & 0.250 \\
\midrule
\multicolumn{8}{l}{\textit{Complementary} ($n=35$)} \\
Parallel Unaware & 18 & 19 & 19 & +2/$-$1 & 1.000 & +0 & 1.000 \\
Parallel Aware & 16 & 17 & 20 & +2/$-$1 & 1.000 & +3 & 0.250 \\
Parallel Message & 18 & 20 & 22 & +2 & 0.500 & +2 & 0.500 \\
Parallel Sync & 21 & 24 & 25 & +3 & 0.250 & +1 & 1.000 \\
Parallel Message (Redis) & 21 & 20 & 22 & +1/$-$2 & 1.000 & +2 & 0.500 \\
Parallel Sync (Redis) & 22 & 24 & 27 & +2 & 0.500 & +3 & 0.250 \\
Communication Oracle & 25 & 28 & 28 & +3 & 0.250 & +1/$-$1 & 1.000 \\
\bottomrule
\end{tabular}
\end{table}

\begin{table}[htp]
\centering
\setlength{\tabcolsep}{3pt}
\caption{Pass rate (\%) of the open-weight configurations on the parallel scenarios by merge policy, averaged over the six configurations: \emph{concat}, \emph{git-merge}, \emph{reconciler}, and \emph{any} (best of the three). \emph{g$>$c} and \emph{r$>$g}: in how many of the six configurations git-merge differs significantly from concatenation, and the reconciler from git-merge.}
\label{table:configs-merge-policy}
\resizebox{\textwidth}{!}{
\begin{tabular}{l rrrr rr rrrr rr}
\toprule
 & \multicolumn{6}{c}{Contradictory} & \multicolumn{6}{c}{Complementary} \\
\cmidrule(lr){2-7}\cmidrule(lr){8-13}
Scenario & concat & git-merge & reconciler & any & g$>$c & r$>$g & concat & git-merge & reconciler & any & g$>$c & r$>$g \\
\midrule
Parallel Unaware & 7 & 9 & 19 & 19 & 0/6 & 4/6 & 41 & 44 & 52 & 53 & 0/6 & 4/6 \\
Parallel Aware & 8 & 9 & 18 & 19 & 0/6 & 5/6 & 48 & 50 & 57 & 58 & 0/6 & 2/6 \\
Parallel Message & 7 & 10 & 24 & 24 & 0/6 & 6/6 & 55 & 58 & 65 & 66 & 0/6 & 3/6 \\
Parallel Sync & 9 & 13 & 27 & 27 & 2/6 & 5/6 & 52 & 55 & 62 & 64 & 0/6 & 3/6 \\
Parallel Message (Redis) & 8 & 11 & 23 & 23 & 0/6 & 4/6 & 51 & 55 & 60 & 61 & 1/6 & 1/6 \\
Parallel Sync (Redis) & 10 & 14 & 28 & 28 & 1/6 & 6/6 & 51 & 55 & 62 & 64 & 0/6 & 2/6 \\
Comm-Oracle & 25 & 36 & 47 & 48 & 4/6 & 5/6 & 62 & 67 & 71 & 71 & 1/6 & 1/6 \\
\bottomrule
\end{tabular}
}
\end{table}

\clearpage 

\section{Performance by repository}

\cref{table:per-repo-performance-opus} reports per-repository oracle and mean pass rates for \claudeopus{}, broken down by repository, language, and family, and \cref{table:configs-per-repo} reports the same for the open-weight configurations, averaged over the six configurations.

\begin{table}[htp]
\centering
\small
\setlength{\tabcolsep}{6pt}
\caption{Per-repository performance of \claudeopus{} on its 85 instances (\swepro{} v1).
$n$ is the number of instances; \emph{Oracle} is the single-agent oracle pass rate (\%),
the per-repo ceiling; \emph{Mean} is the mean pass rate (\%) across all scenarios for that
repository and family.}
\label{table:per-repo-performance-opus}
\begin{tabular}{l l rrr rrr}
\toprule
 & & \multicolumn{3}{c}{Contradictory} & \multicolumn{3}{c}{Complementary} \\
\cmidrule(lr){3-5}\cmidrule(lr){6-8}
Repo & Lang & $n$ & Oracle & Mean & $n$ & Oracle & Mean \\
\midrule
flipt & Go & 4 & 75 & 41 & 2 & 100 & 83 \\
navidrome & Go & 6 & 50 & 39 & 7 & 100 & 87 \\
teleport & Go & 1 & 100 & 37 & 2 & 100 & 83 \\
vuls & Go & 5 & 60 & 41 & 4 & 75 & 65 \\
\midrule
ansible & Python & 9 & 67 & 54 & 3 & 100 & 85 \\
openlibrary & Python & 3 & 67 & 60 & 4 & 50 & 72 \\
qutebrowser & Python & 6 & 67 & 48 & 2 & 100 & 81 \\
\midrule
NodeBB & JS/TS & 8 & 75 & 48 & 5 & 100 & 89 \\
element-web & JS/TS & 3 & 33 & 39 & 5 & 80 & 61 \\
webclients & JS/TS & 5 & 40 & 33 & 1 & 0 & 6 \\
\midrule
All & & 50 & 62 & 45 & 35 & 86 & 76 \\
\bottomrule
\end{tabular}
\end{table}

\begin{table}[htp]
\centering
\small
\setlength{\tabcolsep}{6pt}
\caption{Per-repository performance of the open-weight configurations on all 155 instances, averaged over the six configurations. $n$ is the number of instances; \emph{Oracle} is the single-agent oracle pass rate (\%); \emph{Mean} is the mean pass rate (\%) across all scenarios for that repository and family.}
\label{table:configs-per-repo}
\begin{tabular}{l l rrr rrr}
\toprule
 & & \multicolumn{3}{c}{Contradictory} & \multicolumn{3}{c}{Complementary} \\
\cmidrule(lr){3-5}\cmidrule(lr){6-8}
Repo & Lang & $n$ & Oracle & Mean & $n$ & Oracle & Mean \\
\midrule
flipt & Go & 8 & 40 & 27 & 6 & 36 & 47 \\
vuls & Go & 9 & 52 & 38 & 8 & 79 & 72 \\
teleport & Go & 4 & 54 & 52 & 6 & 72 & 61 \\
navidrome & Go & 9 & 74 & 44 & 11 & 80 & 74 \\
\midrule
ansible & Python & 13 & 81 & 58 & 8 & 94 & 89 \\
openlibrary & Python & 6 & 47 & 38 & 7 & 60 & 61 \\
qutebrowser & Python & 9 & 59 & 44 & 5 & 83 & 65 \\
\midrule
element-web & JS/TS & 7 & 43 & 34 & 9 & 80 & 72 \\
NodeBB & JS/TS & 12 & 65 & 45 & 8 & 60 & 56 \\
webclients & JS/TS & 7 & 64 & 51 & 2 & 33 & 34 \\
tutanota & JS/TS & 1 & 17 & 6 & 0 & -- & -- \\
\midrule
All &  & 85 & 60 & 43 & 70 & 72 & 67 \\
\bottomrule
\end{tabular}
\end{table}

\clearpage 

\section{Absolute Token numbers}
\label{appendix:raw-token-counts}

\cref{table:token-numbers-opus} reports the mean per-run token counts (input, cache-read, output, cache-write and effective) for \claudeopus{} across all scenarios, and \cref{table:configs-token-numbers} reports the mean effective tokens per run for the six open-weight configurations. Input, cache-read, output, cache-write and reasoning counts per configuration are in the released data.

\begin{table}[htp]
\centering
\small
\setlength{\tabcolsep}{3pt}
\caption{Mean per-run token counts (thousands) for \claudeopus{} on its 85 instances (\swepro{} v1; \cref{sec:methods}, \cref{appendix:openrouter-pricing}).}
\label{table:token-numbers-opus}
\resizebox{\textwidth}{!}{
\begin{tabular}{l rrrrr rrrrr}
\toprule
 & \multicolumn{5}{c}{Contradictory} & \multicolumn{5}{c}{Complementary} \\
\cmidrule(lr){2-6}\cmidrule(lr){7-11}
Scenario & Input & Cache Read & Output & Cache Write & Effective & Input & Cache Read & Output & Cache Write & Effective \\
\midrule
\multicolumn{11}{l}{\textit{Reference}} \\
Oracle & 1.4 & 854.8 & 14.6 & 37.7 & 197.6 & 0.1 & 2525.4 & 26.7 & 72.4 & 458.4 \\
Orchestrator & 1.7 & 980.2 & 25.6 & 84.1 & 312.0 & 0.1 & 2400.5 & 35.6 & 123.5 & 541.7 \\
Comm-Oracle & 1.2 & 3008.6 & 32.4 & 77.6 & 541.8 & 0.1 & 3471.9 & 33.0 & 86.1 & 598.5 \\
Oracle + Gold Tests & 1.6 & 666.2 & 9.9 & 29.2 & 147.0 & 1.5 & 1223.5 & 15.8 & 42.6 & 245.7 \\
Parallel + Gold Tests & 2.7 & 642.2 & 9.7 & 36.5 & 152.2 & n/a & n/a & n/a & n/a & n/a \\
\midrule
\multicolumn{11}{l}{\textit{Sequential}} \\
\quad A$\to$B (Repo) & 2.4 & 969.0 & 16.6 & 51.1 & 233.3 & 1.5 & 2744.7 & 28.9 & 91.5 & 511.8 \\
\quad A$\to$B (Hist) & 2.4 & 1027.7 & 17.7 & 52.3 & 246.1 & 0.8 & 2918.2 & 29.0 & 88.1 & 525.5 \\
\quad B$\to$A (Repo) & 2.4 & 1124.3 & 17.4 & 49.0 & 250.8 & 0.2 & 2816.8 & 29.1 & 81.9 & 509.4 \\
\quad B$\to$A (Hist) & 2.6 & 1116.4 & 17.2 & 48.7 & 249.0 & 0.5 & 2923.1 & 27.8 & 76.5 & 508.0 \\
\midrule
\multicolumn{11}{l}{\textit{Reentrant}} \\
\quad ABAB (Repo) & 4.8 & 1506.4 & 26.6 & 83.6 & 372.0 & 0.2 & 3587.6 & 39.1 & 127.1 & 681.7 \\
\quad ABAB (Hist) & 4.9 & 1656.6 & 29.1 & 89.1 & 405.3 & 4.5 & 3784.1 & 38.6 & 118.3 & 694.3 \\
\quad BABA (Repo) & 4.8 & 1591.9 & 26.8 & 82.2 & 380.4 & 2.2 & 3675.2 & 38.9 & 127.9 & 691.9 \\
\quad BABA (Hist) & 6.2 & 1658.8 & 28.1 & 84.1 & 396.6 & 1.0 & 3919.0 & 40.8 & 121.9 & 718.5 \\
\midrule
\multicolumn{11}{l}{\textit{Parallel}} \\
\quad Unaware & 2.3 & 966.2 & 16.4 & 50.5 & 231.6 & 0.3 & 2803.3 & 30.4 & 88.5 & 521.0 \\
\quad Aware & 3.2 & 939.2 & 15.9 & 48.7 & 225.1 & 0.3 & 3073.8 & 31.4 & 89.2 & 553.8 \\
\quad Message & 2.5 & 1297.0 & 19.3 & 52.2 & 280.8 & 1.6 & 3866.3 & 34.9 & 92.0 & 654.9 \\
\quad Sync & 2.7 & 1434.1 & 23.2 & 60.8 & 322.7 & 0.3 & 3813.3 & 35.6 & 94.1 & 653.5 \\
\quad Message (Redis) & 2.5 & 1481.8 & 22.3 & 55.9 & 318.0 & 1.5 & 3501.0 & 32.5 & 89.2 & 603.3 \\
\quad Sync (Redis) & 2.7 & 2029.1 & 32.4 & 72.4 & 440.1 & 0.9 & 3601.9 & 36.8 & 93.0 & 637.8 \\
\bottomrule
\end{tabular}}
\end{table}

\begin{table}[htp]
\centering
\small
\setlength{\tabcolsep}{3pt}
\caption{Mean effective tokens per run (thousands) for the six open-weight configurations (fresh input $+$ $0.1\cdot$cache-read $+$ $5\cdot$output; \cref{sec:methods}). Claude Code connected to our model server reports no cache reads, so its input counts as fresh; absolute numbers are therefore not comparable across harnesses.}
\label{table:configs-token-numbers}
\resizebox{\textwidth}{!}{
\begin{tabular}{l rrrrrr rrrrrr}
\toprule
 & \multicolumn{6}{c}{Contradictory} & \multicolumn{6}{c}{Complementary} \\
\cmidrule(lr){2-7}\cmidrule(lr){8-13}
 & \multicolumn{3}{c}{Claude Code} & \multicolumn{3}{c}{Codex} & \multicolumn{3}{c}{Claude Code} & \multicolumn{3}{c}{Codex} \\
\cmidrule(lr){2-4}\cmidrule(lr){5-7}\cmidrule(lr){8-10}\cmidrule(lr){11-13}
Scenario & Qwen & DeepSeek & GLM & Qwen & DeepSeek & GLM & Qwen & DeepSeek & GLM & Qwen & DeepSeek & GLM \\
\midrule
\multicolumn{13}{l}{\textit{Reference}} \\
Oracle & 2219 & 4835 & 5498 & 750 & 626 & 444 & 3081 & 6293 & 7013 & 856 & 770 & 536 \\
Orchestrator & 3018 & 3727 & 4039 & 315 & 275 & 116 & 2889 & 4306 & 4366 & 343 & 288 & 137 \\
Comm-Oracle & 3395 & 8210 & 7239 & 1286 & 1213 & 584 & 3453 & 7858 & 7497 & 1202 & 1011 & 692 \\
Oracle + Gold Tests & 1418 & 2815 & 3547 & 531 & 355 & 235 & 2278 & 4454 & 5345 & 733 & 514 & 353 \\
Parallel + Gold Tests & 1703 & 3221 & 3354 & 597 & 378 & 309 & n/a & n/a & n/a & n/a & n/a & n/a \\
\midrule
\multicolumn{13}{l}{\textit{Sequential}} \\
\quad A$\to$B (Repo) & 2536 & 5390 & 5082 & 811 & 639 & 474 & 3018 & 5604 & 5719 & 1009 & 710 & 537 \\
\quad A$\to$B (Hist) & 2395 & 5287 & 4948 & 897 & 647 & 493 & 2997 & 6201 & 5505 & 983 & 717 & 519 \\
\quad B$\to$A (Repo) & 2675 & 5901 & 5462 & 928 & 696 & 453 & 3628 & 6640 & 7925 & 1092 & 839 & 627 \\
\quad B$\to$A (Hist) & 2286 & 5521 & 5286 & 951 & 616 & 524 & 3332 & 7103 & 7126 & 1118 & 864 & 623 \\
\midrule
\multicolumn{13}{l}{\textit{Reentrant}} \\
\quad ABAB (Repo) & 4510 & 10171 & 7721 & 1776 & 1310 & 764 & 4470 & 10968 & 9831 & 1987 & 1404 & 871 \\
\quad ABAB (Hist) & 3764 & 9608 & 7820 & 1680 & 1237 & 758 & 4343 & 10899 & 10272 & 1983 & 1430 & 870 \\
\quad BABA (Repo) & 4806 & 10872 & 9536 & 1891 & 1387 & 764 & 4928 & 12086 & 10874 & 2196 & 1516 & 942 \\
\quad BABA (Hist) & 3850 & 9834 & 9018 & 1783 & 1253 & 868 & 4694 & 11474 & 9829 & 2119 & 1541 & 1001 \\
\midrule
\multicolumn{13}{l}{\textit{Parallel}} \\
\quad Unaware & 2384 & 5161 & 5160 & 782 & 574 & 467 & 3166 & 6910 & 6610 & 1001 & 773 & 571 \\
\quad Aware & 2368 & 4637 & 5023 & 784 & 561 & 463 & 2993 & 6048 & 6126 & 995 & 774 & 663 \\
\quad Message & 3107 & 6759 & 6128 & 1106 & 828 & 492 & 3718 & 7911 & 6908 & 1178 & 917 & 578 \\
\quad Sync & 3486 & 8053 & 7065 & 1233 & 1087 & 543 & 4229 & 8097 & 6575 & 1236 & 996 & 794 \\
\quad Message (Redis) & 2986 & 7063 & 6362 & 1120 & 806 & 571 & 3805 & 7733 & 7062 & 1119 & 890 & 609 \\
\quad Sync (Redis) & 3652 & 8262 & 6744 & 1210 & 968 & 558 & 3919 & 7914 & 7621 & 1282 & 973 & 643 \\
\bottomrule
\end{tabular}
}
\end{table}

\clearpage

\section{Paired Significance Tests On Scenarios}
\label{appendix:significance}

\Cref{table:significance} reports an exact McNemar test on joint success and a paired Wilcoxon signed-rank test on effective token cost, both computed over the same instances for \claudeopus{} and for every scenario comparison cited in \cref{sec:results-and-analysis}. \Cref{table:configs-significance} reports the same comparisons for the six open-weight configurations.
\emph{Pass} gives the mean joint success over configurations, and $p$ is an exact McNemar test on the discordant instances summed over configurations. $k$/6 is the number of configurations in which the comparison is significant on its own ($p<0.05$). \emph{Cost} is the mean over configurations of the ratio of mean effective tokens $A/B$, and its $p$-value is a Wilcoxon signed-rank test over all (configuration, instance) pairs, with each configuration's tokens divided by its mean oracle tokens. Per-configuration results are in the released data.

\begin{table}[htp]
\centering
\small
\setlength{\tabcolsep}{5pt}
\caption{Paired significance tests for \claudeopus{} between scenarios on the same instances (contradictory $n=50$, complementary $n=35$). \emph{Pass} gives each scenario's joint success (\%); $p$ is an exact McNemar test on the instances that only one scenario passes. \emph{Cost} is the ratio of mean effective tokens $A/B$, with the $p$-value of a Wilcoxon signed-rank test paired by instance. $^{*}$: $p<0.05$; $^{**}$: $p<0.01$.}
\label{table:significance}
\begin{tabular}{l rr r rr}
\toprule
 & \multicolumn{2}{c}{Pass (\%)} & & \multicolumn{2}{c}{Cost} \\
\cmidrule(lr){2-3}\cmidrule(lr){5-6}
Comparison ($A$ vs.\ $B$) & $A$ & $B$ & $p$ & $A/B$ & $p$ \\
\midrule
\multicolumn{6}{l}{\textit{Contradictory}} \\
single agent (oracle) vs parallel unaware & 62 & 16 & $<$0.001$^{**}$ & 0.85 & 0.005$^{**}$ \\
single agent (oracle) + gold vs parallel + gold & 98 & 38 & $<$0.001$^{**}$ & 0.97 & 0.002$^{**}$ \\
sequential A$\to$B (repo) vs parallel unaware & 54 & 16 & $<$0.001$^{**}$ & 1.01 & 0.818 \\
reentrant ABAB (repo) vs sequential A$\to$B (repo) & 50 & 54 & 0.727 & 1.59 & $<$0.001$^{**}$ \\
sequential B$\to$A vs A$\to$B (repo) & 50 & 54 & 0.754 & 1.08 & 0.553 \\
sequential A$\to$B hist vs repo & 50 & 54 & 0.727 & 1.06 & 0.163 \\
orchestrator vs single agent (oracle) & 58 & 62 & 0.688 & 1.58 & $<$0.001$^{**}$ \\
parallel aware vs unaware & 26 & 16 & 0.180 & 0.97 & 0.484 \\
parallel message vs aware & 20 & 26 & 0.453 & 1.25 & $<$0.001$^{**}$ \\
parallel sync vs aware & 22 & 26 & 0.688 & 1.43 & $<$0.001$^{**}$ \\
parallel sync vs message & 22 & 20 & 1.000 & 1.15 & $<$0.001$^{**}$ \\
parallel message redis vs message (file) & 30 & 20 & 0.180 & 1.13 & 0.006$^{**}$ \\
parallel sync redis vs sync (file) & 32 & 22 & 0.125 & 1.36 & $<$0.001$^{**}$ \\
parallel sync redis vs unaware & 32 & 16 & 0.039$^{*}$ & 1.90 & $<$0.001$^{**}$ \\
comm.\ oracle vs parallel sync redis & 56 & 32 & 0.002$^{**}$ & 1.23 & 0.001$^{**}$ \\
comm.\ oracle vs sequential A$\to$B (repo) & 56 & 54 & 1.000 & 2.32 & $<$0.001$^{**}$ \\
single agent (oracle) vs comm.\ oracle & 62 & 56 & 0.508 & 0.36 & $<$0.001$^{**}$ \\
single agent (oracle) vs sequential A$\to$B (repo) & 62 & 54 & 0.344 & 0.85 & 0.002$^{**}$ \\
\midrule
\multicolumn{6}{l}{\textit{Complementary}} \\
single agent (oracle) vs parallel unaware & 86 & 57 & 0.021$^{*}$ & 0.88 & 0.017$^{*}$ \\
sequential A$\to$B (repo) vs parallel unaware & 77 & 57 & 0.065 & 0.98 & 0.777 \\
reentrant ABAB (repo) vs sequential A$\to$B (repo) & 80 & 77 & 1.000 & 1.33 & $<$0.001$^{**}$ \\
sequential B$\to$A vs A$\to$B (repo) & 71 & 77 & 0.774 & 1.00 & 0.827 \\
sequential A$\to$B hist vs repo & 77 & 77 & 1.000 & 1.03 & 0.104 \\
orchestrator vs single agent (oracle) & 77 & 86 & 0.375 & 1.18 & 0.023$^{*}$ \\
parallel aware vs unaware & 60 & 57 & 1.000 & 1.06 & 0.245 \\
parallel message vs aware & 63 & 60 & 1.000 & 1.18 & 0.006$^{**}$ \\
parallel sync vs aware & 71 & 60 & 0.344 & 1.18 & 0.001$^{**}$ \\
parallel sync vs message & 71 & 63 & 0.453 & 1.00 & 0.326 \\
parallel message redis vs message (file) & 69 & 63 & 0.754 & 0.92 & 0.037$^{*}$ \\
parallel sync redis vs sync (file) & 77 & 71 & 0.625 & 0.98 & 0.441 \\
parallel sync redis vs unaware & 77 & 57 & 0.065 & 1.22 & 0.003$^{**}$ \\
comm.\ oracle vs parallel sync redis & 83 & 77 & 0.727 & 0.94 & 0.179 \\
comm.\ oracle vs sequential A$\to$B (repo) & 83 & 77 & 0.727 & 1.17 & 0.004$^{**}$ \\
single agent (oracle) vs comm.\ oracle & 86 & 83 & 1.000 & 0.77 & $<$0.001$^{**}$ \\
single agent (oracle) vs sequential A$\to$B (repo) & 86 & 77 & 0.375 & 0.90 & 0.107 \\
\bottomrule
\end{tabular}
\end{table}

\begin{table}[htp]
\centering
\small
\setlength{\tabcolsep}{4pt}
\caption{Paired significance tests for the six open-weight configurations between scenarios on the same instances (contradictory $n=85$, complementary $n=70$ per configuration). $k$/6 is the number of configurations in which the comparison is significant on its own. $^{*}$: $p<0.05$; $^{**}$: $p<0.01$.}
\label{table:configs-significance}
\resizebox{\textwidth}{!}{
\begin{tabular}{l rr rr r rr}
\toprule
 & \multicolumn{4}{c}{Pass} & & \multicolumn{2}{c}{Cost} \\
\cmidrule(lr){2-5}\cmidrule(lr){7-8}
Comparison ($A$ vs.\ $B$) & $A$ & $B$ & $p$ & $k$/6 & $A/B$ & $p$ & $k$/6 \\
\midrule
\multicolumn{8}{l}{\textit{Contradictory}} \\
single agent (oracle) vs parallel unaware & 60 & 19 & $<$0.001$^{**}$ & 6/6 & 0.99 & 0.392 & 0/6 \\
single agent (oracle) + gold vs parallel + gold & 96 & 32 & $<$0.001$^{**}$ & 6/6 & 0.89 & $<$0.001$^{**}$ & 3/6 \\
sequential A$\to$B (repo) vs parallel unaware & 60 & 19 & $<$0.001$^{**}$ & 6/6 & 1.04 & 0.021$^{*}$ & 0/6 \\
reentrant ABAB (repo) vs sequential A$\to$B (repo) & 52 & 60 & $<$0.001$^{**}$ & 1/6 & 1.84 & $<$0.001$^{**}$ & 6/6 \\
sequential B$\to$A vs A$\to$B (repo) & 46 & 60 & $<$0.001$^{**}$ & 4/6 & 1.07 & $<$0.001$^{**}$ & 2/6 \\
sequential A$\to$B hist vs repo & 53 & 60 & $<$0.001$^{**}$ & 2/6 & 1.01 & 0.483 & 0/6 \\
orchestrator vs single agent (oracle) & 46 & 60 & $<$0.001$^{**}$ & 4/6 & 0.66 & $<$0.001$^{**}$ & 6/6 \\
parallel aware vs unaware & 19 & 19 & 0.890 & 0/6 & 0.97 & 0.939 & 0/6 \\
parallel message vs aware & 24 & 19 & 0.002$^{**}$ & 1/6 & 1.32 & $<$0.001$^{**}$ & 6/6 \\
parallel sync vs aware & 27 & 19 & $<$0.001$^{**}$ & 2/6 & 1.55 & $<$0.001$^{**}$ & 6/6 \\
parallel sync vs message & 27 & 24 & 0.111 & 0/6 & 1.17 & $<$0.001$^{**}$ & 3/6 \\
parallel message redis vs message (file) & 23 & 24 & 0.567 & 0/6 & 1.03 & 0.495 & 0/6 \\
parallel sync redis vs sync (file) & 28 & 27 & 1.000 & 0/6 & 0.99 & 0.893 & 0/6 \\
parallel sync redis vs unaware & 28 & 19 & $<$0.001$^{**}$ & 2/6 & 1.48 & $<$0.001$^{**}$ & 6/6 \\
comm.\ oracle vs parallel sync redis & 48 & 28 & $<$0.001$^{**}$ & 5/6 & 1.06 & 0.252 & 1/6 \\
comm.\ oracle vs sequential A$\to$B (repo) & 48 & 60 & $<$0.001$^{**}$ & 3/6 & 1.50 & $<$0.001$^{**}$ & 6/6 \\
single agent (oracle) vs comm.\ oracle & 60 & 48 & $<$0.001$^{**}$ & 3/6 & 0.64 & $<$0.001$^{**}$ & 6/6 \\
single agent (oracle) vs sequential A$\to$B (repo) & 60 & 60 & 1.000 & 0/6 & 0.95 & 0.003$^{**}$ & 3/6 \\
\midrule
\multicolumn{8}{l}{\textit{Complementary}} \\
single agent (oracle) vs parallel unaware & 72 & 53 & $<$0.001$^{**}$ & 4/6 & 0.96 & 0.003$^{**}$ & 2/6 \\
sequential A$\to$B (repo) vs parallel unaware & 71 & 53 & $<$0.001$^{**}$ & 6/6 & 0.92 & $<$0.001$^{**}$ & 3/6 \\
reentrant ABAB (repo) vs sequential A$\to$B (repo) & 70 & 71 & 0.583 & 0/6 & 1.79 & $<$0.001$^{**}$ & 6/6 \\
sequential B$\to$A vs A$\to$B (repo) & 63 & 71 & $<$0.001$^{**}$ & 1/6 & 1.20 & $<$0.001$^{**}$ & 5/6 \\
sequential A$\to$B hist vs repo & 69 & 71 & 0.341 & 0/6 & 1.00 & 0.465 & 0/6 \\
orchestrator vs single agent (oracle) & 59 & 72 & $<$0.001$^{**}$ & 3/6 & 0.55 & $<$0.001$^{**}$ & 5/6 \\
parallel aware vs unaware & 58 & 53 & 0.020$^{*}$ & 2/6 & 0.98 & 0.458 & 1/6 \\
parallel message vs aware & 66 & 58 & $<$0.001$^{**}$ & 2/6 & 1.15 & $<$0.001$^{**}$ & 5/6 \\
parallel sync vs aware & 64 & 58 & 0.008$^{**}$ & 1/6 & 1.26 & $<$0.001$^{**}$ & 4/6 \\
parallel sync vs message & 64 & 66 & 0.470 & 0/6 & 1.10 & 0.004$^{**}$ & 1/6 \\
parallel message redis vs message (file) & 61 & 66 & 0.022$^{*}$ & 0/6 & 1.00 & 0.895 & 0/6 \\
parallel sync redis vs sync (file) & 64 & 64 & 0.917 & 0/6 & 0.98 & 0.540 & 1/6 \\
parallel sync redis vs unaware & 64 & 53 & $<$0.001$^{**}$ & 3/6 & 1.20 & $<$0.001$^{**}$ & 6/6 \\
comm.\ oracle vs parallel sync redis & 71 & 64 & $<$0.001$^{**}$ & 1/6 & 0.99 & 0.021$^{*}$ & 2/6 \\
comm.\ oracle vs sequential A$\to$B (repo) & 71 & 71 & 0.913 & 0/6 & 1.29 & $<$0.001$^{**}$ & 5/6 \\
single agent (oracle) vs comm.\ oracle & 72 & 71 & 1.000 & 0/6 & 0.81 & $<$0.001$^{**}$ & 5/6 \\
single agent (oracle) vs sequential A$\to$B (repo) & 72 & 71 & 0.813 & 0/6 & 1.05 & 0.614 & 3/6 \\
\bottomrule
\end{tabular}
}
\end{table}

\clearpage

\section{Details on Communication analysis}
\label{appendix:communication_analysis}
We analyze
message logs and agent transcripts from all valid \claudeopus{} runs in the communication scenarios and communication oracle over both the file and Redis transports, track when messages are read relative to the recipient's first edit, and label each conversation's content.
The content is labeled by an outcome-blind LLM judge (Claude Sonnet~5) validated on a stratified sample of 30 conversations labeled independently by an author. The judge agrees with the human labels on 90\% of conversations with Cohen's $\kappa=0.67$ which is substantial.

\textbf{Agents often send early but read late.} Communication is two-way in 94\% of runs. Agents send their first message before editing in 79\% of contradictory and 88\% of complementary turns. However,
only 67\%/68\% of the messages are ever read, 21\%/23\% are read before the recipient's first edit, and 29\%/27\% arrive after the recipient has already finished. This is the same unresponsiveness that CooperBench \citep{khatua2026cooperbench} reports, where direct questions go unanswered. Agents check for incoming messages 
mainly at the start and end of a turn and can cause \textit{mistimed communication}, and we find the main cause for coordination failure is ill-timing rather than unwillingness to reply. 

\textbf{Agents discuss the shared artifact.} The validated LLM judge finds that the shared artifact is discussed in 88\% of contradictory and 76\% of complementary non-oracle conversations. 
Agents reach an agreement in 59\% and 74\% of runs, in which both agents accept specific names, signatures, data shapes, or an ownership split.
Agreement is more frequent under \emph{Sync} than \emph{Message}: 72\% vs.\ 46\% on contradictory and 90\% vs.\ 57\% on complementary. Runs with an agreement pass at 36\% vs.\ 24\% and 75\% vs.\ 57\%, suggesting that agents need to \textit{agree on shared interface or edit ownership}, rather than merely identify \textit{where} their tasks overlap. In contrast to \citet{khatua2026cooperbench} that finds channels ``jammed with vague, ill-timed, and inaccurate messages,'' the message content is mostly on target. 

\textbf{Agreements do not prevent conflicting edits.} On contradictory tasks, 61\% of runs with a concrete agreement still produce a merge conflict, and those runs pass at only 19\%. Transcripts show agents agreeing on a design but then both editing the same function, or one agent simply mirroring the other's implementation. Their patches can still \textit{conflict with overlapping changes} even when the agents aim to implement the \textit{same design}. 

\textbf{Knowing the partner's task clarifies edit ownership.} The communication oracle reveals the partner's requirement. On complementary tasks, 91\% of oracle conversations reach agreement and only 6\% produce merge conflicts. All 48 contradictory oracle conversations reach a concrete agreement, 85\% assign ownership of the contested code, and merge conflicts fall from 65\% to 15\% of runs. In the contradictory oracle runs, assigning an owner to the contested code is associated with a higher pass rate of 61\%, compared with 29\% without one. In 15\% of runs, one agent implements both requirements while the other withdraws.
However, we note it is also the \textit{least scalable} strategy in reality because with $n$ agents it is infeasible for one agent doing all the work at company scale. 

\section{Implementation Details}
\label{appendix:details}

This section documents the model and harness configurations, communication tools, merge resolver, and token accounting used in our experiments.

\begin{itemize}[leftmargin=1em, labelsep=0.4em, itemsep=2pt, topsep=2pt]
\item \textbf{Redis documentation.}
\phantomsection\label{appendix:redis-documentation}
\url{https://redis.io/docs/latest/}

\item \textbf{Codex configuration (dataset authoring).}
\phantomsection\label{appendix:codex-configuration}
\url{https://github.com/openai/codex}, version 0.150.1, medium reasoning effort.

\item \textbf{Scout models.}
\phantomsection\label{appendix:qwen-configuration}
Qwen3.8 27B (\texttt{Qwen/Qwen3.8-27B-FP8}) served with vLLM.

\item \textbf{Merge resolver.}
\phantomsection\label{appendix:merge-resolver}
\url{https://huggingface.co/CooperBench/Qwen2.5-Coder-0.5B-Merge-Resolver}.

\item \textbf{OpenRouter pricing.}
\phantomsection\label{appendix:openrouter-pricing}
\url{https://web.archive.org/web/20260924162009/https://openrouter.ai/anthropic/claude-opus-4.8}

\item \textbf{Cross-configuration token accounting.}
\phantomsection\label{appendix:token-accounting}
For comparisons across configurations, the effective-token formula counts fresh input, cache-read, and output tokens, and excludes cache-write (cache-creation) tokens, which are not reported for every model; we exclude them for every configuration, including \claudeopus{}.
Codex reports input tokens including cache reads, so we subtract them; Claude Code connected to vLLM reports no cache reads, so all of its input counts as fresh.
The single-configuration \claudeopus{} results include cache-write tokens in the input term.

\item \textbf{Claude Code version.}
\phantomsection\label{appendix:claude-code-version}
Claude Code version 2.1.269, used for all Claude Code configurations:
\url{https://code.claude.com/docs/en/changelog#2-1-269}

\item \textbf{Codex harness.}
\phantomsection\label{appendix:codex-harness}
\url{https://github.com/openai/codex}, version 0.150.1, run non-interactively (\texttt{codex exec}) with reasoning effort \texttt{xhigh}.

\item \textbf{Open-weight models.}
\phantomsection\label{appendix:open-weight-models}
\qwen{} (\url{https://huggingface.co/Qwen/Qwen3.8-Flash-Next-FP8}), \deepseek{} (\url{https://huggingface.co/deepseek-ai/DeepSeek-V4-Flash-0731}), and \glm{} (\url{https://huggingface.co/unsloth/GLM-5.3-Flash-FP8}), each served with vLLM (\url{https://github.com/vllm-project/vllm}) from the \texttt{vllm/vllm-openai} images (\texttt{v0.28.0} for \deepseek{}; \texttt{nightly} for \qwen{} and \glm{}), with tensor parallelism over four GPUs, an FP8 KV cache, and a 1M-token context window.
Sampling uses each model's recommended settings: temperature 1.0 for all three, top-$p$ 0.95 for \qwen{} (with top-$k$ 20) and \deepseek{}, and top-$p$ 1.0 for \glm{}.
\qwen{} and \deepseek{} use speculative decoding (multi-token prediction with 3 draft tokens, and \texttt{dspark} with 7, respectively).

\end{itemize}

\section{Prompts}
\label{appendix:scenario-prompts}

Every scenario in \cref{subsec:experimental-settings} is assembled from a
\emph{base task prompt} with an optional scenario-specific \emph{overlay} appended.
Placeholders in double braces are filled per instance and scenario:
\texttt{\{\{PROBLEM\}\}}, \texttt{\{\{PROBLEM\_A\}\}}, \texttt{\{\{PROBLEM\_B\}\}}
are the requirement statements; \texttt{\{\{AGENT\}\}}/\texttt{\{\{PEER\}\}} and
\texttt{\{\{PEER\_PROBLEM\}\}} identify the partner and its requirement in the
communication oracle; \texttt{\{\{SUBAGENT\_MODEL\}\}} is the model the orchestrator
must spawn; and \texttt{\{\{TEST\_FILES\}\}} lists the gold-test files in the
gold-test scenarios. The mapping from scenario to prompt is:

\begin{center}
\small
\begin{tabular}{@{}ll@{}}
\toprule
Scenario & Prompt (base $+$ overlay) \\
\midrule
Oracle & base (both requirements), Listing \ref{prompt:base_oracle} \\
Orchestrator & orchestrator prompt, Listing \ref{prompt:orchestrator} \\
Sequential / Reentrant & base (single) $+$ hand-off overlay, Listing \ref{prompt:base_single}, \ref{prompt:handoff} \\
Parallel unaware & base (single); no overlay \\
Parallel aware / message / sync & base (single) $+$ coordination overlay, Listing \ref{prompt:coord_overlays} \\
Communication oracle & communication-oracle prompt, Listing  \ref{prompt:comm_oracle} \\
Oracle $+$ gold tests & base (both) $+$ gold-test overlay, Listing  \ref{prompt:goldtests} \\
Parallel $+$ gold tests & base (single) $+$ gold-test overlay, Listing  \ref{prompt:goldtests} \\
\bottomrule
\end{tabular}
\end{center}

\paragraph{Base task prompts.}
The single-requirement base (used by every parallel and hand-off scenario, one
requirement per agent) and the two-requirement base (used by the single-agent oracle):

\begin{lstlisting}[
caption={Base task prompt, single-requirement form (one agent, one requirement).},
label={prompt:base_single},
basicstyle=\ttfamily\scriptsize,
breaklines=true,
columns=fullflexible,
keepspaces=true,
showstringspaces=false
]
You are working in a git repository. Resolve the following requirement in the codebase. Make the changes directly to the files.

## Requirement

{{PROBLEM}}
\end{lstlisting}

\begin{lstlisting}[
caption={Base task prompt, oracle form (one agent, both requirements).},
label={prompt:base_oracle},
basicstyle=\ttfamily\scriptsize,
breaklines=true,
columns=fullflexible,
keepspaces=true,
showstringspaces=false
]
You are working in a git repository. Resolve both of the following requirements in the codebase. Make the changes directly to the files.

## Requirement A

{{PROBLEM_A}}

## Requirement B

{{PROBLEM_B}}
\end{lstlisting}

\paragraph{Orchestrator.}
Replaces the base prompt for the orchestrator scenario; the acting agent delegates
both requirements to two subagents and does no implementation itself.

\begin{lstlisting}[
caption={Orchestrator prompt.},
label={prompt:orchestrator},
basicstyle=\ttfamily\scriptsize,
breaklines=true,
columns=fullflexible,
keepspaces=true,
showstringspaces=false
]
You are the orchestrator of a small team working in a git repository. There are two independent requirements to resolve, A and B. Your job is to coordinate the work, not to do it.

Hard rules:

- You MUST delegate all implementation to exactly two subagents -- one per requirement. Spawn two subagents and have each one implement its assigned requirement in full.
- You MUST NOT implement anything yourself. Do not edit, create, or fix any repository file directly. Every change to the codebase must be made by a subagent, not by you.
- Each of the two subagents MUST run on the model `{{SUBAGENT_MODEL}}`. Spawn them with that model.
- Your role is limited to: reading enough context to plan, deciding the assignment, dispatching the two subagents, and confirming their work is integrated into the working tree.

You are free to decide how to run this: which subagent takes which requirement, what instructions and context you give each, and whether you dispatch them concurrently or one after the other. The only fixed constraints are the three rules above -- two subagents, one per requirement, and no hands-on work by you.

## Requirement A

{{PROBLEM_A}}

## Requirement B

{{PROBLEM_B}}

When both subagents have finished, make sure the repository's working tree contains the combined result of both requirements.
\end{lstlisting}

\paragraph{Parallel coordination overlays.}
Appended to the single-requirement base. \emph{Unaware} adds no overlay. \emph{Aware}
states a peer exists but gives no channel. \emph{Message} and \emph{Sync} expose the
same send/receive interface, framed as free-form messaging or as explicit
synchronization points, respectively.

\begin{lstlisting}[
caption={Parallel coordination overlays (aware, message, sync). Unaware appends nothing.},
label={prompt:coord_overlays},
basicstyle=\ttfamily\scriptsize,
breaklines=true,
columns=fullflexible,
keepspaces=true,
showstringspaces=false
]
=== aware ===
Another agent is working in this same repository at the same time on a related requirement. You will not see their changes while you work.

=== message ===
Another agent is working in this same repository at the same time on a related requirement. You can exchange messages with them: run `./send-msg "your message"` to send, and `./recv-msg` to read any new messages from the other agent. Use this to share intent, interfaces, and assumptions. Check `./recv-msg` periodically, at least before you make interface decisions and before you finish.

=== sync ===
Another agent is working in this same repository at the same time on a related requirement. You can exchange messages with them: run `./send-msg "your message"` to send, and `./recv-msg` to read any new messages from the other agent. Treat the work as a sequence of sync points: before starting implementation, send your plan and the interfaces you intend to define, then run `./recv-msg` to read theirs; after each significant change, send it and run `./recv-msg` to reconcile; before finishing, confirm your assumptions against theirs. Do not finalize an interface decision without running `./recv-msg` first.
\end{lstlisting}

\paragraph{Communication oracle.}
Extends parallel messaging by giving each agent its partner's requirement verbatim
together with an instruction to jointly solve both.

\begin{lstlisting}[
caption={Communication-oracle prompt.},
label={prompt:comm_oracle},
basicstyle=\ttfamily\scriptsize,
breaklines=true,
columns=fullflexible,
keepspaces=true,
showstringspaces=false
]
You are agent {{AGENT}}. Another agent, {{PEER}}, is working in this same repository at the same time on the following requirement:

{{PEER_PROBLEM}}

You will not see their changes while you work. Both sets of changes are merged afterwards, and both requirements must hold after the merge. You can exchange messages with them: `./send-msg "text"` sends, `./recv-msg` prints new messages from the other agent, and `./recv-msg --wait` waits up to 90 seconds for one. Please communicate to jointly solve your respective requirements as efficiently as possible.
\end{lstlisting}

\paragraph{Sequential and reentrant hand-off overlays.}
Appended to the single-requirement base for the second and later turns, varying whether
the acting agent sees only the current file state (\emph{repo}) or also the earlier
agent's edit history (\emph{hist}).

\begin{lstlisting}[
caption={Hand-off overlays (repo-only and with-history).},
label={prompt:handoff},
basicstyle=\ttfamily\scriptsize,
breaklines=true,
columns=fullflexible,
keepspaces=true,
showstringspaces=false
]
=== repo (current file state only) ===
Another agent has already worked in this repository before you. You are picking up the current state of the files as they left it. Continue from that state.

=== hist (file state plus edit history) ===
Another agent has already worked in this repository before you. You are picking up the current state of the files as they left it, and their edit history is available in the git log (use `git log` and `git diff` to see what they changed and why). Continue from that state.
\end{lstlisting}

\paragraph{Gold-test access overlay.}
Appended in the \emph{oracle + gold tests} and \emph{parallel + gold tests} scenarios,
letting an agent run the gold tests for its own requirement(s) while it works.

\begin{lstlisting}[
caption={Gold-test access overlay.},
label={prompt:goldtests},
basicstyle=\ttfamily\scriptsize,
breaklines=true,
columns=fullflexible,
keepspaces=true,
showstringspaces=false
]
The tests that define "done" for your requirement(s) are provided in the `_gold_tests/` directory, and you can run them at any time with `./run-tests`. This applies your current changes in a clean copy of the project, runs those tests, and prints which pass and fail. Use it to check your work, and make sure they pass before you finish.

Your tests:
{{TEST_FILES}}

(`_gold_tests/` and `./run-tests` are harness scaffolding, excluded from your submission -- do not edit them or include them in your changes.)
\end{lstlisting}

\end{document}